\documentclass[10pt,twocolumn,letterpaper]{article}

\usepackage[pagenumbers]{cvpr} 

\usepackage{mathtools}
\usepackage{graphicx}
\usepackage{amsmath}
\usepackage{amssymb}
\usepackage{booktabs}
\usepackage{shtabularlines}
\usepackage{multirow}
\usepackage[table,xcdraw,usenames,dvipsnames]{xcolor}
\usepackage[normalem]{ulem}
\useunder{\uline}{\ul}{}
\definecolor{myblue}{RGB}{88, 190, 237}
\definecolor{myyellow}{RGB}{249, 227, 130}
\definecolor{mygreen}{RGB}{183, 221, 169}
\usepackage{multirow}
\usepackage{makecell}
\usepackage{tabularx}
\usepackage{enumitem}
\usepackage{rotating}
\usepackage{hyperref}
\usepackage{pifont}
\makeatletter
\@namedef{ver@everyshi.sty}{}
\makeatother
\usepackage{tikz}
\usepackage{wrapfig}

\definecolor{LightCyan}{rgb}{0.9059,0.9961,1}
\usepackage{multirow}
\usepackage{graphicx}
\usepackage[font=footnotesize,labelfont=bf,aboveskip=1pt,belowskip=-11pt]{caption}  
\usepackage{makecell}
\usepackage{tabulary}
\definecolor{demphcolor}{RGB}{144,144,144}

\usepackage{pifont}
\newlength\savewidth

\newcommand{\tablestyle}[2]{\setlength{\tabcolsep}{#1}\renewcommand{\arraystretch}{#2}\centering\footnotesize}
\makeatletter\renewcommand\paragraph{\@startsection{paragraph}{4}{\z@}
  {.5em \@plus1ex \@minus.2ex}{-.5em}{\normalfont\normalsize\bfseries}}\makeatother
\newdimen\abovecrulesep
\newdimen\belowcrulesep
\makeatletter
\patchcmd{\@@@cmidrule}{\aboverulesep}{\abovecrulesep}{}{}
\patchcmd{\@xcmidrule}{\belowrulesep}{\belowcrulesep}{}{}
\makeatother
\usepackage{etoolbox}
\preto\align{\par\nobreak\small\noindent}
\expandafter\preto\csname align*\endcsname{\nobreak\small}
\preto\equation{\small}
\newcommand{\cmark}{\ding{51}}%
\newcommand{\xmark}{\ding{55}}%
\renewcommand{\sout}[1]{\unskip}

\usepackage[capitalize]{cleveref}
\crefname{section}{Sec.}{Secs.}
\Crefname{section}{Section}{Sections}
\Crefname{table}{Table}{Tables}
\crefname{table}{Tab.}{Tabs.}

\newcommand{\ourmodel}{X-Decoder}

\newcommand{\xyz}[1]{{\color{myblue}{\small{\bf #1}}}}

\def\cvprPaperID{870} 
\def\confName{CVPR}
\def\confYear{2023}

\begin{document}
\title{Generalized Decoding for Pixel, Image, and Language}

\author{
\small{
Xueyan Zou$^{*\S}$, ~Zi-Yi Dou$^{*\sharp}$, ~Jianwei Yang$^{*}$\textsuperscript{\tiny $\ddagger\spadesuit$}, ~Zhe Gan$^{\dagger}$, ~Linjie Li$^{\dagger}$, ~Chunyuan Li$^{\ddagger}$, Xiyang Dai$^{\dagger}$, Harkirat Behl$^{\ddagger}$
}
\and
\small{
Jianfeng Wang$^{\dagger}$, Lu Yuan$^{\dagger}$, Nanyun Peng$^{\sharp}$, Lijuan Wang$^{\dagger}$, Yong Jae Lee\textsuperscript{$\P\S$}, Jianfeng Gao\textsuperscript{$\P\ddagger$}}
\and
{
\small
$^{\S}$ \textbf{University of Wisconsin-Madison} \;
$^{\sharp}$ \textbf{UCLA} \;
$^\ddagger$ \textbf{Microsoft Research at Redmond} \;  
$^{\dagger}$ \textbf{Microsoft Cloud \& AI} \;
}
\and
\footnotesize{
\small{$^*$Equal Technical Contribution} \;
\small{$\P$~Equal Advisory Contribution} \;
\small{$\spadesuit$~Project Lead} \;
}
\and
\centerline{\tt\tiny  
\{xueyan,yongjaelee\}@cs.wisc.edu 
\ \{zdou,violetpeng\}@cs.ucla.edu 
 \ \{jianwyan,jfgao,zhgan,linjli,chunyl,jianfw,luyuan,lijuanw,hbehl,xidai\}@microsoft.com
}
}

\twocolumn[{%
\renewcommand\twocolumn[1][]{#1}%
\maketitle
\vspace{-2.3em}
\begin{center}
    \centering
    \includegraphics[width=0.98\textwidth]{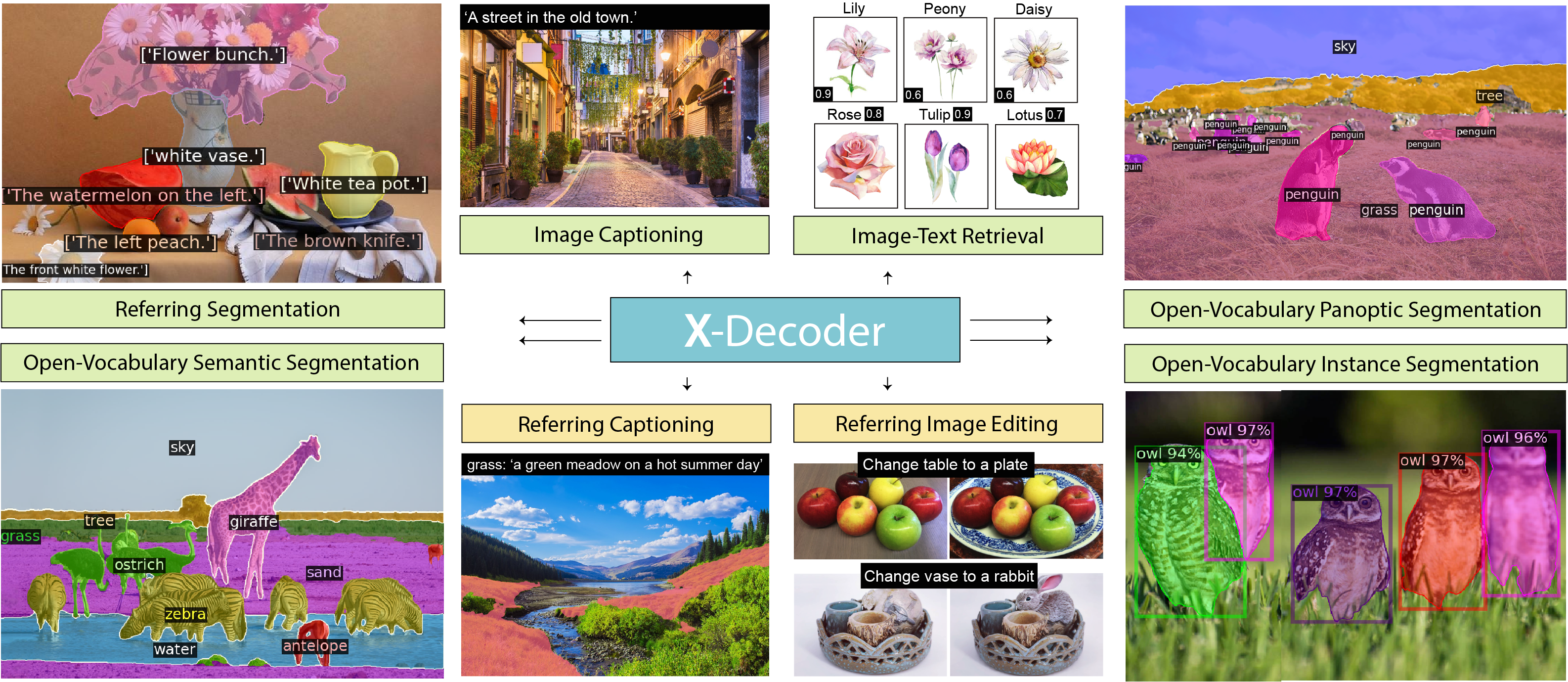}
\vspace{0.2em}
\captionof{figure}{With one suite of parameters, X-Decoder after pretraining supports all types of image segmentation tasks ranging from open-vocabulary instance/semantic/panoptic segmentation to referring segmentation, and vision-language tasks including image-text retrieval, and image captioning (labeled in \textcolor{LimeGreen}{green} boxes). It further empowers composite tasks like referring captioning using X-Decoder itself and image editing that combines with generative models such as Stable Diffusion~\cite{rombach2022high} (labeled in \textcolor{Goldenrod}{yellow} boxes).}
\label{fig:intro}
\vspace{1.5em}
\end{center}%
}]

\maketitle
\vspace{-3pt}
\begin{abstract}
\vspace{-4pt}
We present \ourmodel, a generalized decoding model that can predict pixel-level segmentation and language tokens seamlessly. \ourmodel~takes as input two types of queries: ($i$) generic non-semantic queries and ($ii$) semantic queries induced from text inputs, to decode different pixel-level and token-level outputs in the same semantic space. With such a novel design, \ourmodel~is the first work that provides a unified way to support all types of image segmentation and a variety of vision-language (VL) tasks. Further, our design enables seamless interactions across tasks at different granularities and brings mutual benefits by learning a common and rich pixel-level visual-semantic understanding space, without any pseudo-labeling. After pretraining on a mixed set of a limited amount of segmentation data and millions of image-text pairs, \ourmodel{} exhibits strong transferability to a wide range of downstream tasks in both zero-shot and finetuning settings. Notably, it achieves (1) state-of-the-art results on open-vocabulary segmentation and referring segmentation on eight datasets; (2) better or competitive finetuned performance to other generalist and specialist models on segmentation and VL tasks; and (3) flexibility for efficient finetuning and novel task composition (\textit{e.g.}, referring captioning and image editing shown in Fig.~\ref{fig:intro}). Code, demo, video and visualization are available at: \xyz{\url{https://x-decoder-vl.github.io}}.
\let\thefootnote\relax\footnotetext{The work is developed during an internship at Microsoft.}
\end{abstract}
\vspace{-6mm}
\section{Introduction}

Visual understanding at different levels of granularity has been a longstanding problem in the vision community. The tasks span from image-level tasks (\emph{e.g.}, image classification~\cite{deng2009imagenet}, image-text retrieval, image captioning~\cite{chen2015microsoftcoco}, and visual question answering (VQA)~\cite{antol2015vqa}), region-level localization tasks (\emph{e.g.}, object detection and phrase grounding~\cite{plummer2015flickr30k}), to pixel-level grouping tasks (\textit{e.g.}, image instance/semantic/panoptic segmentation~\cite{long2015fully,kirillov2019panoptic,hafiz2020survey}). Until recently, most of these tasks have been separately tackled with specialized model designs, preventing the synergy of tasks across different granularities from being exploited. In light of the versatility of transformers~\cite{vaswani2017attention}, we are now witnessing a growing interest in building general-purpose models that can learn from and be applied to a diverse set of vision and vision-language tasks, through multi-task learning~\cite{hu2021unit,gupta2022towards},  sequential decoding \cite{wang2022ofa,yang2022unitab,chen2022unified, lu2022unified}, or unified learning strategy~\cite{yuan2021florence,yang2022unified, yu2022coca, zhang2022glipv2}. While these works have shown encouraging cross-task generalization capabilities, most target the unification of image-level and region-level tasks, leaving the important pixel-level understanding underexplored. In~\cite{chen2022unified, lu2022unified}, the authors attempt to unify segmentation into a decoding of a coordinate sequence or a color map, which, however, produces suboptimal performance and limited support for open-world generalization.

Arguably, understanding images down to the pixel level is one of the most important yet challenging problems in that: (1) pixel-level annotations are costly and undoubtedly much more scarce compared to other types of annotations; (2) grouping every pixel and recognizing them in an open-vocabulary manner is less studied; and (3) more importantly, it is non-trivial to learn from data at two substantially different granularities while also obtaining mutual benefits. Some recent efforts have attempted to bridge this gap from different aspects. In~\cite{cheng2022masked}, Chen \textit{et~al.}~propose a unified architecture Mask2Former that tackles all three types of segmentation tasks but in a closed set. To support open vocabulary recognition, a number of works study how to transfer or distill rich semantic knowledge from image-level vision-language foundation models such as CLIP~\cite{radford2021learning} and ALIGN~\cite{jia2021scaling} to specialist models~\cite{ghiasi2021open,ding2022open,rao2022denseclip}. However, all these initial explorations focus on specific segmentation tasks of interest and do not show generalization to tasks at different granularities. In this work, we take one step further to build a generalized decoder called \ourmodel{}\footnote{Here, `X' denotes versatile, and also represents `piXel'.} towards the unification of pixel-level and image-level  vision-language understanding, as shown in Figure~\ref{fig:intro}.

\textbf{A generalized decoding framework.} 
We formulate all tasks including pixel-level image segmentation, image-level retrieval and vision-language tasks into a generic decoding procedure. Specifically, \ourmodel~is built on top of a vision backbone and a transformer encoder for extracting multi-scale image features, following the framework of Mask2Former~\cite{cheng2022masked}. The key novelty lies in the decoder design. First, it takes two sets of queries as input: ($i$) generic non-semantic queries that aim to decode segmentation masks for universal segmentation, similar to Mask2Former~\cite{cheng2022masked}, and ($ii$) newly introduced textual queries to make the decoder language-aware for a diverse set of language-related vision tasks. Second, it predicts two types of outputs: pixel-level masks and token-level semantics, and their different combinations can seamlessly support all tasks of interest. Third, we use a single text encoder to encode the textual corpus involved in all tasks, including concepts in segmentation, phrases in referring segmentation, tokens in image captioning and questions in VQA, \etc. As a result, our \ourmodel{} can naturally facilitate the synergy across tasks and advocate the learning of a shared visual-semantic space, while respecting the heterogeneous nature of different tasks. 

\textbf{An end-to-end learning paradigm.} With our generalized decoder design, we propose an end-to-end pretraining method to learn from all granularities of supervision. We unite three types of data: panoptic segmentation, referring segmentation, and image-text pairs. Unlike previous works that use pseudo-labeling techniques to extract fine-grained supervision from image-text pairs~\cite{zhang2022glipv2,ghiasi2021open}, \ourmodel{} directly groups and proposes a few meaningful segmentation candidates, so that it can map the regions easily to the contents described in the captions on the fly. Meanwhile, the referring segmentation task bridges generic segmentation and image captioning by sharing the pixel-level decoding with the former and semantic queries with the latter. 

\textbf{Strong zero-shot and task-specific transferability to a wide range of segmentation and VL tasks.} Pre-trained with a limited amount of segmentation data and millions of image-text pairs, our \ourmodel{} supports a diversity of tasks in a zero-shot and open-vocabulary manner. Concretely, our model can be directly applied for all three types of segmentation tasks in a wide range of domains, establishing new state-of-the-art on ten settings of seven datasets. When transferred to specific tasks, our model also exhibits consistent superiority to previous works. Finally, we observe some intriguing properties in our model that it can support some novel task compositions and efficient finetuning, thanks to the flexibility endowed by our model design.
\section{From Specialist to Generalist Models}

\subsection{Pixel-Level Understanding}

Pixel-level image understanding, also known as image segmentation, has been a long-standing problem~\cite{fu1981survey,pal1993review}. 

\textit{\textbf{Generic Segmentation.}} There are mainly three well-defined tasks for pixel-level understanding, including semantic~\cite{long2015fully}, instance~\cite{hafiz2020survey}, and panoptic~\cite{kirillov2019panoptic} segmentation. Semantic segmentation cares about the per-pixel semantic within an image~\cite{long2015fully, chen2017rethinking, chen2022vision}, whereas instance segmentation groups pixels of the same semantic meaning into object instances. Models for both tasks have evolved from CNN-based architectures~\cite{long2015fully} to transformer-based ones~\cite{chen2022vision}, and from two-stage models~\cite{he2017mask} to one-stage models~\cite{bolya2019yolact,tian2020conditional} and to the recent query-based approaches~\cite{dong2021solq,zou2022end}. With the capability of per-pixel and instance-level understanding, a natural step was taken to formulate panoptic segmentation~\cite{kirillov2019panoptic,wang2021max,cheng2022masked}. Most recently, Mask2Former~\cite{cheng2022masked}  proposed to address all three tasks with a unified encoder-decoder architecture. Nevertheless, all these works cope with a limited number of categories, \ie, models can hardly recognize concepts absent in the training set. In MSeg~\cite{lambert2020mseg}, the authors manually merge different datasets and train a more generalized model on the composite set, which is still limited to being a closed set.

\textit{\textbf{Open-Vocabulary Segmentation.}} Recently, a number of works opt to transfer or distill the rich visual-semantic knowledge from foundation models like CLIP~\cite{radford2021learning} and ALIGN~\cite{jia2021scaling} to specific segmentation tasks. Prominent examples include LSeg~\cite{li2022language}, OpenSeg~\cite{ghiasi2021open}, and \cite{huynh2022open}. Instead of using existing models, GroupViT~\cite{xu2022groupvit} performed language-image pretraining from scratch with a bottom-up grouping ViT~\cite{dosovitskiy2020image}, while DenseCLIP~\cite{rao2022denseclip} demonstrated the superiority of foundation models in finetuning settings compared with supervised models. Recently, MaskCLIP~\cite{ding2022open} proposed to tackle open-vocabulary panoptic and semantic segmentation by leveraging CLIP, and achieved SoTA performance on ADE20K~\cite{zhou2017scene} and PASCAL~\cite{mottaghi2014role,everingham2011pascal}.

\textit{\textbf{Referring Segmentation}} by nature is open-vocabulary in that it does not presume a fixed number of phrases in the training and inference times. Models are usually designed specifically to learn from target datasets using various multimodal fusion strategies~\cite{hu2016segmentation,liu2017recurrent,margffoy2018dynamic,ye2019cross,yu2016modeling}. Since the emergence of vision transformers, works like LAVT~\cite{yang2022lavt} enhance the cross-modal interactions from the very beginning, which led to SoTA on RefCOCO~\cite{yu2016modeling}, RefCOCO+~\cite{yu2016modeling} and G-Ref~\cite{mao2016generation,nagaraja2016modeling}. CLIPSeg~\cite{luddecke2022image} extended the textual query to a visual query and showed superior performance not only on referring segmentation but also on semantic segmentation.

In this work, we propose \ourmodel{}, which is the first model to tackle generic and referring segmentation tasks all in one model. Furthermore, the generalized decoder jointly learns from segmentation data and image-text pairs end-to-end, and thus can augment the synergy across tasks for rich pixel-level and image-level understanding.

\subsection{Vision-Language Understanding}
Vision-language (VL) pretraining has proven to be effective for various VL tasks~\cite{lu2019vilbert,tan-bansal-2019-lxmert,su2019vl,li2019visualbert}. The field has evolved from a transformer fusion model~\cite{chen2020uniter,li2020oscar,zhang2021vinvl} with pre-extracted object features~\cite{anderson2018bottom} to end-to-end transformers~\cite{kim2021vilt,li2021align,dou2021empirical}, that directly learn from raw image pixels.
Recently, researchers~\cite{wang2021simvlm,beit3,singh2022flava} have found that image-text data at scale can be helpful for visual representation learning (\eg, enabling zero-shot image classification~\cite{radford2021learning,jia2021scaling} and action recognition~\cite{yuan2021florence,yu2022coca}). VL pre-trained models can be further extended to region-level tasks, such as phrase grounding and open-vocabulary object detection~\cite{kamath2021mdetr,gu2021open,zhong2022regionclip,minderer2022simple}, and unified frameworks that aim to combine image-text pairs with region-level data have also been proposed~\cite{cai2022x,li2022grounded,zhang2022glipv2,yao2022detclip,fiber2022}. A comprehensive review on this topic is provided in~\cite{gan2022vision}.

We are witnessing 
a clear trend from building specialist models to generalist ones. 
Early efforts~\cite{hu2021unit,gupta2022towards} build a multi-task learning paradigm to accommodate a diversity of tasks. 
However, the interactions among different tasks in these works are less studied, and the combination usually leads to performance degradation compared with specialist models. Recently, 
a number of works aim to reformulate the tasks into a unified sequential decoding process~\cite{chen2022unified,yang2022unitab,wang2022ofa,lu2022unified,kolesnikov2022uvim}. In this work, instead of developing a unified interface for vision and VL tasks, our \ourmodel{} builds a generalized decoding paradigm that can seamlessly connect the tasks by taking the common (\eg, semantic) but respecting the natural differences (\eg, spatial mask \textit{v.s.} sequential language), leading to significant improvements for different segmentation and VL tasks across the board.
\section{X-Decoder}
\begin{figure}
    \centering
    \includegraphics[width=.8\linewidth]{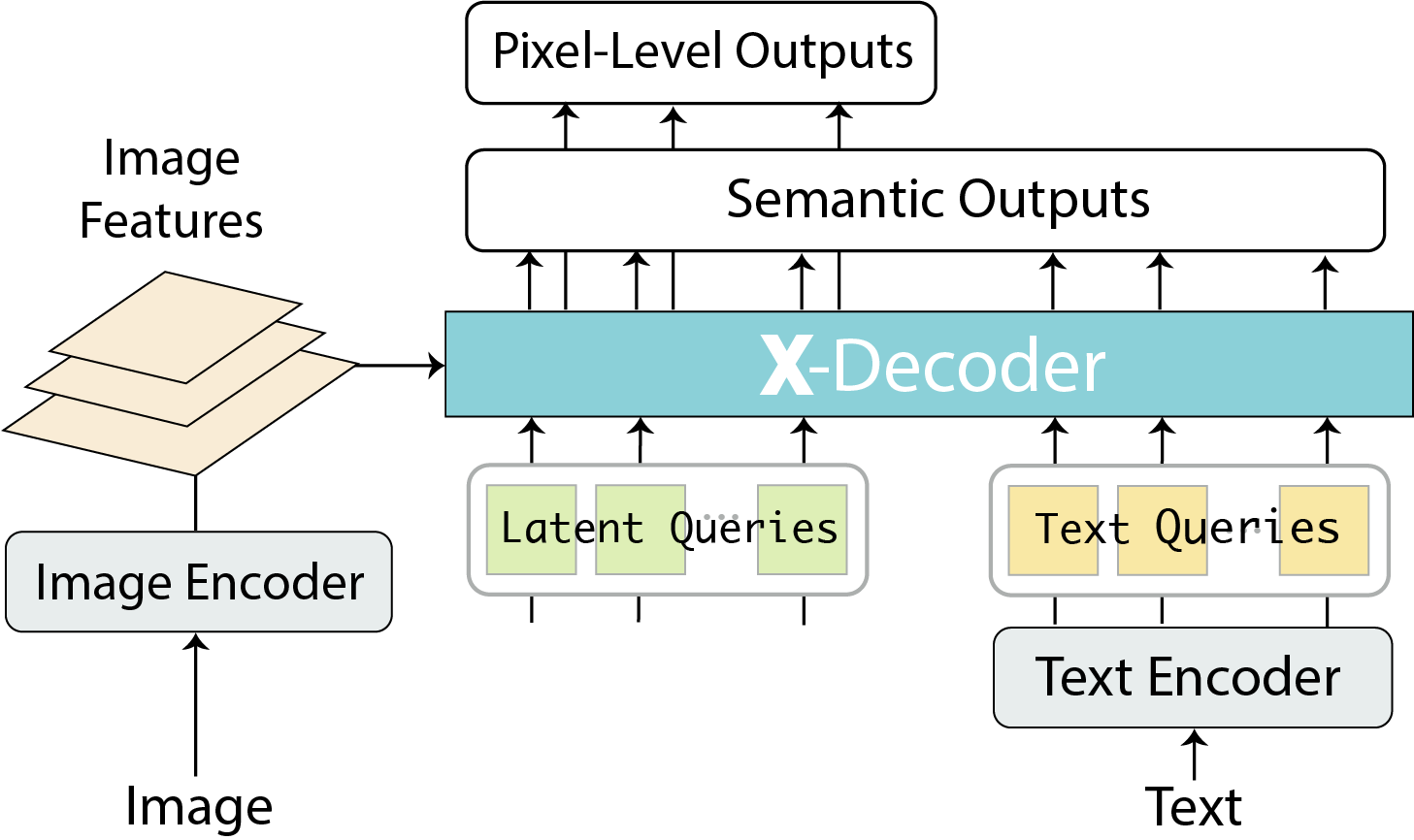}
    \caption{Overall pipeline for our model. It consists of an image encoder, a text encoder and our own designed \ourmodel{}.}
    \label{fig:xdecoder_overall}
\end{figure}

\begin{figure*}[t]
    \centering
    \includegraphics[width=1.0\textwidth]{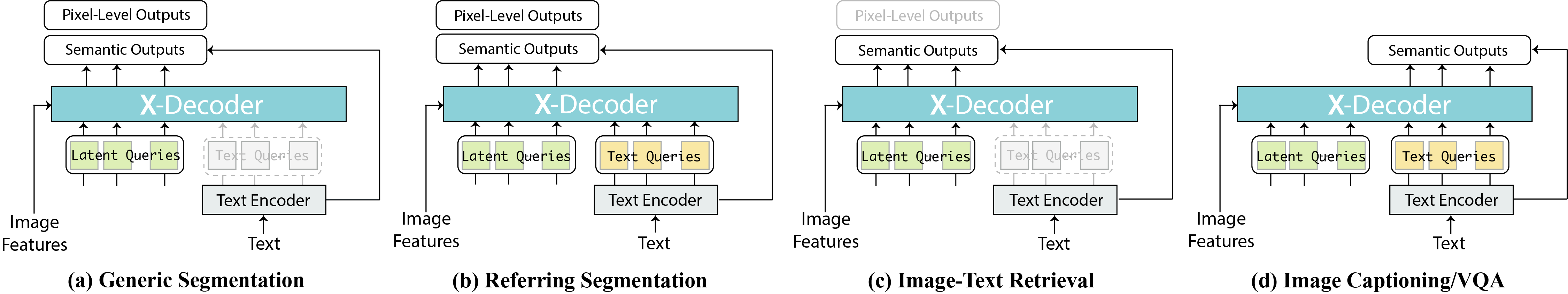}
    \caption{Unifying four different types of tasks with our proposed \ourmodel{}. From left to right, they are: (a) generic semantic/instance/panoptic segmentation; (b) referring segmentation; (c) image-text retrieval and (d) image captioning and VQA. The components with white text indicate not applied.}
    \label{fig:task_unification}
\end{figure*}

\subsection{Formulation}

Our model follows the generic design of encoder-decoder architecture as shown in Fig.~\ref{fig:xdecoder_overall}. Given an input image $\mathbf{I} \in \mathcal{R}^{H \times W \times 3}$, we first use an image encoder $\mathbf{Enc}_I$ to extract features $\mathbf{Z}$. Afterwards, we use the text encoder $\mathbf{Enc}_T$ to encode a textual query $\mathbf{T}$ into {$\mathbf{Q^t}=\langle q^t_1, \cdots, q^t_n\rangle$} of length $n$. The visual features, textual queries and the $m$ non-semantic or latent queries {$\mathbf{Q^h}=\langle q^h_1, \cdots, q^h_m\rangle$} are fed to our \ourmodel{} to predict the outputs:
\begin{equation}
    \langle\mathbf{O^p}, \mathbf{O^s}\rangle = \mathbf{XDec}\left(\langle\mathbf{Q^h}, \mathbf{Q^t}\rangle; \mathbf{Z}\right)
    \label{EQ:XDecoder_overall}
\end{equation}
where $\mathbf{O^p}$ and $\mathbf{O^s}$ are the pixel-level masks and token-level semantics, respectively. In the above formula, we note three critical designs to empower the generalization ability of our \ourmodel{} to a variety of vision and vision-language tasks.

\smallskip
\noindent
\textbf{We define two types of queries and outputs for \ourmodel{}}. As discussed earlier, the queries for the decoder are categorized into latent queries $\mathbf{Q^h}$ and text queries $\mathbf{Q^t}$, which undertake generic vision and vision-language tasks, respectively, and their combinations can further support various language-aware tasks such as referring segmentation, VQA, \textit{etc}. Likewise, the output is categorized into pixel-level mask $\mathbf{O}^p$ and semantic embedding $\mathbf{O}^s$. By simply using different combinations, we can adapt our \ourmodel{} to various tasks with the same suite of parameters.

\noindent
\textbf{We employ a single text encoder $\mathbf{Enc}_T$ to encode the textual corpus from all tasks.} The common text encoder is used to encode referring phrases, text descriptions, image captions in the task of referring segmentation, image-text retrieval and image captioning, respectively. Furthermore, we reformulate the mask classification in segmentation into a mask-text matching problem between $\mathbf{O^s}$ and the textual embeddings of prompted textual concepts similar to~\cite{yang2022unified,ghiasi2021open}. Sharing the text encoder for all textual corpus could maximally exchange knowledge from different tasks and learn a richer and more coherent semantic space.

\noindent
\textbf{We fully decouple the image and text encoder}. In many previous unified encoder-decoder models~\cite{kamath2021mdetr,yang2022unitab,chen2022unified}, the image and text are fused in the encoder side. This design makes it intractable not only for global image-text contrastive learning~\cite{radford2021learning,yang2022unified}, but also generative pretraining~\cite{wang2022git}. In contrast, by fully decoupling the image and text encoder and using the outputs all as queries, \ourmodel{} can learn from both intra-image supervisions and inter-image ones, which is essential to learn stronger pixel-level representations and support different granularity of tasks.
\vspace{-2pt}

\subsection{Unification of Tasks}
\label{subsec:task_unification}
Based on the above designs, \ourmodel{} can be used to seamlessly unify different vision and vision-language tasks, simply with different combinations of queries as inputs.
     \textit{\textbf{Generic Segmentation}}. For this task, there are no textual queries as inputs. Hence, Eq.~\eqref{EQ:XDecoder_overall} becomes:
    \begin{equation}
        \langle\mathbf{O^p}, \mathbf{O^s}\rangle = \mathbf{XDec}(\mathbf{Q^h}; \mathbf{Z})
    \label{EQ:XDecoder_generic_seg}        
    \end{equation}
    where $\mathbf{O^p}$, $\mathbf{O^s}$ have the same size of $\mathbf{Q^h}$. Eq.~\eqref{EQ:XDecoder_generic_seg} reduces to Mask2former~\cite{cheng2022masked}, but with open-vocabulary capacity since we use mask-text matching for mask classification. 
    
     \textit{\textbf{Referring Segmentation}}. It requires both latent and text queries as inputs, thus shares the same formula as Eq.~\eqref{EQ:XDecoder_overall}. Similar to generic segmentation, we only use the first $m$ decoded outputs corresponding to the latent queries. Compared with Eq.~\eqref{EQ:XDecoder_generic_seg}, referring segmentation can be regarded as language-conditioned generic segmentation.
    
     \textit{\textbf{Image-Text Retrieval}.}  The decoupled image and text encoder in our \ourmodel{} makes it straightforward for inter-image retrieval tasks. Specifically, we only feed the latent queries to the decoder and obtain the semantic representation of an image:
    \begin{equation}
    \vspace{-4pt}
    \mathbf{O^s} = \mathbf{XDec}\left(\mathbf{Q^h}; \mathbf{Z}\right)
    \label{EQ:XDecoder_retrieval}
    \end{equation}
    where $\mathbf{O^s}$ has the same length as $\mathbf{Q^h}$, and the last~($m$-th) token in $\mathbf{O^s}$ is then used to compute the similarities between images and texts.
    
     \textit{\textbf{Image Captioning and VQA}.} For both tasks, \ourmodel{} takes both latent and text queries and decodes the outputs:
    \begin{equation}
        \mathbf{O^s} = \mathbf{XDec}\left(\langle\mathbf{Q^h}, \mathbf{Q^t}\rangle; \mathbf{Z}\right)
    \label{EQ:XDecoder_captioning}
    \end{equation}
    where $\mathbf{O^s}$ correspondingly has equal size to $\mathbf{Q^t}$, and no masks are predicted. There are two slight differences between the two tasks. First, the caption prediction follows a causal masking strategy while VQA does not. Second, we use all the outputs in $\mathbf{O^s}$ for captioning, but only the last one to predict the answer for VQA. 
    
The adaptation of our \ourmodel{} to each task is further depicted in Fig.~\ref{fig:task_unification}. Based on this unification, we can pretrain our \ourmodel{} jointly with all tasks using a proper combination of queries and losses, and further finetune for individual tasks without any extra heads.\footnote{VQA is used for pretraining following common practice.} As discussed earlier, a lineup of works exploited a sequential decoding interface for the unification~\cite{cho2021unifying,wang2022unifying,chen2022unified,chen2022unified,yang2022unitab,lu2022unified}. 
However, in this work, {we advocate the unification by \emph{functionality} rather than interface}, namely, we maximally share the common parts of different tasks while keeping the remaining unchanged for individual tasks. 

\subsection{Unified Architecture}
We follow Mask2Former~\cite{cheng2022masked} to build our decoder architecture. Given an image $\mathbf{I} \in \mathcal{R}^{H \times W \times 3}$, we extract hierarchical visual features from $L$ layers:
\begin{equation}
    \mathbf{Z} = \mathbf{Enc}_I(\mathbf{I}) = \langle \mathbf{z}_l\rangle_{l=1}^{L}
\end{equation}
where $\mathbf{z}_l \in \mathcal{R}^{H_l\times W_l \times d}$ and $\{H_l, W_l\}$ is the size of feature map at level $l$ and $d$ is the feature dimension. These hierarchical feature maps are important for pixel-level understanding at different scales.

\smallskip\noindent\textbf{One Decoder $\mathbf{XDec}$ for All Tasks.} Given the visual features $\mathbf{Z}$, \ourmodel{} uses a stack of transformer layers to refine the queries and render the outputs. At layer $l$, it first cross-attends the visual features and then performs self-attention among latent and text queries:

\begin{align}
    \langle\hat{\mathbf{Q}}^h_{l-1}, \hat{\mathbf{Q}}^t_{l-1}\rangle &= \mathbf{CrossAtt}(\langle{\mathbf{Q}}^h_{l-1}, {\mathbf{Q}}^t_{l-1}\rangle; \mathbf{Z}) \label{EQ:crossattn} \\
    \langle{\mathbf{Q}}^h_{l}, {\mathbf{Q}}^t_{l}\rangle &= \mathbf{SelfAtt}(\langle\hat{\mathbf{Q}}^h_{l-1}, \hat{\mathbf{Q}}^t_{l-1}\rangle) \label{EQ:selfattn}
\end{align}
In Eq.~\eqref{EQ:crossattn}, we let all queries cross-attend the visual features. For latent queries, we use a masked cross-attention mechanism as in~\cite{cheng2022masked}, and full attention for the textual queries. In Eq.~\eqref{EQ:selfattn}, we specifically design the self-attention mechanism to prompt the synergy of tasks: $(i)$ we use the last latent query to extract the global image representation and the remaining for generic segmentation; $(ii)$ for image captioning, each textual query can attend itself, its predecessors and all latent queries; $(iii)$ for referring segmentation, latent queries will attend all text queries to use it as the language condition. 

\begin{table*}[!ht]
\centering
\tablestyle{3pt}{0.95}
\resizebox{0.96\linewidth}{!}{
\begin{tabular}{lc|ccc|ccc|c|cc|cc|cc|cc} 
\toprule
\multirow{3}{*}{Method}                          & \multirow{3}{*}{Type}                                                          & \multicolumn{6}{c|}{\uline{\it{Generic Segmentation}}}                                                       & \uline{\it{Referring}}          & \multicolumn{4}{c|}{\uline{\it{Retrieval}}}                                                & \multicolumn{2}{c|}{\uline{\it{Captioning}}}    & \multicolumn{2}{c}{\uline{\it{VQA}}}         \\
                                                 &                                                                                & \multicolumn{3}{c}{ADE}                                  & \multicolumn{3}{c|}{COCO}                         & g-Ref              & \multicolumn{2}{c}{COCO-Karpathy}        & \multicolumn{2}{c|}{F30k-Karpathy} & \multicolumn{2}{c|}{COCO-Karpathy} & \multicolumn{2}{c}{VQAv2-test}  \\
                                                 &                                                                                & PQ            & mAP           & \multicolumn{1}{c}{mIoU} & PQ                & mAP           & mIoU          & cIoU               & IR@1          & \multicolumn{1}{c}{TR@1} & IR@1          & TR@1               & CIDEr              & BLEU          & dev           & std             \\ 
\hline
Mask2Former (T)~\cite{cheng2022masked}                                 & \multirow{6}{*}{Segmentation}                                                  & 39.7          & 26.4          & 47.7                     & 53.2              & 43.3          & 63.2          & -                  & -             & -                        & -             & -                  & -                  & -             & -             & -               \\
Mask2Former (B)~\cite{cheng2022masked}                                 &                                                                                & $\star$             & $\star$             & 53.9                     & 56.4              & 46.3          & 67.1          & -                  & -             & -                        & -             & -                  & -                  & -             & -             & -               \\
Mask2Former (L)~\cite{cheng2022masked}                                 &                                                                                & 48.1          & 34.2          & 56.1                     & 57.8              & \textbf{48.6} & 67.4          & -                  & -             & -                        & -             & -                  & -                  & -             & -             & -               \\
Pano/SegFormer (B)~\cite{li2022panoptic,xie2021segformer}                            &                                                                                & $\star$             & $\star$             & 51.0                     & 55.4              & $\star$             & $\star$             & -                  & -             & -                        & -             & -                  & -                  & -             & -             & -               \\
kMaX-DeepLab (L)~\cite{yu2022k}                                &                                                                                & 48.7          & $\star$             & 54.8                     & \textbf{58.1}     & $\star$             & $\star$             & -                  & -             & -                        & -             & -                  & -                  & -             & -             & -               \\
LAVT (B)~\cite{yang2022lavt}                                        &                                                                                & -             & -             & -                        & -                 & -             & -             & 61.2               & -             & -                        & -             & -                  & -                  & -             & -             & -               \\ 
\hline
UNITER (B)~\cite{chen2020uniter}                                      & \multirow{6}{*}{\begin{tabular}[c]{@{}c@{}}Vision Language\\(VL)\end{tabular}} & -             & -             & -                        & -                 & -             & -             & -                  & 50.3          & 64.4                     & 72.5          & 85.9               & -                  & -             & 72.7          & 72.9            \\
UNITER (L)~\cite{chen2020uniter}                                      &                                                                                & -             & -             & -                        & -                 & -             & -             & -                  & 52.9          & 65.6                     & 75.6          & 87.3               & -                  & -             & 73.8          & 74.0            \\
VinVL (B)~\cite{zhang2021vinvl}                                       &                                                                                & -             & -             & -                        & -                 & -             & -             & -                  & 58.1          & 74.6                     & $\star$             & $\star$                  & 129.3              & 38.2          & 76.0          & 76.1            \\
VinVL (L)~\cite{zhang2021vinvl}                                       &                                                                                & -             & -             & -                        & -                 & -             & -             & -                  & \textbf{58.8} & 75.4                     & $\star$             & $\star$                  & 130.8              & 38.5          & 76.5          & 76.6            \\
ALBEF-4M (B)~\cite{li2021align}                                    &                                                                                & -             & -             & -                        & -                 & -             & -             & -                  & 56.8          & 73.1                     & 82.8          & 94.3               & $\star$                  & $\star$             & 74.5          & 74.7            \\
METER-Swin (B)~\cite{dou2021empirical}                                  &                                                                                & -             & -             & -                        & -                 & -             & -             & -                  & 54.9          & 73.0                     & 79.0          & 92.4               & $\star$                  & $\star$             & 76.4          & 76.4            \\ 
\hline
\uline{UViM} (L)~\cite{kolesnikov2022uvim}                                &                                                                                & $\star$             & $\star$             & $\star$                        & 45.8 $^{1}$ & $\star$             & $\star$             & -                  & -             & -                        & -             & -                  & -                  & -             & -             & -               \\
UniT (T)~\cite{hu2021unit}                                         & \multirow{10}{*}{General Purpose}                                              & -             & -             & -                        & -                 & -             & -             & -                  & -             & -                        & -             & -                  & -                  & -             & 67.6          & $\star$               \\
\uline{GPV} (T)~\cite{gupta2022towards}                                  &                                                                                & -             & -             & -                        & -                 & -             & -             & -                  & -             & -                        & -             & -                  & 102.3~$^{2}$ & $\star$             & 62.5          & $\star$               \\
UniTAB (B)~\cite{yang2022unitab}                                     &                                                                                & -             & -             & -                        & -                 & -             & -             & -                  & -             & -                        & -             & -                  & 119.8              & 36.1          & 70.7          & 71.0            \\
Pix2Seq v2 (B)~\cite{chen2022unified}                                  &                                                                                & -             & $\star$             & -                        & -                 & 38.2          & -             & -                  & -             & -                        & -             & -                  & $\star$                  & 34.9          & -             & -               \\ 

\uline{Unified-IO} (B)~\cite{lu2022unified}                          &                                                                                & -             & $\star$             & -                        & -                 & $\star$             & -             & -                  & -             & -                        & -             & -                  & $\star$                  & $\star$             & 61.8          & $\star$               \\
\uline{Unified-IO} (L)~\cite{lu2022unified}                          &                                                                                & -             & $\star$             & -                        & -                 & $\star$             & -             & -                  & -             & -                        & -             & -                  & $\star$                  & $\star$             & 67.8          & $\star$               \\
GLIPv2 (T)~\cite{zhang2022glipv2}                                       &                                                                                & -             & $\star$             & -                        & -                 & -/42.0        & -             & $\star$                  & -             & -                        & -             & -                  & 122.1              & $\star$             & 71.6          & 71.8            \\
GLIPv2 (B)~\cite{zhang2022glipv2}                                       &                                                                                & -             & $\star$             & -                        & -                 & -/45.8        & -             & $\star$                  & -             & -                        & -             & -                  & 128.5              & $\star$             & 73.1          & 73.3            \\
GLIPv2 (H)~\cite{zhang2022glipv2}                                       &                                                                                & -             & $\star$             & -                        & -                 & -/48.9        & -             & $\star$                  & -             & -                        & -             & -                  & 131.0              & $\star$             & 74.6          & 74.8            \\
\hline
\rowcolor[rgb]{0.961,0.961,0.961} X-Decoder (T)  & {\cellcolor[rgb]{0.961,0.961,0.961}}                                           & 41.6          & 27.7          & 51.0                     & 52.6              & 41.3/42.3     & 62.4          & 59.8 $\lvert$ 61.9          & 49.3          & 66.7                     & 74.4          & 89.1               & 122.3              & 37.8          & 70.6          & 70.9            \\
\rowcolor[rgb]{0.961,0.961,0.961} X-Decoder (B)  & {\cellcolor[rgb]{0.961,0.961,0.961}}                                           & 46.8          & 33.5          & 54.6                     & 56.2              & 45.8/45.8     & 66.0          & 62.4 $\lvert$ 64.5          & 54.5          & 71.2                     & 80.8          & 93.2               & 129.0              & 39.6          & 74.1          & 74.2            \\
\rowcolor[rgb]{0.961,0.961,0.961} X-Decoder (L)  & {\cellcolor[rgb]{0.961,0.961,0.961}}                                           & \textbf{49.6} & \textbf{35.8} & \textbf{58.1}            & 56.9             & 46.7/47.1     & \textbf{67.5} & \textbf{64.6 $\lvert$ 64.6} & 58.6          & \textbf{76.1}            & \textbf{84.4} & \textbf{94.4}      & \textbf{ 132.1}    & \textbf{40.2} & \textbf{76.8} & \textbf{77.0}   \\
\bottomrule
\end{tabular}
}
\vspace{3pt}
\caption{\textbf{Task-specific transfer} of \ourmodel{} to different segmentation and VL tasks. Note: ``$\star$" denotes the model has the capability for the task but does not have number reported. ``-" means the model does not have the ability for the specific task. ``L*" is the large model with deformable encoder. ``\uline{model name}" means the model does not have task specific finetune. ``1" is the reported pretrained number for UViM, the corresponding X-Decoder (L) has pretrained PQ 56.7. ``2" is the reported coco test2014 value for GPV. ``a$\lvert$b" means ``pretrain$\lvert$finetune". ``a/b" indicate ``val/test". }
\label{tab:task_transfer}
\end{table*}

\begin{figure}
    \centering
    \includegraphics[width=.98\linewidth]{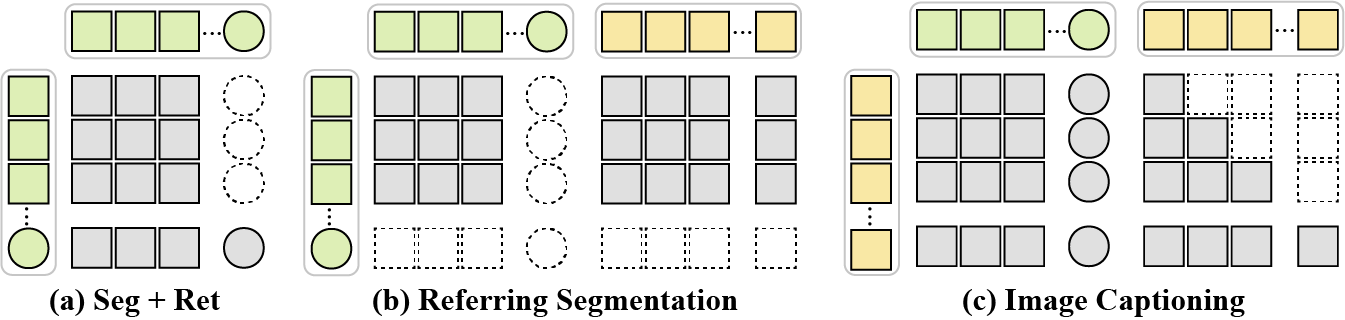}
    \caption{Interaction among latent queries (\textcolor{LimeGreen}{gree}), between latent and text queries (\textcolor{Goldenrod}{yellow}) for (a) Generic segmentation and image/text retrieval (b) referring segmentation and (c) image captioning. The square latent query is designated for image-text retrieval.}
    \label{fig:selfattention_mechanism}
\end{figure}

Based on these rules, the resulting self-attention in our \ourmodel{} is shown in Fig.~\ref{fig:selfattention_mechanism}. 

The output of our \ourmodel{} is also categorized into two types: 1) pixel-wise mask and 2) semantic outputs. \ourmodel{} always produces the masks only for the $m$ latent queries, \ie, $\mathbf{O}^{p} = \{o_1^{p},\cdots,o_m^{p}\} \in \{0,1\}^{m \times H \times W}$ for all the latent queries. As for the semantic outputs, \ourmodel{} predicts the outputs for both latent and text queries, \ie, $\mathbf{O}^{s} = \{o_1^{s},\cdots,o_{m+n}^{s}\} \in \mathcal{R}^{(m+n) \times d}$, to cover both mask recognition and caption generation.

\smallskip\noindent\textbf{One Encoder $\mathbf{Enc}_T$ for All Texts.} Our text encoder consists of a number of transformer layers. Given the raw text such as a phrase or caption, we convert it to discrete tokens using an off-the-shelf tokenizer and then send it to the text encoder. We apply causal masking  to ensure its outputs are compatible with caption decoding. For segmentation, we follow~\cite{radford2021learning,yang2022unified} to convert the class name into a phrase with a text prompt (\eg, ``dog'' $\rightarrow$ ``an image of dog''), and encode the phrase as above. 

\subsection{End-to-End Pre-training}

We train our \ourmodel{} in an end-to-end manner with two types of losses corresponding to the outputs.

\noindent
\textbf{Semantic Loss}. There are three losses on the semantic outputs corresponding to three tasks. For image-text retrieval, we compute the language-image contrastive loss as~\cite{radford2021learning}. We take the last valid token feature of $\mathbf{Q^t}$ from the text encoder to represent a text as $\hat{q}^t$ and take the last entry in $\mathbf{O}^s$ derived from \ourmodel{} as $\hat{o}^s$. As a result, we obtain $B$ pairs of features $\langle \hat{q}^t_i, \hat{o}^s_i\rangle_{i=1}^{B}$ for a minibatch of $B$ image-text pairs. Afterwards, we compute the dot-product between these $B\times B$ feature pairs to obtain an affinity matrix $\mathbf{S}_{it} \in \mathcal{R}^{B\times B}$, and compute the bidirectional cross-entropy loss:
\begin{equation}
    \mathcal{L}_{it} = \mathbf{CE}(\mathbf{S}_{it}, \mathbf{y}_{it}) + \mathbf{CE}(\mathbf{S}^T_{it}, \mathbf{y}_{it})
\end{equation}
where $\mathbf{y}_{it}$ are the class labels corresponding to diagonal entries in $\mathbf{S}_{it}$, and $\mathbf{S}_{it}^T$ is the transpose of $\mathbf{S}_{it}$.

For mask classification, we encode all $C$ class names including ``background'' into $C$ text queries and take the last valid token feature from each to represent the concept. Afterward, we take the decoder outputs corresponding to the first $(m-1)$ latent queries and compute the dot-product between these outputs and concept embeddings to obtain an affinity matrix $\mathbf{S}_{cls} \in \mathcal{R}^{(m-1) \times C}$ and compute the loss $\mathcal{L}_{cls} = \mathbf{CE}(\mathbf{S}_{cls}, \mathbf{y}_{cls})$, with the ground-truth class $\mathbf{y}_{cls}$.

For image captioning, we first extract the embeddings for all tokens in the vocabulary of size $V$ from the text encoder. Given the last $n$ semantic outputs from \ourmodel{}, we compute the dot-product with all token embeddings to obtain an affinity matrix $\mathbf{S}_{cap} \in \mathcal{R}^{n \times V}$. Then we compute the cross-entropy loss $\mathcal{L}_{cap} = \mathbf{CE}(\mathbf{S}_{cap}, \mathbf{y}_{cap})$, with the ground-truth next-token id $\mathbf{y}_{cap}$.

\noindent
\textbf{Mask Loss}. Given the predictions $\langle \mathbf{O}^{p}, \mathbf{O^s}\rangle$ derived from $m$ latent queries, we use Hungarian matching~\cite{carion2020end,cheng2022masked} to find the matched entries of first $(m-1)$ outputs to ground-truth annotations. Afterward, we follow~\cite{cheng2022masked} to use binary cross-entropy loss $\mathcal{L}_{bce}$ and dice loss $\mathcal{L}_{dice}$ to compute the loss for masks. We combine the above four losses to pretrain our \ourmodel{}. More details can be found in Appendix.
\section{Experiments}
\subsection{Experimental Setup}
\noindent

\textbf{Datasets and Settings.}
We pretrain \ourmodel{} on three types of data including panoptic segmentation, image-text pairs (itp), and referring segmentation. For panoptic and referring segmentation, we use 
COCO2017~\cite{lin2014microsoft} with segmentation annotations and exclude the validation sets of Ref-COCOg UMD~\cite{yu2016modeling} and COCO Karpathy~\cite{yin2017obj2text}. In total, there are 104k images for segmentation pretraining,  out of which 30k images are with referring segmentation annotations. For image-text pairs, we use the standard 4M corpora, including Conceptual Captions~\cite{sharma2018conceptual}, SBU Captions~\cite{ordonez2011im2text}, Visual Genome~\cite{krishna2017visual}, and COCO Captions~\cite{chen2015microsoft}. We broadly evaluate our models on all tasks covered by pretraining, including generic (Semantic/Instance/Panoptic) segmentation, referring segmentation, image-text retrieval, and image captioning. In particular, we benchmark on 10 settings of 7 datasets covering a wide range of domains. Moreover, we finetune and report results on VQA for fine-grained visual reasoning.

\smallskip
\noindent
\textbf{Implementation Details.}
Our visual encoder follows~\cite{cheng2022masked} to use 100 latent queries and 9 decoder layers for segmentation, and we add one additional latent query for image-level task. However, we do not adopt a deformable encoder as it does not generalize well to open-vocabulary settings (see in Appendix). We adopt Focal-T~\cite{yang2022focal} and DaViT-B/L~\cite{ding2022davit} as the vision encoder and a transformer text encoder with causal masking~\cite{radford2021learning, yuan2021florence} as language encoder. The models are pretrained on large-scale image-text data~\cite{yuan2021florence} (Base or Large) or UniCL~\cite{yang2022unified} for the tiny model. During pretraining, we set a minibatch for segmentation to $32$ and image-text pairs to $1024$. The image resolution is set to $1024$ for segmentation and $224$ for image-text data respectively. We follow a similar balanced sampling strategy in~\cite{yang2022unified} to ensure the segmentation data are always observed for a consistent number of epochs, regardless of the total number of image-text pairs. Based on this, we pretrain all models for 50 epochs using AdamW~\cite{loshchilov2017decoupled} as the optimizer. During finetuning, we have task-specific designs, please refer to details in Appendix.

\begin{table*}
\tablestyle{2.4pt}{1.0} 
\centering
\label{table:zero_shot}
\resizebox{0.99\linewidth}{!}{
\begin{tabular}{lccccccc|cccccccccccccccc} 
\toprule
\multirow{2}{*}{Model}                                 & \multicolumn{3}{c}{COCO (p/s)} & \multirow{2}{*}{ITP} & \multirow{2}{*}{Fix} & \multirow{2}{*}{EM} & \multirow{2}{*}{\begin{tabular}[c]{@{}c@{}}Pse-\\udo\end{tabular}} & \multicolumn{3}{c}{ADE-150}                                                                         & A-857                                   & VOC                                      & PC-59         & PC-459        & SUN           & \multicolumn{2}{c}{SCAN-20}   & SCAN-41       & \multicolumn{3}{c}{Cityscapes}                                                                      & \multicolumn{2}{c}{BDD}        \\
                                                       & m & cls & cap                  &                      &                      &                     &                                                                    & PQ                                       & mAP                                      & mIoU          & mIoU                                    & mIoU                                     & mIoU          & mIoU          & mIoU          & mIoU          & PQ            & mIoU          & mIoU          & mAP                                      & PQ                                       & mIoU          & PQ             \\ 
\hline
MSeg~(B)~\cite{lambert2020mseg}                                               & \cmark & \cmark   & \xmark                    & \xmark                    & \xmark                    & \xmark                   & \xmark                                                                  & \textcolor[rgb]{0.784,0.784,0.796}{33.7} & \textcolor[rgb]{0.784,0.784,0.796}{32.6} & 19.1          & $\star$                                       & 73.4                                     & 43.4          & $\star$             & 29.6          & 33.4          & $\star$             & $\star$             & 46.9          & \textcolor[rgb]{0.784,0.784,0.796}{24.8} & \textcolor[rgb]{0.784,0.784,0.796}{51.1} & 44.9          & $\star$              \\
GroupViT~(S)                                           & \xmark & \xmark   & \xmark                    & \cmark                    & \xmark                    & \cmark                   & \xmark                                                                  & -                                        & -                                        & $\star$             & $\star$                                       & 52.3                                     & 22.4          & $\star$             & $\star$             & $\star$             & -             & $\star$             & $\star$             & -                                        & -                                        & $\star$             & -              \\
LSeg+~(B)~\cite{li2022language}                                              & \cmark & \cmark   & \xmark                    & \xmark                    & \cmark                    & \cmark                   & \xmark                                                                  & -                                        & -                                        & 18.0          & 3.8                                     & $\star$                                        & 46.5          & 7.8           & $\star$             & $\star$             & -             & $\star$             & $\star$             & -                                        & -                                        & $\star$             & -              \\
ZegFormer~(B)~\cite{hendricks2021decoupling}                                          & \cmark & \cmark   & \xmark                    & \xmark                    & \cmark                    & \cmark                   & \xmark                                                                  & -                                        & -                                        & $\star$             & \textcolor[rgb]{0.855,0.835,0.965}{8.1} & \textcolor[rgb]{0.855,0.835,0.965}{80.7} & $\star$             & $\star$             & $\star$             & $\star$             & -             & $\star$             & $\star$             & -                                        & -                                        & $\star$             & -              \\
OpenSeg~(B)~\cite{gu2021open}                                            & \cmark & \xmark   & \cmark                    & \xmark                    & \cmark                    & \cmark                   & \cmark                                                                  & -                                        & -                                        & 21.1          & 6.3                                     & 70.3                                     & 45.9          & 9.0           & $\star$             & $\star$             & -             & $\star$             & $\star$             & -                                        & -                                        & $\star$             & -              \\
OpenSeg~(B)~\cite{gu2021open}                                            & \cmark & \xmark   & \cmark                    & \cmark                    & \cmark                    & \cmark                   & \cmark                                                                  & -                                        & -                                        & 26.4          & 8.1                                     & 70.2                                     & 44.8          & 11.5          & $\star$             & $\star$             & -             & $\star$             & $\star$             & -                                        & -                                        & $\star$             & -              \\
MaskCLIP~(L)~\cite{ding2022open}                                           & \cmark & \cmark   & \xmark                    & \xmark                    & \cmark                    & \cmark                   & \xmark                                                                  & 15.1                                     & 6.0                                      & 23.7          & 8.2                                     & $\star$                                        & 45.9          & 10.0          & $\star$             & $\star$             & $\star$             & $\star$             & $\star$             & $\star$                                        & $\star$                                        & $\star$             & $\star$              \\ 
\hline
X-Decoder-Seg~(B)                                      & \cmark & \cmark   & \xmark                    & \xmark                    & \xmark                    & \xmark                   & \xmark                                                                  & 15.3                                     & 8.3                                      & 19.5          & 2.9                                     & 95.7                                     & 63.5          & 13.3          & 33.0          & 41.6          & 32.5          & 22.4          & 47.3          & 22.8                                     & 35.2                                     & 44.1          & 14.1           \\
X-Decoder-Seg$^+$~(B)                                  & \cmark & \cmark   & \cmark                    & \xmark                    & \xmark                    & \xmark                   & \xmark                                                                  & 16.9                                     & 9.5                                      & 23.8          & 4.6                                     & 97.8                                     & 64.7          & 12.1          & 32.2          & 35.1          & 33.8          & 18.5          & 47.6          & \textbf{25.9}                            & 36.9                                     & 42.7          & 16.6           \\ 
\hline
\rowcolor[rgb]{0.937,0.937,0.937} X-Decoder (T)        & \cmark & \cmark   & \cmark                    & \cmark                    & \xmark                    & \xmark                   & \xmark                                                                  & 18.8                                     & 9.8                                      & 25.0          & 6.4                                     & 96.2                                     & 62.9          & 12.3          & 34.5          & 37.8          & 30.7          & 21.7          & 47.3          & 16.0                                     & 37.2                                     & 42.4          & 16.4           \\
\rowcolor[rgb]{0.937,0.937,0.937} X-Decoder (B)        & \cmark & \cmark   & \cmark                    & \cmark                    & \xmark                    & \xmark                   & \xmark                                                                  & 21.1                                     & 11.7                                     & 27.2          & 8.2                                     & \textbf{97.9}                            & \textbf{65.1} & 14.7          & 39.6          & 40.3          & 35.4          & 24.8          & 50.8          & 22.3                                     & \textbf{39.5}                            & 45.1          & 17.1           \\
\rowcolor[rgb]{0.937,0.937,0.937} X-Decoder (L)        & \cmark & \cmark   & \cmark                    & \cmark                    & \xmark                    & \xmark                   & \xmark                                                                  & \textbf{21.8}                            & \textbf{13.1}                            & \textbf{29.6} & \textbf{9.2}                                     & 97.7                                     & 64.0          & \textbf{16.1}          & \textbf{43.0} & \textbf{49.5} & \textbf{39.5} & \textbf{29.7} & \textbf{52.0} & 24.9                                     & 38.1                                     & \textbf{47.2} & \textbf{17.8}  \\
\bottomrule
\end{tabular}
}
\vspace{3pt}
\caption{\textbf{One suite of model weights} for open-vocabulary image segmentation. Note: ``ITP'' means image-text pairs. ``Fix'' indicates whether contains fixed text/image encoder. ``EM" means whether the model has extra modules that are designed for open-vocabulary settings (e.g. Adaptor, class agnostic proposal, and etc.). ``Pseudo'' means whether the method uses an extra step to extract pseudo label image-text pairs. ``gray" color means a fully supervised approach. ``light purple" color means a semi-supervised learning approach. ``FL-in21k" means the backbone is pretained with in21k data using a FocalNet backbone. For COCO, different methods use different supervisions of mask~(m), class label~(cls) and caption~(cap). ``$\star$ and -" follows Table~\ref{tab:task_transfer}}
\label{tab:zero-shot-transfer}
\end{table*}
\vspace{-2pt}
\subsection{Task-Specific Transfer}
\label{exp:task_specific}
Without any architecture change except adding a head for VQA, we directly finetune \ourmodel{} to demonstrate its task transfer capability. Table~\ref{tab:task_transfer} presents the comparisons with previous specialized and generalized models.

\noindent\textbf{Comparison with segmentation models}. We list the most recent models for individual tasks, including Mask2Former~\cite{cheng2022masked}, Panoptic SegFormer~\cite{li2022panoptic}, KMaX-DeepLab~\cite{yu2022k} for generic segmentation, and LAVT~\cite{yang2022lavt} for referring segmentation. Notably, our 25 epoch finetuned \ourmodel{}~(L) \textit{\textbf{establishes a new SoTA on ADE20k dataset}} that outperforms the current SoTA KMaX-DeepLab~(L) on ADE Panoptic Segmentation (our model trained with 1024 resolution achieves 51.0 PQ), as well as Instance Segmentation SoTA, Mask2Former-L. On COCO, our model attains comparable performance to Mask2Former and kMaX-DeepLab. There are three reasons to explain minor inferiority. First, we do not use deformable attention in \ourmodel{}, which typically benefits supervised settings but hurts open-vocabulary performance. Second, we use the language-image pretrained model as the backbone, which can understand richer semantics but lags behind the supervised model for classification tasks~\cite{li2022elevater}. Third, we use 100 latent queries for segmentation, which is half of that in Mask2Former~(L). Finally, we compare with LAVT~\cite{yang2022lavt} on COCO G-ref. It is worth pointing out that with lightweight finetuning, our tiny model already outperforms LAVT-Base (\textbf{61.9} \textit{v.s.} \textbf{61.2}). Further increasing the model size can bring additional gains by \textbf{2.6} and \textbf{2.7} points respectively, which helps to \textit{\textbf{set a new record on this benchmark}}.

\noindent
\textbf{Comparison with VL models.} We compare with a set of VL models on image-text retrieval, image captioning and VQA in Table~\ref{tab:task_transfer}. \ourmodel~achieves competitive performance 
across the board. Specifically, \ourmodel{} outperforms strong baseline UNITER~\cite{chen2020uniter} and rivals VinVL~\cite{zhang2021vinvl} on COCO retrieval, and even beats all the methods on Flickr30k~\cite{plummer2015flickr30k}. Unlike all these works, the image and text encoders are fully decoupled in \ourmodel{}, which leads to a much faster inference speed. On captioning and VQA, our models also demonstrate superior performance to their counterparts. For example, it outperforms VinVL by \textbf{1.3} and \textbf{1.7} on CIDEr and BLEU, respectively. Note that most of these works use sophisticatedly designed training objectives, such as masked data modeling, image-text matching and hard-negative mining~\cite{li2021align,wang2021ufo,fiber2022}. In contrast, \ourmodel{} is pretrained with image-text contrastive and image captioning, along with the segmentation losses. The simplicity and effectiveness imply a great potential of using \ourmodel{} as a general pretraining paradigm for VL.

\noindent
\textbf{Comparison with generalist models}. We further compare with prior arts that explore general-purpose vision models. Limited works report the generic segmentation performance. Our model outperforms UViM~\cite{kolesnikov2022uvim} and Pix2Seq~v2~\cite{chen2022unified} significantly on COCO panoptic (\textbf{56.7} \textit{v.s.} \textbf{45.8}) and instance segmentation (\textbf{46.7 \textit{v.s.} 38.2}), respectively. With the same amount of segmentation data, these margins strongly justify our model design, \ie, unifying functionality \emph{without} any tweaks for individual tasks. When compared with GLIPv2~\cite{zhang2022glipv2}, our model achieves comparable performance. Note that GLIPv2 uses over 10M pretraining data, including around 2M with box supervision. Despite the huge gap in pretraining data, \ourmodel{} outperforms GLIPv2 on both captioning and VQA. Furthermore, \ourmodel{} also beats other general-purpose models like UniT~\cite{hu2021unit}, GPV~\cite{gupta2022towards}, UniTAB~\cite{yang2022unitab} and Unified-IO~\cite{lu2022unified}.

\noindent\textbf{Efficient Finetuning}. Finally, we study whether our pretrained \ourmodel{} can be finetuned for segmentation with a low cost. In Table~\ref{tab:efficient-finetuning}, we show that we can simply finetune the class embedding layer, mask embedding layer or the whole decoder to reach a decent segmentation performance and surpass the fully finetuned tiny SoTA models like kMaX-DeepLab~\cite{yu2022k}. These results imply an efficient way of using our pretrained \ourmodel{} models.

\begin{figure*}[!t]
    \centering
    \includegraphics[width=.99\textwidth]{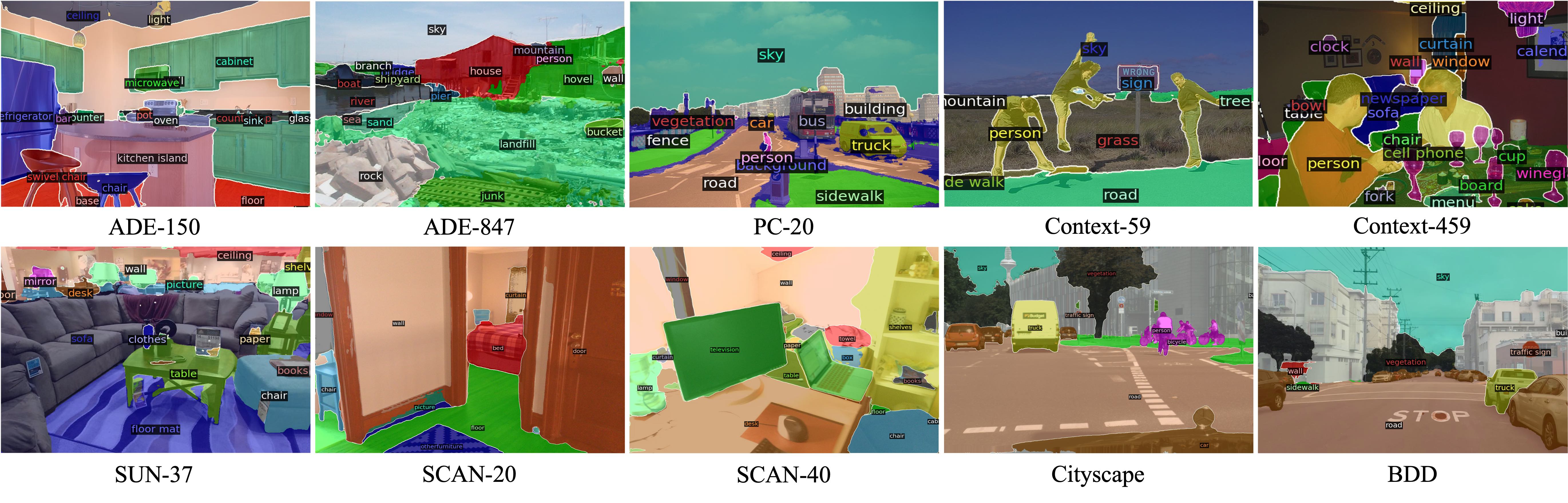}
    \vspace{-3pt}
    \caption{Visualization of zero-shot semantic segmentation on 10 settings of 7 datasets.}
    \vspace{2pt}
    \label{fig:segmentation}
\end{figure*}

\begin{table}[]
\tablestyle{1.8pt}{0.95}
\centering
\resizebox{1.0\linewidth}{!}{
\begin{tabular}{c|ccccc|cccccc} 
\toprule
\multirow{2}{*}{Method}                                                                               & \multirow{2}{*}{C.E.}                 & \multirow{2}{*}{M.E}                  & \multirow{2}{*}{Q.E}                  & \multirow{2}{*}{Dec.}                 & \multirow{2}{*}{\#Param} & \multicolumn{3}{c}{ADE}                                                                                                              & \multicolumn{3}{c}{Cityscapes}                                                                                                  \\
                                                                                                      &                                       &                                       &                                       &                                       &                          & PQ                                          & mAP                                         & mIoU                                     & PQ                                       & mAP                                      & mIoU                                      \\ 
\hline
Mask2Former (T)~\cite{cheng2022masked}                                                                                       & -                                     & -                                     & -                                     & -                                     & -                        & 39.7                                        & 26.4                                        & 47.7                                     & \textcolor[rgb]{0.133,0.161,0.184}{63.9} & 39.1                                     & \textcolor[rgb]{0.133,0.161,0.184}{80.5}  \\
Pano/SegFormer (T)~\cite{li2022panoptic,xie2021segformer}                                                                                   & -                                     & -                                     & -                                     & -                                     & -                        & 36.4                                        & $\star$                                     & 46.5                                     & $\star$                                  & $\star$                                  & $\star$                                   \\
kMaX-DeepLab (T)~\cite{yu2022k}                                                                                     & -                                     & -                                     & -                                     & -                                     & -                        & 41.5                                        & $\star$                                     & 45.0                                     & 64.3                                     & 38.5                                     & 79.7                                      \\
Mask2Former (S)~\cite{cheng2022masked}                                                                                      & -                                     & -                                     & -                                     & -                                     & -                        & $\star$                                     & $\star$                                     & \textcolor[rgb]{0.133,0.161,0.184}{51.3} & \textcolor[rgb]{0.133,0.161,0.184}{64.8} & 40.7                                     & 81.8                                      \\
\textcolor[rgb]{0.694,0.694,0.702}{Mask2Former (B)~\cite{cheng2022masked}}                                                   & \textcolor[rgb]{0.694,0.694,0.702}{-} & \textcolor[rgb]{0.694,0.694,0.702}{-} & \textcolor[rgb]{0.694,0.694,0.702}{-} & \textcolor[rgb]{0.694,0.694,0.702}{-} & -                        & \textcolor[rgb]{0.694,0.694,0.702}{$\star$} & \textcolor[rgb]{0.694,0.694,0.702}{$\star$} & \textcolor[rgb]{0.694,0.694,0.702}{53.9} & \textcolor[rgb]{0.694,0.694,0.702}{66.1} & \textcolor[rgb]{0.694,0.694,0.702}{42.8} & \textcolor[rgb]{0.694,0.694,0.702}{82.7}  \\
\textcolor[rgb]{0.694,0.694,0.702}{Mask2Former (L)~\cite{cheng2022masked}}                             & \textcolor[rgb]{0.694,0.694,0.702}{-} & \textcolor[rgb]{0.694,0.694,0.702}{-} & \textcolor[rgb]{0.694,0.694,0.702}{-} & \textcolor[rgb]{0.694,0.694,0.702}{-} & -                        & \textcolor[rgb]{0.694,0.694,0.702}{48.1}    & \textcolor[rgb]{0.694,0.694,0.702}{34.9}    & \textcolor[rgb]{0.694,0.694,0.702}{56.1} & \textcolor[rgb]{0.694,0.694,0.702}{66.6} & \textcolor[rgb]{0.694,0.694,0.702}{43.6} & \textcolor[rgb]{0.694,0.694,0.702}{82.9}  \\
\rowcolor[rgb]{0.937,0.937,0.937} {\cellcolor[rgb]{0.937,0.937,0.937}}                                & \cmark                                     & \xmark                                     & \xmark                                     & \xmark                                     & 0.26M                    & \uline{44.3}                                & \uline{33.2}                                & \uline{54.6}                             & \uline{65.1}                             & \uline{41.4}                             & \uline{81.7}                              \\
\rowcolor[rgb]{0.937,0.937,0.937} {\cellcolor[rgb]{0.937,0.937,0.937}}                                & \cmark                                     & \cmark                                     & \xmark                                     & \xmark                                     & 1.05M                    & 43.9                                        & 33.2                                        & 53.9                                     & 64.8                                     & 41.2                                     & 81.2                                      \\
\rowcolor[rgb]{0.937,0.937,0.937} {\cellcolor[rgb]{0.937,0.937,0.937}}                                & \cmark                                     & \cmark                                     & \cmark                                     & \xmark                                     & 1.15M                    & 44.0                                        & 32.8                                        & 54.0                                     & 64.6                                     & 41.1                                     & 81.5                                      \\
\rowcolor[rgb]{0.937,0.937,0.937} \multirow{-4}{*}{{\cellcolor[rgb]{0.937,0.937,0.937}}X-Decoder (L)} & \cmark                                     & \cmark                                     & \cmark                                     & \cmark                                     & 38.3M                    & \textbf{47.0}                               & \textbf{35.1}                               & \textbf{56.0}                            & \textbf{65.6}                            & \textbf{42.2}                            & \textbf{81.7}                             \\
\bottomrule
\end{tabular}
}
\vspace{3pt}
\caption{Performance with different efficient finetuning strategies for \ourmodel{} large, and comparisons with fully-finetuned models.}
\label{tab:efficient-finetuning}
\end{table}

\subsection{Zero-Shot Transfer}
\label{exp:zero_shot_sec}

Without any change in model weights, \ourmodel{} can be directly applied to various segmentation tasks and datasets after pretraining. In Table~\ref{tab:zero-shot-transfer}, we evaluate our model in a zero-shot manner on seven commonly used segmentation datasets in 10 different settings from diverse domains, including common indoor (\eg, ADE20K~\cite{zhou2017scene} and Pascal~\cite{everingham2011pascal}), outdoor (\eg, Cityscapes~\cite{cordts2015cityscapes}) and self-driving scenarios (\eg, BDD~\cite{yu2018bdd100k}). We report PQ, mAP and mIoU for panoptic, instance and semantic segmentation respectively. And we visualize the predicted open-vocabulary segmentation result on each dataset in Fig.~\ref{fig:segmentation}.

\begin{table*}[t]
\begin{minipage}{0.49\linewidth}
\tablestyle{1.6pt}{1.0} 
\centering
\resizebox{\linewidth}{!}{
\begin{tabular}{l|lll|lll|lll|l} 
\toprule
\multirow{2}{*}{Model} & \multicolumn{3}{l|}{COCO}                     & \multicolumn{3}{l|}{ADE}                     & \multicolumn{3}{l|}{COCO-Karparthy}                & g-Ref            \\
                      & PQ            & mAP           & mIoU          & PQ            & mAP          & mIoU          & IR@1           & IR@1           & CIDEr         & cIoU           \\ 
\hline
\cellcolor[HTML]{EFEFEF}X-Decoder              & \cellcolor[HTML]{EFEFEF}\textbf{51.4} & \cellcolor[HTML]{EFEFEF}\textbf{40.5} & \cellcolor[HTML]{EFEFEF}\textbf{62.8} & \cellcolor[HTML]{EFEFEF} \uline{14.7}          & \cellcolor[HTML]{EFEFEF}\textbf{9.6} & \cellcolor[HTML]{EFEFEF}\textbf{23.4} & \cellcolor[HTML]{EFEFEF}30.7          & \cellcolor[HTML]{EFEFEF}48.5          & \cellcolor[HTML]{EFEFEF}\textbf{82.0} & \cellcolor[HTML]{EFEFEF}\textbf{59.7}  \\
* text: [yny]          & 51.4          & 39.8          & 61.7          & 14.7          & 9.4          & \uline{22.2}          & 29.9          & \uline{46.9}          & \uline{78.6}          & 57.7           \\
* text: [nyy]          & 51.4          & \uline{38.6}          & \uline{61.7}          & 15.2          & \uline{9.4}          & 23.1          & 30.3          & 47.5          & 78.9          & 59.4           \\
* latent: [yyn]          & \uline{50.9}          & 39.6          & 62.0          & \textbf{15.5} & 9.4          & 22.8          & \uline{29.8}          & 47.6          & 81.1          & \uline{57.6}           \\
\bottomrule
\end{tabular}
}
\caption{\textbf{Ablation of query interaction} in \ourmodel{}. [x,x,x] denotes whether attend [object latent query, image latent query, text query]}
\label{table:ablate_self_attention}
\end{minipage}
\quad
\begin{minipage}{0.49\linewidth}
\tablestyle{1.6pt}{1.0} 
\resizebox{\linewidth}{!}{
\begin{tabular}{l|lll|lll|lll|l}
\toprule
                        & \multicolumn{3}{l|}{COCO}                                       & \multicolumn{3}{l|}{ADE}                                       & \multicolumn{3}{l|}{COCO-Karparthy}                                  & g-Ref                 \\
\multirow{-2}{*}{Model} & PQ                  & mAP                 & mIoU                & PQ                  & mAP                & mIoU                & IR@1                 & TR@1                 & CIDEr               & cIoU                \\ \hline
\rowcolor[HTML]{EFEFEF} 
\rowcolor[rgb]{0.937,0.937,0.937} * bs 1024 & 50.9                   & 39.5                    & 62.4                      & 15.2                   & 10.0                    & 24.6                      & 30.6                             & 48.1                             & 85.0                              & 58.0                       \\

* bs 768             & 51.0                & 39.5                & 62.4                & 15.4 & 10.0                & 24.2 &    29.0              & 46.8                 & 78.6 & 58.8                \\
* bs 512            & 50.7 & 39.3                & 62.0              & 14.9                & 9.7                & 24.3                & \textcolor[rgb]{0,0.502,0} {27.4} & \textcolor[rgb]{0,0.502,0}{43.8}                &      \textcolor[rgb]{0,0.502,0}{   76.1 }        & 58.6 \\
\bottomrule[1.0pt]
\end{tabular}
}
\vspace{1pt}
\caption{\textbf{Ablation of VL batch size.} We mark the significant drop metrics in green.}
\label{tab:ablate_vl_batchsize}
\end{minipage}
\end{table*}

\begin{table*}[t]
\begin{minipage}{0.49\linewidth}
\tablestyle{1.6pt}{1.0} 
\resizebox{\linewidth}{!}{
\begin{tabular}{l|lll|lll|lll|l}
\toprule
                        & \multicolumn{3}{l|}{COCO}                                       & \multicolumn{3}{l|}{ADE}                                       & \multicolumn{3}{l|}{COCO-Karparthy}                                  & g-Ref                 \\
\multirow{-2}{*}{Model} & PQ                  & mAP                 & mIoU                & PQ                  & mAP                & mIoU                & IR@1                 & TR@1                 & CIDEr               & cIoU                \\ \hline
\rowcolor[HTML]{EFEFEF} 
X-Decoder               & \textbf{51.4}                & \textbf{40.5}                & 62.8                & 14.7                & \textbf{9.6}               & \textbf{23.4}                & \textbf{30.7}                & \textbf{48.5}                & \textbf{82.0}                & \textbf{59.7}                \\
- Retrieval             & 51.4                & 40.4                & 62.6                & {\ul \textit{14.0}} & 9.2                & {\ul \textit{21.8}} & n/a                 & n/a                 & {\ul \textit{78.8}} & 59.2                \\
- Captioning            & {\ul \textit{51.1}} & 40.4                & \textbf{63.2}              & 15.0                & 9.6                & 23.2                & {\ul \textit{29.9}} & 48.1                & n/a                 & {\ul \textit{57.7}} \\
- Referring             & {\ul \textit{51.1}} & {\ul \textit{39.7}} & {\ul \textit{62.3}} & \textbf{15.2}               & {\ul \textit{8.9}} & 22.6                & 30.0                & {\ul \textit{47.6}} & {\ul \textit{78.8}} & n/a                 \\ 
\bottomrule[1.0pt]
\end{tabular}
}
\vspace{1pt}
\caption{\textbf{Ablation of pretraining tasks} by removing one at a time. We bold the best entry and underline the worst entry in each column.}
\label{tab:ablate_pretraining_task}
\end{minipage}
\quad
\begin{minipage}{0.49\linewidth}
\tablestyle{1.6pt}{1.0} 
\centering
\resizebox{\linewidth}{!}{
\begin{tabular}{l|ccc|ccc|ccc|c} 
\toprule
\multirow{2}{*}{Model}                          & \multicolumn{3}{l|}{COCO}                                                    & \multicolumn{3}{l|}{ADE}                                                     & \multicolumn{3}{l|}{COCO-Karparthy}                                                                     & \multicolumn{1}{l}{g-Ref}  \\
                                                & \multicolumn{1}{l}{PQ} & \multicolumn{1}{l}{mAP} & \multicolumn{1}{l|}{mIoU} & \multicolumn{1}{l}{PQ} & \multicolumn{1}{l}{mAP} & \multicolumn{1}{l|}{mIoU} & \multicolumn{1}{l}{IR@1}         & \multicolumn{1}{l}{TR@1}         & \multicolumn{1}{l|}{CIDEr}        & \multicolumn{1}{l}{cIoU}   \\ 
\hline
\rowcolor[rgb]{0.937,0.937,0.937} Full Datasets & 50.9                   & 39.5                    & 62.4                      & 15.2                   & 10.0                    & 24.6                      & 30.6                             & 48.1                             & 85.0                              & 58.0                       \\
- coco                                          & 50.9                   & 39.9                    & 62.2                      & 15.3                   & 9.8                     & 24.4                      & 27.4                             & 38.2                             & 32.6                              & 59.4                       \\
- cc3m                                          & 51.2                   & 39.7                    & 62.6                      & 15.5                   & 10.1                    & 24.6                      & 31.0                             & 50.0                             & 81.2                              & 58.3                       \\
- vg                                            & 51.1                   & 39.8                    & 62.4                      & \textcolor[rgb]{0,0.502,0}{14.6}                   & \textcolor[rgb]{0,0.502,0}{9.7}                     & \textcolor[rgb]{0,0.502,0}{23.8}                      & \textcolor[rgb]{0.502,0,0}{36.1} & \textcolor[rgb]{0.502,0,0}{56.1} & \textcolor[rgb]{0.502,0,0}{107.1} & 58.3                       \\
- sbu                                           & 51.1                   & 39.8                    & 62.4                      & 15.3                   & 9.5                     & 24.6                      & 30.3                             & 48.3                             & 81.2                              & 58.3                       \\
\bottomrule
\end{tabular}
}
\caption{\textbf{Ablation of VL datasets} in \ourmodel{}. A single VL dataset is removed in each row. And we mark the metrics that significantly drop/increase in green/red. }
\label{table:ablate_vl_datasets}
\end{minipage}
\end{table*}

\noindent
\textbf{Comparison with baselines}. We build two \ourmodel{} variants: (1) \ourmodel{}-Seg, which is only trained with COCO panoptic segmentation using a text encoder for class names;
and (2) \ourmodel{}-Seg$^+$, where we take the heuristic way to extract noun phrases from COCO captions and use them as extra supervision on top of the matched decoder outputs. First, \ourmodel{}-Seg shows clear advantages on open-vocabulary segmentation over MSeg~\cite{lambert2020mseg}, that manually conducts label mapping across different datasets. Second, the extra supervision from COCO captions improves model performance on 9 out of 15 metrics, which indicates the benefit of joint learning with image-level supervision. Third, when pretraining with the full \ourmodel{}, the performance is significantly boosted. Notably, the mIoU metric is improved by \textbf{7.4}, \textbf{3.4} and \textbf{2.6} on SUN, ADE-150 and PC-459, respectively. 

\noindent
\textbf{Comparison with state-of-the-art}. We further compare with the most advanced methods for open-vocabulary image segmentation in Table~\ref{tab:zero-shot-transfer}. Clearly, our models achieve the best results across all datasets. Among the base-sized models, \ourmodel{}~(B) outperforms OpenSeg~(B)~\cite{ghiasi2021open} on two challenging datasets, ADE-150 and PC-459 for semantic segmentation. Scaling \ourmodel~to large size 
 further improves mIoU by \textbf{2.4} and \textbf{1.4} on these two datasets. 
Among prior arts, MaskCLIP~\cite{ding2022open} is the first proposed for open-vocabulary panoptic segmentation by combining Mask2Former with CLIP models. 
With COCO caption supervisions, our simple baseline \ourmodel{}-Seg$^+$ already performs comparably. The full version of our tiny model \ourmodel{}~(T) surpasses MaskCLIP across the board except A-847. We note that these comparisons are not strictly fair in terms of supervision, settings and models used. However, these results demonstrate the effectiveness of our \ourmodel{} to learn from the different granularity of supervisions \emph{end-to-end} for open-vocabulary segmentation, which leads to \textit{\textbf{new SoTA on 10 settings of 7 datasets across three segmentation tasks}}.

\begin{figure*}[!t]
    \centering
    \vspace{8pt}
    \includegraphics[width=.99\textwidth]{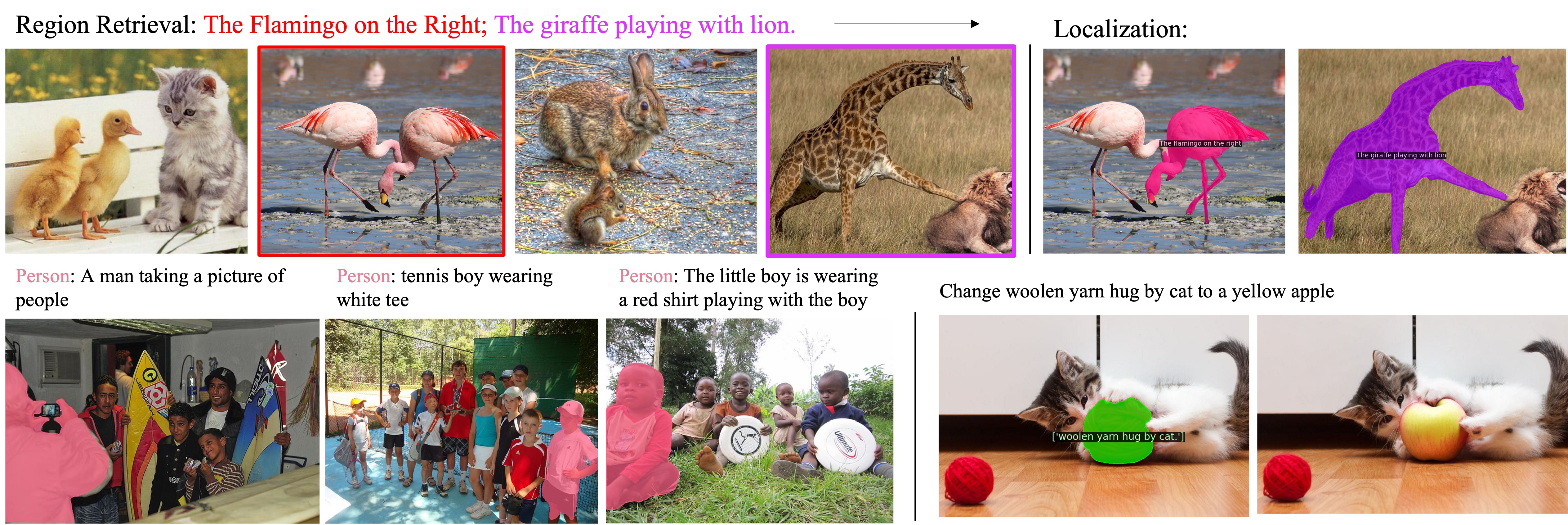}
    \caption{Top: Region Retrieval. Bottom: Referring Captioning (Left), Referring Image Editing (right). Please refer to more results in Appendix.}
    \label{fig:visualization}
\end{figure*}

\subsection{Model Inspection}
\label{exp:query_promt}

\noindent\textbf{Pretraining Tasks}. By default, we exploit four pretraining tasks including generic and referring segmentation, captioning and retrieval. In Table~\ref{tab:ablate_pretraining_task}, we keep the generic segmentation while ablating the importance of the other pretraining tasks. Accordingly, we have the following observations:
\begin{itemize}[leftmargin=*,topsep=2pt]
\setlength\itemsep{-2pt}

\item \textbf{Image-text retrieval can help open-vocabulary segmentation.} On ADE, the mIoU drops from \textbf{23.4} to \textbf{21.8}, and PQ drops \textbf{0.7} without image-text retrieval. Since we share the same semantic space for both tasks, a good visual-semantic alignment learned from the retrieval task can directly benefit the recognition of novel concepts.

\item\textbf{Image captioning helps referring segmentation and vice versa.} We observe a drop of \textbf{2.0} pts on COCO g-Ref without captioning task, and a \textbf{3.2} pts drop of CIDEr from removing referring task. The two tasks share the same text encoder for text queries. Joint training, therefore, improves the understanding of text inputs.

\item\textbf{Image captioning and retrieval can mutually benefit each other.} When removing captioning during pretraining, the image retrieval R@1 drops by \textbf{0.8}, and the captioning CIDEr drops significantly by \textbf{3.2} pts from removing retrieval task. Our \ourmodel{} promotes harmony of generative and contrastive learning.      
\end{itemize}

\noindent The above observations verify that the unified design of \ourmodel{} can prompt the synergy of different tasks.

\smallskip
\noindent\textbf{Query Interactions}. The interaction among tasks is highly dependent on the interaction between latent and text queries. We have described how the queries interact with each other by default in Fig.~\ref{fig:selfattention_mechanism}. Here, we investigate how our model behaves with different interactions. In Table~\ref{table:ablate_self_attention}, we show the performance across tasks with ablated versions and have the following takeaways:
\begin{itemize}[leftmargin=*, topsep=2pt]
\setlength\itemsep{-2pt}

\item \textbf{Image captioning requires both fine-grained and global image information}. Comparing the first with the second and third row in the table, we find the CIDEr score significantly drops if we cut off the information flow from the global latent query or other latent queries to text queries (\textbf{82.0} $\rightarrow$ \textbf{78.6} and \textbf{78.9}, respectively).

\item \textbf{Language-condition is important for referring segmentation}. In the last row, we turn off the interaction from text queries to latent queries. This significantly hurts referring segmentation (\textbf{59.7}$\rightarrow$ \textbf{57.6}). On the one hand, this indicates that we can convert generic segmentation to referring segmentation using post-hoc matching with referring texts. On the other hand, sending the text phrase as input to \ourmodel{} is essential to modulate our model to specifically decode the targets.
\end{itemize}

\noindent\textbf{VL Batch Size \& Dataset} The default batch size of VL task is $1024$, here we explore the gradual decreasing of VL batch size. In addition, each VL dataset is removed individually to investigate the pre-trained performance on different tasks.

\begin{itemize}[leftmargin=*, topsep=2pt]
\item \textbf{Decreasing VL batch size hurts VL tasks and open-vocab Segmentation performance}. As shown in Table.~\ref{tab:ablate_vl_batchsize}, decreasing the VL task batch size from $1024$ to $256$ significantly hurts the retrieval and captioning tasks' performance, where ir@1, tr@1, CIDEr decrease by 3.2, 4.3, and 8.9 points respectively. Further, the open-vocabulary performance also drops 0.3 points on each metric.

\item \textbf{VG dataset hurts pretraining VL tasks performance but improves open-vocab segmentation}. As shown in Table~\ref{table:ablate_vl_datasets}, removing the visual genome from the pretraining VL dataset significantly improves captioning task with 22.1 points during pretraining, but only 0.2 points after finetuning. Moreover, open-vocabulary semantic segmentation drops around 0.8 points.

\end{itemize}

\subsection{Task Composition}
\label{exp:task_composite}

\ourmodel{} has the unique benefit of task interaction, thanks to the sophisticated architecture design on latent and text queries as well as the decoder architecture. It enables joint task inference and iterative task inference with a single set of weights. In Fig.~\ref{fig:visualization}, we show our model can perform region-based retrieval and referring based captioning without any architecture/weight change. For example, given a set of animal images (row 1, Fig.~\ref{fig:visualization}) and text query, our model first retrieves the correct image (flamingo and giraffe) and then grounds the query with pixel-level predictions. Further, our model can easily adapted to referring captioning by first localizing a given word and then modulating the predicted mask in the cross-attention layers. Lastly, we also integrate X-Deocder with diffusion model to do referring image editing demonstrated in the latter half of the second row in Fig.~\ref{fig:visualization}.
\vspace{-4pt}
\section{Conclusion}

We present \ourmodel{}, a model that seamlessly supports pixel-level and image-level vision-language understanding. With a simple and generalized design, \ourmodel{} can unite and support generic segmentation, referring segmentation and VL tasks effortlessly, achieving strong generalizability and competitive or even SoTA performance. We hope this work can shed a light on the design of the next-generation general-purpose vision system.

\paragraph{Acknowledgements.} We appreciated the constructive discussion with Haotian Zhang. This work was also supported in part by NSF CAREER IIS2150012, the Wisconsin Alumni Research Foundation, and the Institute of Information \& communications Technology Planning \& Evaluation (IITP) grant funded by the Korea government (MSIT) (No. 2022- 0-00871, Development of AI Autonomy and Knowledge Enhancement for AI Agent Collaboration).

{\small
\bibliographystyle{ieee_fullname}
\bibliography{egbib}
}

\clearpage
\appendix

\section{Experiment Settings}
\subsection{Pretraining}
In the main paper, all the pre-trained models are trained with 50 epochs of COCO data and roughly 45 epochs of 10 million image-text pairs. The batch size of COCO images and image text pairs are 32 and 1024 respectively. And 32 GPUs are used for pretraining. The AdamW optimizer is used in pretraining with the initial learning rate 1e-4. A step-wise scheduler is used to decay the learning rate by 0.1 on the fraction $[0.88889, 0.96296]$ of training steps.

\subsection{Finetuning}
\textbf{Image-Text Retrieval.} For both COCO and Flickr30k image-text retrieval, we finetune the models for 10 epochs using AdamW as the optimizer. We set the image resolution to 384 and the batch size to 2048. The learning rates are 3e-5 for the X-Decoder part and 3e-6 for the vision and language backbones. 

\textbf{Image Captioning.} Similar to image-text retrieval, we finetune the captioning models for 10 epochs using AdamW as the optimizer. We set the image resolution to 480 and the batch size to 256. The learning rates are 2e-5 for the X-Decoder part and 2e-6 for the vision and language backbones. We use beam search during caption generation with the beam size set to 5. We do not use CIDEr optimization for our captioning models.

\textbf{VQA.} For VQA, we add a new classification layer on the top of the model and finetune the models for 10 epochs using AdamW as the optimizer. We set the image resolution to 640 and the batch size to 256. The learning rates are 1e-4 for the X-Decoder part, 1e-5 for the vision and language backbones, and 1e-3 for the VQA classification layer.

\textbf{Generic Segmentation.}
For generic segmentation, we finetune the pretrained checkpoint with 24 epochs with start learning rate 1e-4. We decay the learning rate by factor 10 at epoch 21 and 23, respectively. The batch size of ADE20k is 64, and 32 for COCO. 

\textbf{Referring Segmentation.}
For referring segmentation, we also finetune the pretrained checkpoint with 24 epochs. However, as RefCOCO has been used in pretraining, thus the initial learning rate is 1e-5. It also decays twice at 21 and 23 epochs. We use a batch size of 64 during training. Further, in addition to the normal setting that multiple backbone and language encoder learning rates with 0.1, here we also multiply the transformer encoder learning rate by 0.1.

\section{Open-Vocab Segmentation Benchmark}
We propose an open vocabulary segmentation benchmark on 9 datasets with different evaluation metrics. The goal of this benchmark is to provide a comprehensive and standard evaluation protocol for open-vocabulary segmentation on different vocabulary sizes and image domains.

\begin{table}[h]
\footnotesize \setlength{\tabcolsep}{1.2pt}
\centering
\begin{tabular}{lc|ccc|cc} 
\toprule
\multirow{2}{*}{Dataset} & \multirow{2}{*}{Scene}& \multicolumn{3}{c|}{Annotation Format} & \multirow{2}{*}{\# Images} & \multirow{2}{*}{\# Classes}  \\
                      &   & Sem & Inst & Pano                      &                            &                              \\ 
\hline
ADE-150        & common          & \cmark   & \cmark    & \cmark                         & 2000                       & 150                          \\
ADE-847       & common           & \cmark   & \xmark    & \xmark                         & 2000                       & 847                          \\
Pascal Voc    & common           & \cmark   & \xmark    & \xmark                         & 1449                       & 20                           \\
Pascal Context-59  & common      & \cmark   & \xmark    & \xmark                         & 5105                       & 59                           \\
Pascal Context-459 & common      & \cmark   & \xmark    & \xmark                         & 5105                       & 459                          \\
SUN RGB-D     & in-door           & \cmark   & \xmark    & \xmark                         & 5050                       & 37                           \\
ScanNet-20    & in-door           & \cmark   & \xmark    & \cmark                         & 5436                       & 20                           \\
ScanNet-41     & in-door          & \cmark   & \xmark    & \xmark                         & 5436                       & 41                           \\
Cityscapes     & driving           & \cmark   & \cmark    & \cmark                         & 500                        & 19/8/19                      \\
BDD           & driving           & \cmark   & \xmark    & \cmark                         & 1000                       & 19//40                       \\
\bottomrule
\end{tabular}
\vspace{2pt}
\caption{Open-Vocabulary Segmentation Benchmark Statistics.}
\label{tab:open-vocab-benchmark}
\end{table}

\begin{figure}
    \vspace{-70pt}
    \includegraphics[width=.98\linewidth]{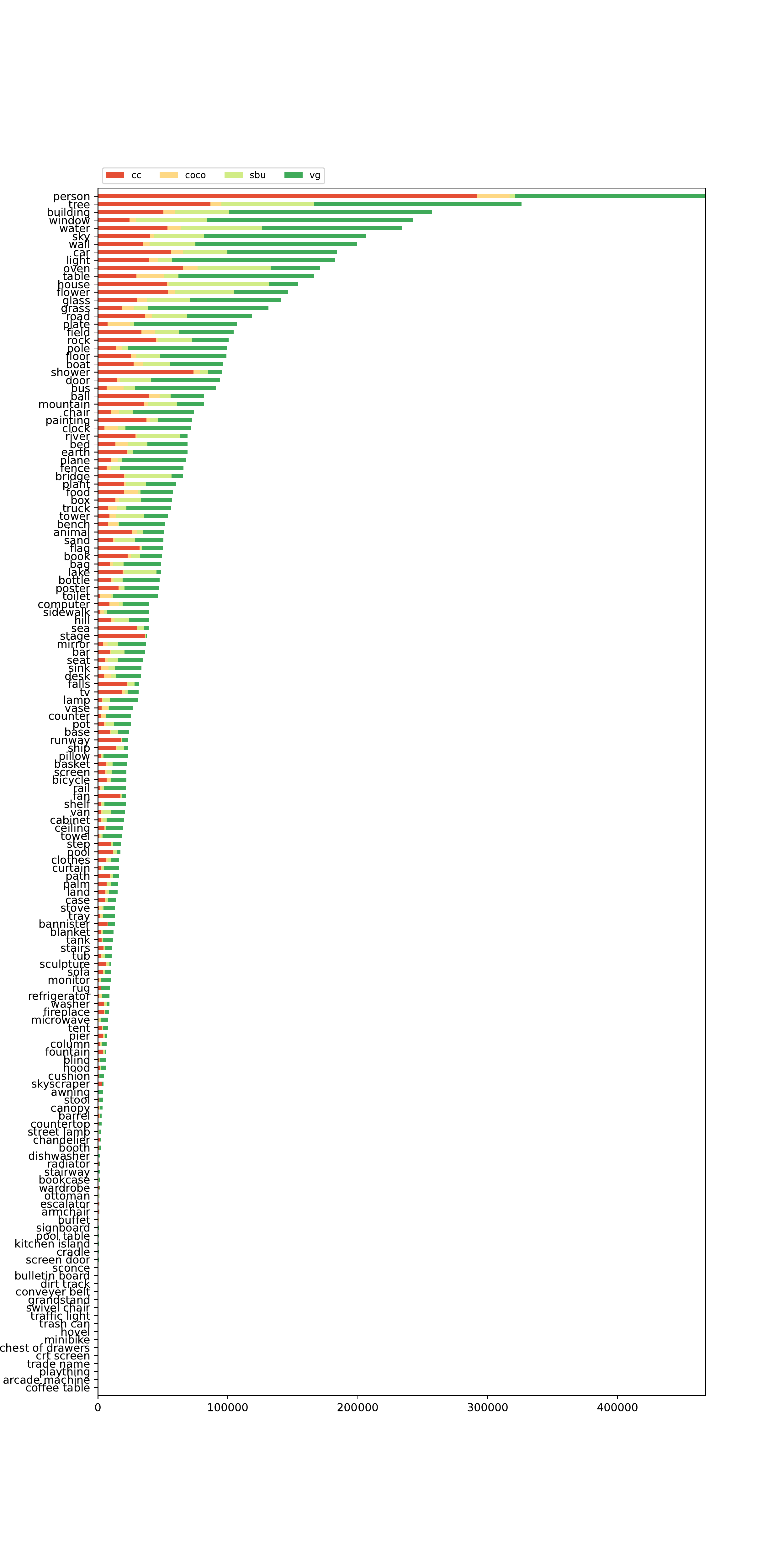}
    \vspace{-44pt}
    \caption{Image captions overlap with ADE20K-150}
    \label{fig:concept_ade}
\end{figure}

\begin{figure}
    \includegraphics[width=.98\linewidth]{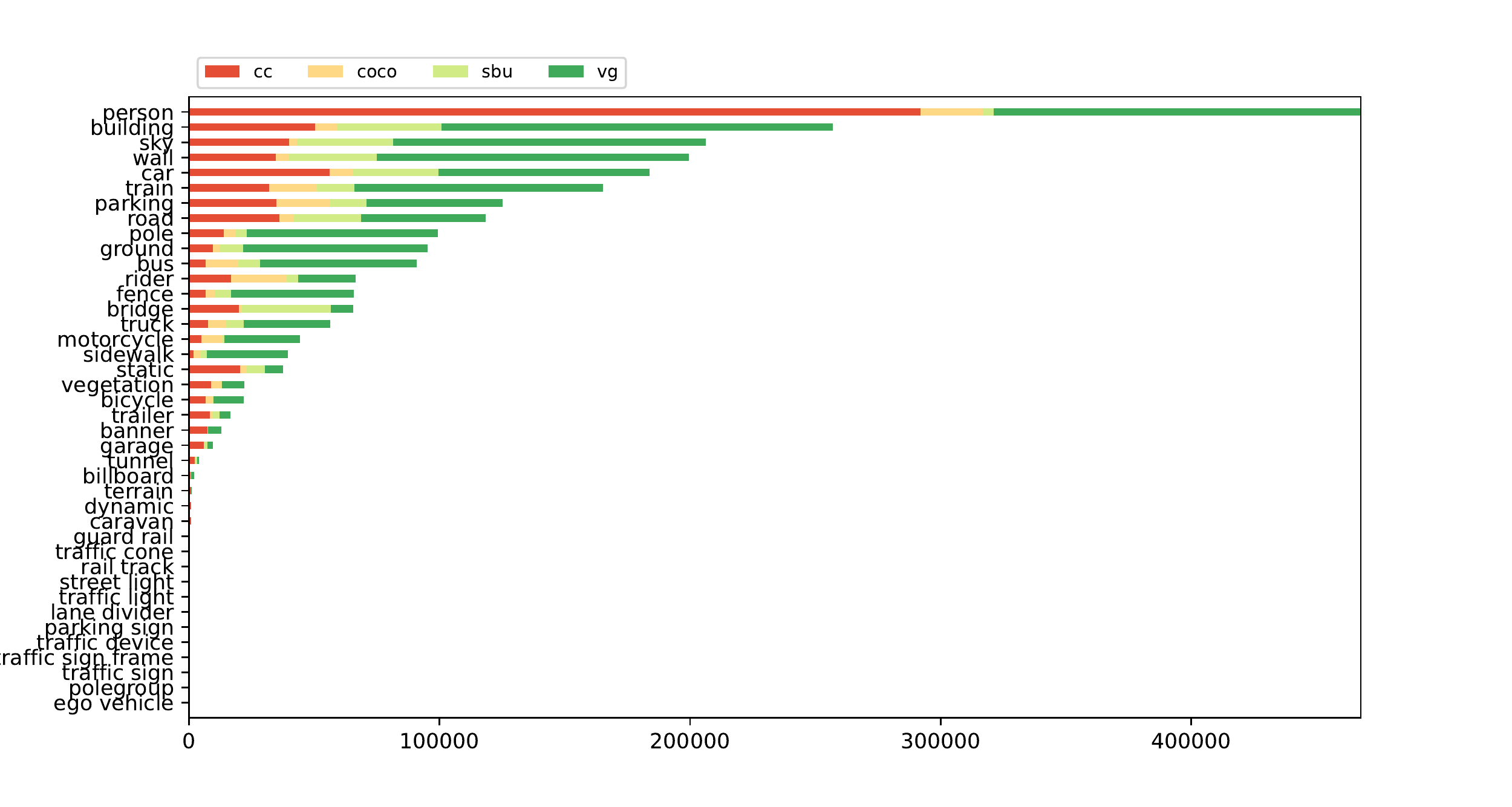}
    \vspace{-4pt}
    \caption{Image captions overlap with BDD-Panoptic}
    \label{fig:concept_bddpano}
\end{figure}

\begin{figure}
    \includegraphics[width=.98\linewidth]{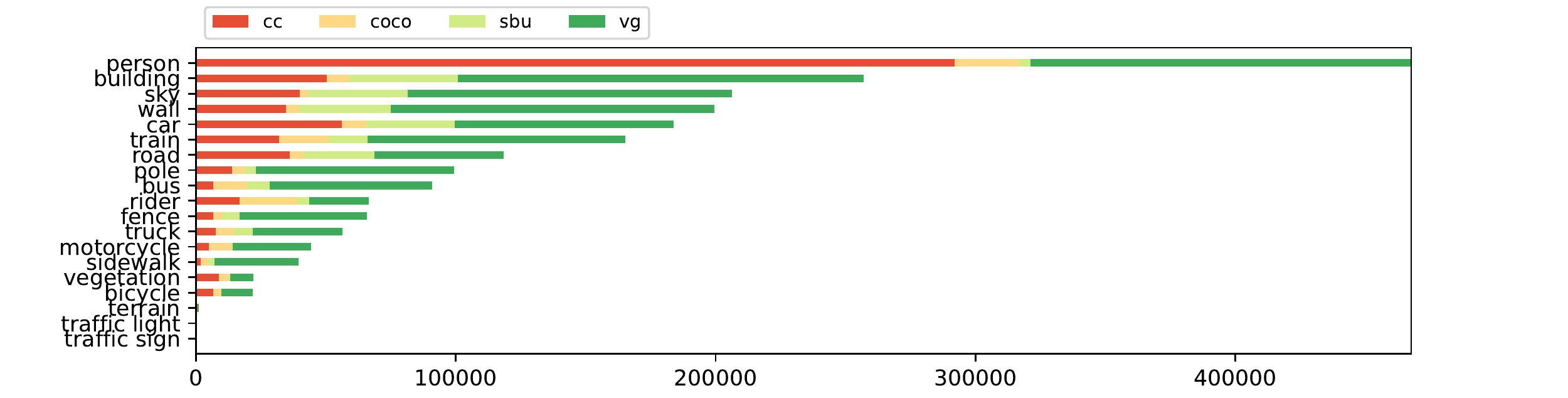}
    \caption{Image captions overlap with BDD-Semantic/Cityscapes}
    \label{fig:concept_bddsem}
\end{figure}

\begin{figure}
    \includegraphics[width=.98\linewidth]{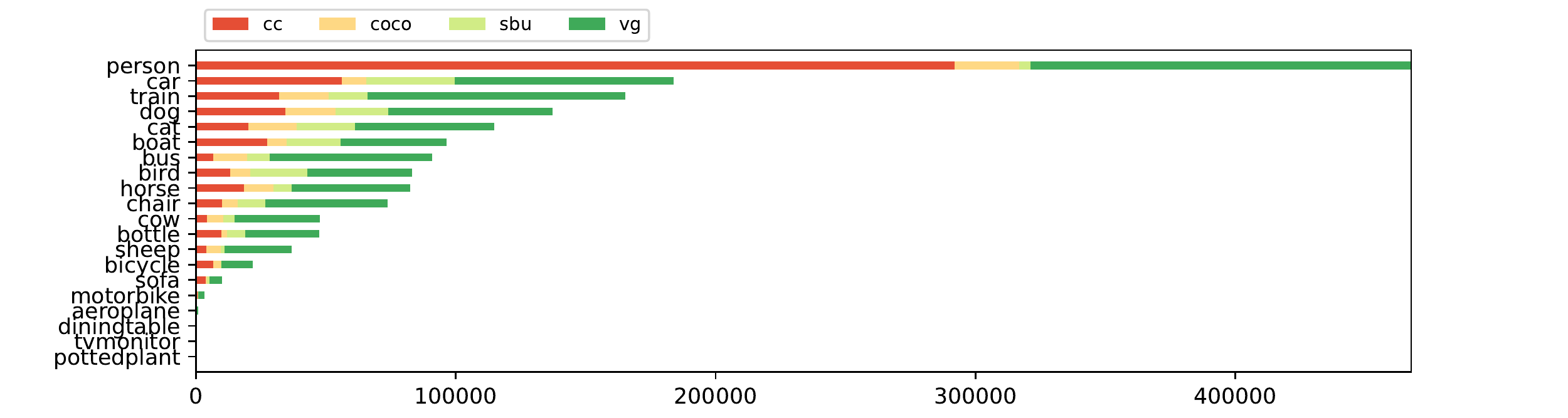}
    \caption{Image captions overlap with Pascal VOC}
    \label{fig:concept_p20}
\end{figure}

\begin{figure}
    \vspace{-66pt}
    \includegraphics[width=.98\linewidth]{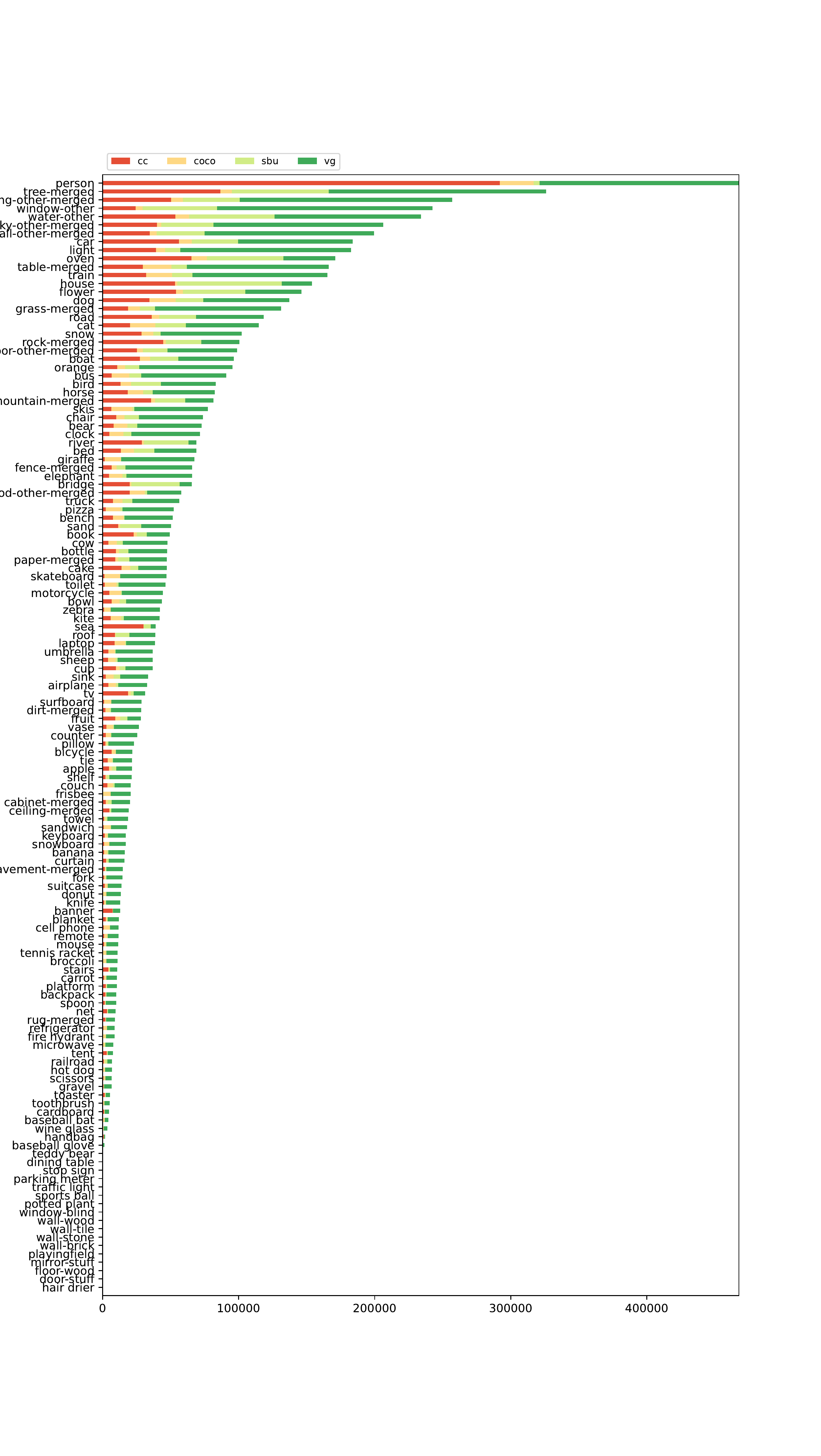}
    \vspace{-35pt}
    \caption{Image captions overlap with COCO}
    \label{fig:concept_coco}
\end{figure}

\begin{figure}
    \includegraphics[width=.98\linewidth]{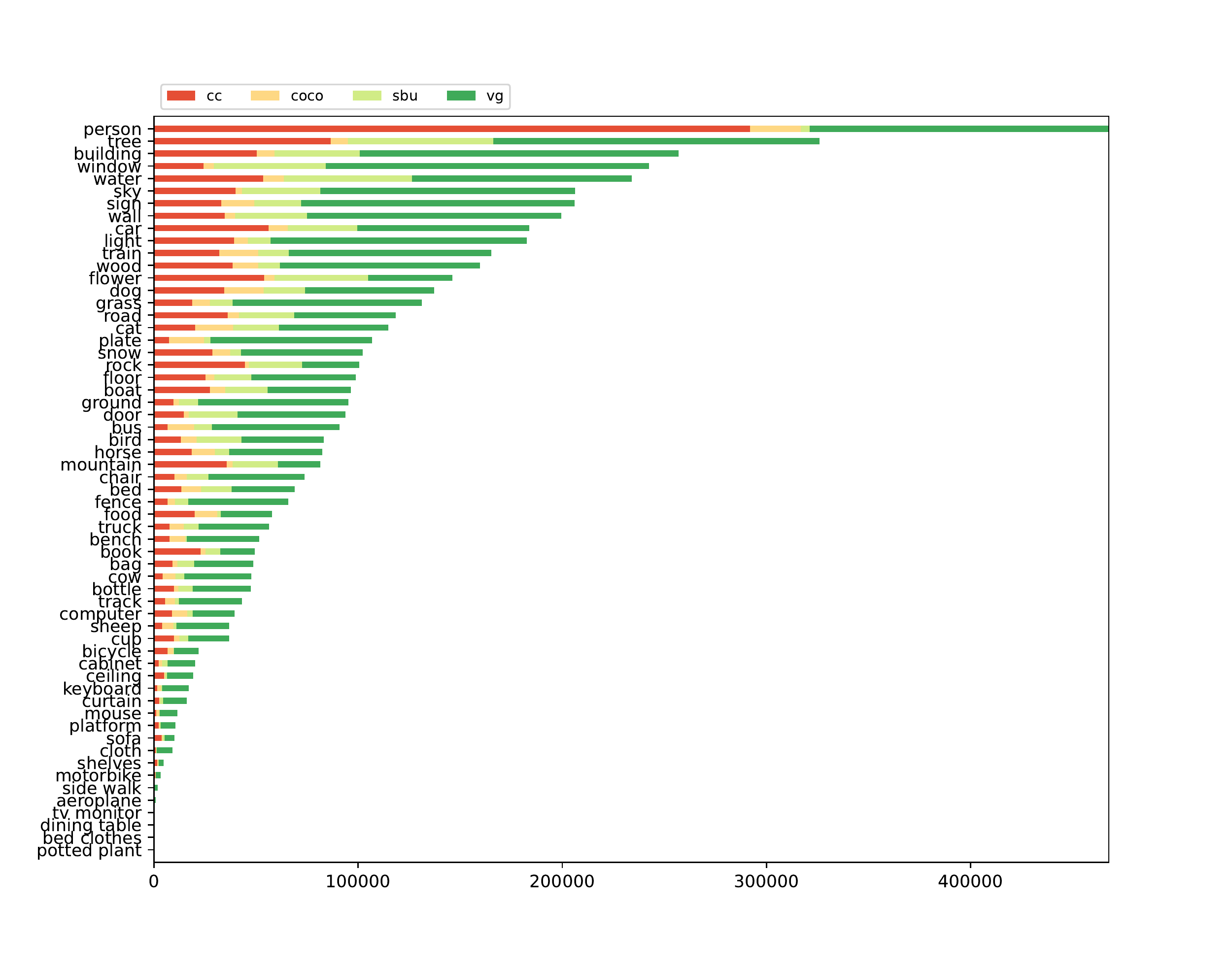}
    \vspace{-4pt}
    \caption{Image captions overlap with Pascal Context-59}
    \label{fig:concept_p59}
\end{figure}

\begin{figure}
    \includegraphics[width=.98\linewidth]{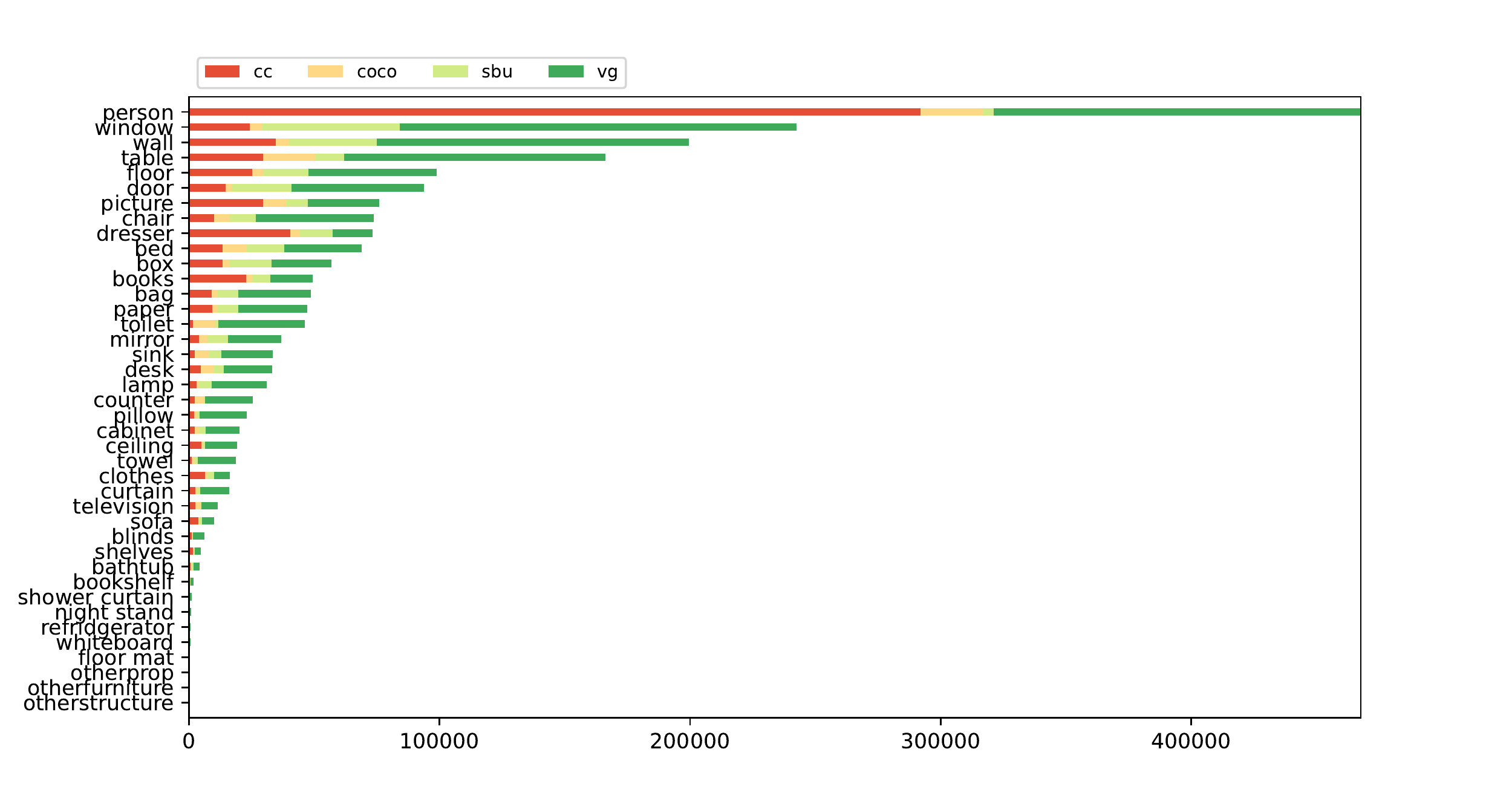}
    \caption{Image captions overlap with ScanNet-40}
    \label{fig:concept_scan40}
\end{figure}

Table~\ref{tab:open-vocab-benchmark} shows the dataset statistics in the benchmark.  It supports all generic segmentation tasks including semantic/instance/panoptic segmentation. It covers a variety of scopes ranging from 20 to 847 classes. In addition, the evaluation scene includes common objects, in-door scenes as well as autonomous driving scenarios. To enable a better understanding of the open-vocabulary ability on the training/evaluation datasets. We evaluate the coverage of training datasets captions and evaluation datasets concepts in Fig.~\ref{fig:concept_ade}-\ref{fig:concept_scan40} (we split the caption into single words and phrases to find mappings in categories). The major results of the open-vocabulary segmentation are evaluated in the main paper, Tab.~\ref{tab:zero-shot-transfer}.

\begin{table}[h]
\footnotesize \setlength{\tabcolsep}{4.0pt}
\resizebox{0.99\linewidth}{!}{
\begin{tabular}{l|ll|ll|l} 
\toprule
\multirow{2}{*}{Method}                                                                       & \multicolumn{4}{c|}{COCO Karpathy} & VQAv2      \\
& IR@1 & TR@1 & CIDEr & BLEU & test-dev \\
\midrule
X-Decoder~(T)                               & 49.3                                 & 66.7                                      & 122.3                                & 37.8                                 & 70.6                                                            \\
X-Decoder-VL~(T)                               & \textcolor{red}{44.3~{\tiny $\downarrow$5.0}}                           & \textcolor{red}{60.3~{\tiny $\downarrow$6.4}}                                     & \textcolor{red}{113.2~{\tiny $\downarrow${9.1}}}                   & \textcolor{red}{34.8~{\tiny $\downarrow${3.0}}}                                 & \textcolor{red}{69.4~{\tiny $\downarrow${1.2}}}                                                          \\
\bottomrule
\end{tabular}
}
\vspace{3pt}
\caption{Compare finetuning result between \ourmodel{} and \ourmodel{}-VL which merely uses 4M image-text pairs for pretraining.}
\label{tab:vlonly}
\end{table}

\begin{table*}[t]
\centering
\footnotesize \setlength{\tabcolsep}{5.2pt}
\begin{tabular}{lcc|cccccc|c|cc|cc} 
\toprule
\multirow{3}{*}{Method} & \multirow{3}{*}{Backbone} & \multirow{3}{*}{Deformable Attn.} & \multicolumn{6}{c|}{Generic Segmentation}                                                    & Referring     & \multicolumn{2}{c|}{Retrieval}     & \multicolumn{2}{c}{Captioning}     \\
                        &                           &                          & \multicolumn{3}{c}{COCO}                      & \multicolumn{3}{c|}{ADE (open)}              & g-Ref         & \multicolumn{2}{c|}{COCO-Karpathy} & \multicolumn{2}{c}{COCO-Karpathy}  \\
                        &                           &                          & PQ            & mAP           & mIoU          & PQ            & mAP          & mIoU          & cIoU          & IR@1          & TR@1               & CIDEr         & BLEU               \\ 
\hline
X-Decoder (T)           & Swin                      & \xmark                        & 50.2          & 38.8          & 61.9          & 17.3          & 9.4          & 23.7          & 55.3          & 28.0          & 43.7               & 79.9          & 24.2               \\
X-Decoder (T)           & Swin                      & \cmark                        & \textbf{52.3} & \textbf{42.7} & \textbf{64.5} & 17.0          & 9.3          & 22.1          & 59.1 & 28.1          & 43.1               & \textbf{87.2} & \textbf{26.9}      \\
\rowcolor[rgb]{0.937,0.937,0.937} X-Decoder (T)           & Focal                     & \xmark                        & 51.4          & 40.5          & 62.8          & \textbf{18.8} & \textbf{9.8} & \textbf{25.0} & \textbf{59.8}          & 30.7          & 48.5               & 79.9          & 24.2               \\
X-Decoder (T)           & Davit                     & \xmark                        & 51.0          & 39.7          & 62.4          & 17.3          & 9.4          & 23.6          & 58.4          & \textbf{31.4} & \textbf{48.8}      & 86.8          & 26.0               \\
\hline
\rowcolor[rgb]{0.937,0.937,0.937} X-Decoder (L)           & Davit                     & \xmark                        & 56.9          & 46.7          & 67.7          & \textbf{21.8}          & \textbf{13.1}          & \textbf{29.6}          & 64.2          & 44.7 &  60.3     & \textbf{111.0}          & \textbf{32.6}               \\
X-Decoder (L)           & Davit                     & \cmark                        &    \textbf{57.4}       & \textbf{48.0}          &   \textbf{69.7}        &   19.1        &      12.6     & 26.6          & \textbf{65.1}          & \textbf{46.2} &   \textbf{61.8}    &  108.2         &     30.1           \\
\bottomrule
\end{tabular}
\vspace{2pt}
\caption{Model architecture inspection among Swin~\cite{liu2021swin}, FocalNet~\cite{yang2022focal} and DaViT~\cite{ding2022davit}. ``Deformable Attn.'' means multi-scale deformable attention~\cite{zhu2020deformable} that is used in Mask2Former~\cite{cheng2022masked}. All numbers are reported in zero-shot manner without any task-specific finetuning, and the row colored in gray corresponds to the architecture used the main paper.}
\vspace{2pt}
\label{tab:arch_comparison}
\end{table*}

\begin{table*}[t]
\centering
\footnotesize \setlength{\tabcolsep}{2.6pt}
\begin{tabular}{lcccc|ccccccccccccccc} 
\toprule
\multirow{2}{*}{Model}                          & \multicolumn{3}{c}{COCO (p/s)} & \multirow{2}{*}{ITP} & \multicolumn{3}{c}{ADE-150}                  & VOC           & PC-59         & PC-459        & SUN           & \multicolumn{2}{c}{SCAN-20}   & SCAN-41       & \multicolumn{3}{c}{Cityscapes}                & \multicolumn{2}{c}{BDD}        \\
                                                & m & cls & cap                  &                      & PQ            & mAP          & mIoU          & mIoU          & mIoU          & mIoU          & mIoU          & mIoU          & PQ            & mIoU          & mIoU          & mAP           & PQ            & mIoU          & PQ             \\ 
\midrule
X-Decoder-Seg~(T)                               & \cmark & \cmark   & \xmark                    & \xmark                    & 13.7          & 6.3          & 18.0          & 89.3          & 59.3          & 11.5          & 16.3          & 8.6           & 16.3          & 6.4           & 46.6          & 14.9          & 30.2          & 36.9          & 13.0           \\
X-Decoder-Seg+(T)                               & \cmark & \cmark   & \cmark                    & \xmark                    & 15.0          & 7.8          & 21.3          & 93.1          & 61.7          & 10.4          & 28.7          & 30.7          & 30.8          & 17.1          & 48.2          & \textbf{16.7} & 37.1          & 40.0          & 13.5           \\
X-Decoder (T)                                   & \cmark & \cmark   & \cmark                    & \xmark                    & 16.6          & 8.3          & 22.3          & 94.4          & 57.6          & 11.9          & 33.1          & \textbf{39.7} & 26.4          & \textbf{21.9} & \textbf{51.0} & 15.6          & 35.5          & \textbf{45.0} & 14.4           \\
\rowcolor[rgb]{0.937,0.937,0.937} X-Decoder (T) & \cmark & \cmark   & \cmark                    & \cmark                    & \textbf{18.8} & \textbf{9.8} & \textbf{25.0} & \textbf{96.2} & \textbf{62.9} & \textbf{12.3} & \textbf{34.5} & 37.8          & \textbf{30.7} & 21.7          & 47.3          & 16.0          & \textbf{37.2} & 42.4          & \textbf{16.4}  \\
\hline
X-Decoder (L-IN21K)                                   & \cmark & \cmark   & \cmark                    & \cmark                    & 19.9          & 11.7          & 29.6          & 95.8          & 54.2          & \textbf{20.5}          & 42.4          & 44.9          & 29.5          & 27.4          & 47.2          & 18.3                                     & 33.3                                     & 44.9          & 15.2           \\
\rowcolor[rgb]{0.937,0.937,0.937} X-Decoder (L)                                   & \cmark & \cmark   & \cmark                    & \cmark                    & \textbf{21.8}                            & \textbf{13.1}                            & \textbf{29.6}          & \textbf{97.7}                                     & \textbf{64.0}          & 16.1          & \textbf{43.0} & \textbf{49.5} & \textbf{39.5} & \textbf{29.7} & \textbf{52.0} & \textbf{24.9}                                     & \textbf{38.1}                                     & \textbf{47.2} & \textbf{17.8} \\

\bottomrule
\end{tabular}
\vspace{2pt}
\caption{More open-vocabulary segmentation results. We report the results for our \ourmodel{} pretrained with COCO segmentation and caption annotations only in 3rd row. Additionally, we compare the model initialized with two different pre-trained large vision backbones, FocalNet-Large and DaViT-d5 trained on ImageNet-21K (row 5) and hundreds of millions of image-text pairs (row 6), respectively.}
\label{Tab:zero_shot_sup}
\end{table*}

\section{Extra Ablation Studies}
\subsection{Complementariness between Vision and VL}
In our main paper, we observed that the vision-language pretraining objectives including image-text contrastive learning and image captioning have clear benefits to image segmentation, particularly in the zero-shot setting. Here, we further study the role of segmentation objectives in vision-language understanding. To investigate, we remove the segmentation data (COCO panoptic segmentation and referring segmentation) and only pretrain \ourmodel{} on the four million image-text pairs, denoted by \ourmodel{}-VL. Afterwards, we transfer the model to downstream VL tasks. As we can see from Table~\ref{tab:vlonly}, the performance significantly drops across all tasks after removing the segmentation data for pretraining. We suspect that segmentation data can help models to learn more fine-grained visual understanding and consequently benefit vision-language tasks. Along with our findings in the main paper, we conclude that \emph{pixel-level segmentation and vision-language learning are complementary to each other for zero-shot and task-specific transfer}.

\subsection{Model Architecture Inspection}
In Table.~\ref{tab:arch_comparison}, we report the results using three different vision backbone architectures, including Swin~\cite{liu2021swin}, FocalNet~\cite{yang2022focal} and DaViT~\cite{ding2022davit}. All models in the first block are with tiny size and trained on the combination of image-label and image-text pairs, following the settings in UniCL~\cite{yang2022unified}. In the second block, all the models are initialized with Florence~\cite{yuan2021florence} pre-trained DaVit-d5 model. Through the comparisons, we have the following observations: (1) FocalNet and DaViT achieve better performance than Swin across all metrics. Particularly, FocalNet achieves the best performance on generic and referring segmentation, while DaViT is better on the zero-shot vision-language evaluations; (2) After adding the deformable attention, we can see a boost on supervised segmentation but significant (especially large model) degradation on the open-vocabulary segmentation on ADE20K dataset. Based on these experimental results, we make the design choices as mentioned in our main submission: (1) we remove deformable attention in the favor of open-vocabulary segmentation; (2) we use FocalNet as the tiny vision encoder and train it by ourselves using UniCL, while using DaViT~\cite{yuan2021florence} as the base and large vision encoder.

\subsection{Open-Vocabulary Generic Segmentation Settings Inspection}
In Tab.~\ref{Tab:zero_shot_sup}, we study the progressive enrichment of data and training settings as well as the pre-trained model usage. X-Decoder-Seg is the baseline of adding a text encoder to Mask2Former~\cite{cheng2022masked} with a learnable language encoder. X-Decoder-Seg$^+$ takes use of caption nouns for Hungarian matching to enrich the vocabulary size. In addition to the main paper, we add row 3 in Tab.~\ref{Tab:zero_shot_sup} to demonstrate the performance of X-Decoder with only coco image text pairs. Comparing 3rd row and 4th row, we find adding extra image-text pairs for pretraining clearl improve open-vocabulary segmentation performance especially when the vocabulary size is large (e.g. ADE-150, CONTEXT-59/459). The way of pretraining vision backbone also matters. Comparing the last two rows side by side, though the backbone model sizes are similar, using ImageNet-21K for pretraining leads to inferior performance on most of the datasets except for CONTEXT-459 which contains most number of categories. These results demonstrate the benefits of using more image-text pairs for pretraining the vision backbone or our \ourmodel{}.

\begin{table*}[]
\scriptsize \setlength{\tabcolsep}{1.2pt}
\centering
\resizebox{1.0\linewidth}{!}{
\begin{tabular}{l|lccc|l} 
\toprule
\multirow{2}{*}{Dataset} & \multirow{2}{*}{Categories}                & \multirow{2}{*}{\# Class} & \multicolumn{2}{l|}{\# Images} & \multirow{2}{*}{URL}                                                                                      \\
                         &                                            &                           & Train & Val                    &                                                                                                           \\ 
\midrule
Phones                   & [phone]                                  & 1                         & 25    & 11                     & {\scriptsize \url{https://universe.roboflow.com/workspace-c4esq/phone-xccez/dataset/1                                     } } \\
Elephants                & [elephant]                               & 1                         & 883   & 99                     & {\scriptsize \url{https://universe.roboflow.com/ds/4YwrXd1bFy?key=Q5GY9ITu14                                              } } \\
Hand-Metal               & [hand, metal]                            & 2                         & 504   & 65                     & {\scriptsize \url{https://universe.roboflow.com/nk950357-gmail-com/lab-k8hyn                                              } } \\
Watermelon               & [watermelon]                             & 1                         & 65    & 23                     & {\scriptsize \url{https://universe.roboflow.com/gnous-b5xq6/my\_project-38aqt/dataset/4                                   } } \\
House-Parts              & [aluminium door, aluminium window, ...]  & 22                        & 700   & 201                    & {\scriptsize \url{https://universe.roboflow.com/testcoco/abc-fqun0/dataset/1                                              } } \\
HouseHold-Items          & [bottle, mouse, perfume, phone]          & 4                         & 45    & 3                      & {\scriptsize \url{https://universe.roboflow.com/maths/household-items-sltdd/dataset/2                                     } } \\
Strawberry               & [R\_strawberry, people]                  & 2                         & 971   & 87                     & {\scriptsize \url{https://universe.roboflow.com/strawberry-25w7z/strawberry\_coco\_1/dataset/1                            } } \\
Fruits                   & [apple, lemon, orange, pear, strawberry] & 5                         & 120   & 9                      & {\scriptsize \url{https://universe.roboflow.com/ds/PEVo9xHLFl?key=PXeJGF0D5q                                              } } \\
Nutterfly-Squireel       & [butterfly, squirrel]                    & 2                         & 951   & 237                    & {\scriptsize \url{https://universe.roboflow.com/handwashhygeine/nature-3tkys                                              } } \\
Hand                     & [Hand-Segmentation, hand]                & 2                         & 210   & 60                     & {\scriptsize \url{https://universe.roboflow.com/rmutsb-xxgii/hand-segmentation-gqzuh/dataset/1                            } } \\
Garbage                  & [bin, garbage, pavement, road]           & 4                         & 325   & 142                    & {\scriptsize \url{https://universe.roboflow.com/project-blmh9/d2-bj1a0/dataset/1                                          } } \\
Chicken                  & [chicken]                                & 1                         & 19    & 1                      & {\scriptsize \url{https://universe.roboflow.com/nena-trikic-ljxt3/chickenstf/dataset/1                                    } } \\
Rail                     & [rail]                                   & 1                         & 3067  & 1069                   & {\scriptsize \url{https://universe.roboflow.com/wzk789wzk-gmail-com/rail\_dataset/dataset/4                               } } \\
Airplane-Parts           & [Airplane, Body, Cockpit, Engine, Wing]  & 5                         & 39    & 7                      & {\scriptsize \url{https://universe.roboflow.com/foxehcorp/foxehcorp\_airplane\_dataset/dataset/4/download                 } } \\
Brain-Tumor              & [tumor]                                  & 1                         & 236   & 28                     & {\scriptsize \url{https://universe.roboflow.com/detection-qskiw/segmnetation/dataset/2                                    } } \\
Poles                    & [poles]                                  & 1                         & 11    & 3                      & {\scriptsize \url{https://universe.roboflow.com/ohsee/pole2/dataset/2                                                     } } \\
Electric-Shaver          & [caorau]                                 & 1                         & 288   & 24                     & {\scriptsize \url{https://universe.roboflow.com/fpt-university-1tkhk/caurau                                               } } \\
Bottles                  & [bottle, can, label]                     & 3                         & 357   & 16                     & {\scriptsize \url{https://universe.roboflow.com/beerup/bottels2/dataset/1                                                 } } \\
Toolkits                 & [Allen-key, block, gasket, ...]          & 8                         & 48    & 6                      & {\scriptsize \url{https://universe.roboflow.com/mst/mask-2ihnt/dataset/1                                                  } } \\
Trash                    & [Aluminium foil, Cigarette, ...]         & 12                        & 832   & 92                     & {\scriptsize \url{https://universe.roboflow.com/sara-najafi/trash\_segmentation2/dataset/2                                } } \\
Salmon-Fillet            & [Salmon\_fillet]                         & 1                         & 1991  & 64                     & {\scriptsize \url{https://universe.roboflow.com/rishik-mishra-rljwe/f1225                                                 } } \\
Puppies                  & [puppy]                                  & 1                         & 15    & 3                      & {\scriptsize \url{https://universe.roboflow.com/marcin-bak/puppies-fmoxu/dataset/2                                        } } \\
Tablets                  & [tablets]                                & 1                         & 237   & 13                     & {\scriptsize \url{https://universe.roboflow.com/detection-qskiw/tablets-instance-segmentation/dataset/1                   } } \\
Cows                     & [cow]                                    & 1                         & 630   & 60                     & {\scriptsize \url{https://universe.roboflow.com/new-workspace-5abdm/maskrcnn-ofglr/dataset/2                              } } \\
Ginger-Garlic            & [garlic, ginger]                         & 2                         & 28    & 8                      & {\scriptsize \begin{tabular}[c]{@{}l@{}}\url{https://universe.roboflow.com/george-brown-college-1omrb/}\\\url{ginger-and-garlic-object-segmentation/dataset/1}\end{tabular}}  \\
\bottomrule
\end{tabular}
}
\vspace{2pt}
\caption{Meta information of \textit{SegInW} benchmark. We list the source links, annotated category names and number of categories for each dataset.}
\vspace{6pt}
\label{tab:seginw_meta}
\end{table*}

\begin{figure*}[t]
    \centering
     \begin{subfigure}[b]{0.14\linewidth}
         \centering
         \includegraphics[width=\textwidth, height=\textwidth]{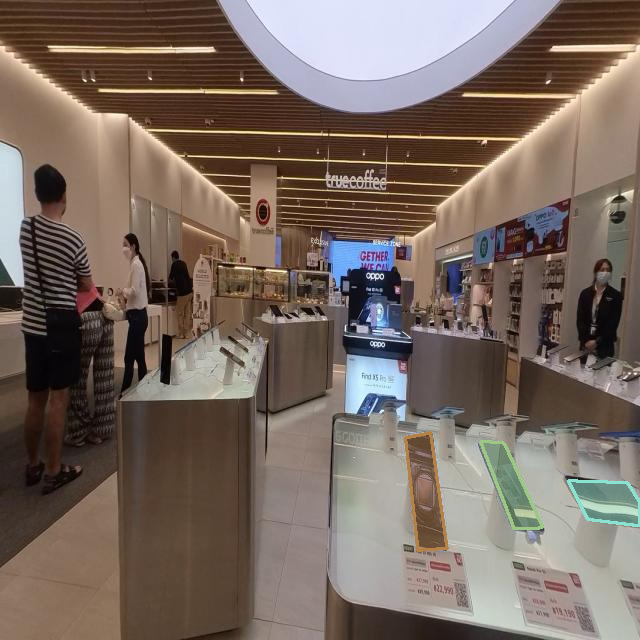}
         \caption{\textit{Phones}}
         \label{fig:five over 1}
     \end{subfigure}
     \hfill
     \begin{subfigure}[b]{0.14\linewidth}
         \centering
         \includegraphics[width=\textwidth]{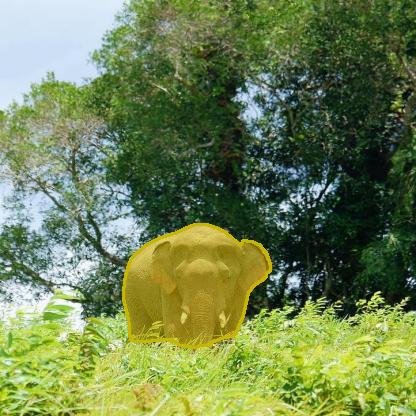}
         \caption{\textit{Elephants}}
         \label{fig:five over 2}
     \end{subfigure}
     \hfill
     \begin{subfigure}[b]{0.14\linewidth}
         \centering
         \includegraphics[width=\textwidth]{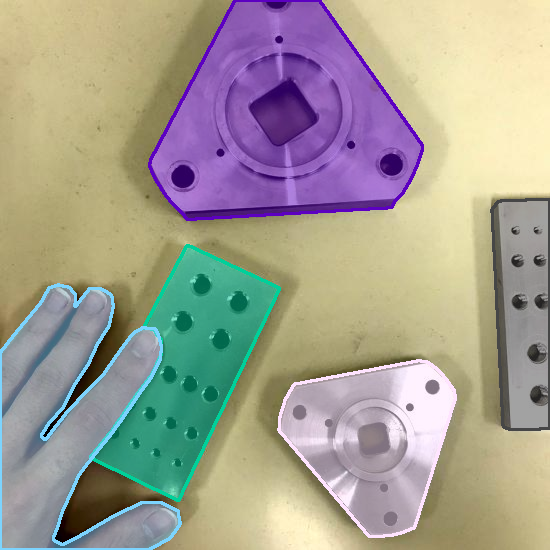}
         \caption{\textit{Hand-Metal}}
         \label{fig:five over 3}
     \end{subfigure}
     \hfill
     \begin{subfigure}[b]{0.14\linewidth}
         \centering
         \includegraphics[width=\textwidth]{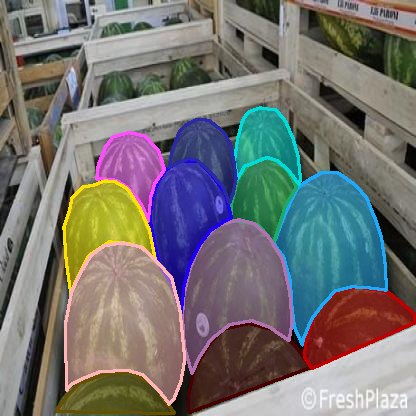}
         \caption{\textit{Watermelon}}
         \label{fig:five over 4}
     \end{subfigure}
     \hfill
     \begin{subfigure}[b]{0.14\linewidth}
         \centering
         \includegraphics[width=\textwidth]{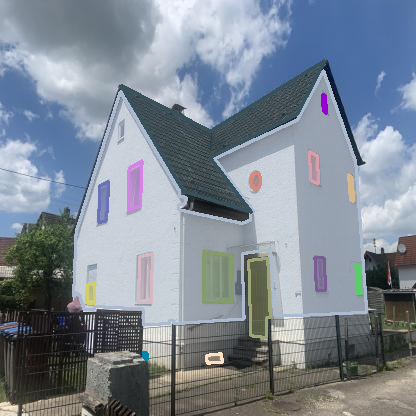}
         \caption{\textit{House-parts}}
         \label{fig:five over 5}
     \end{subfigure}    
     \vspace{10pt}
     
    \centering
     \begin{subfigure}[b]{0.14\linewidth}
         \centering
         \includegraphics[width=\textwidth]{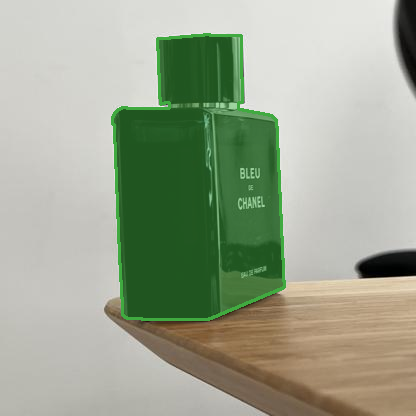}
         \caption{\textit{Household-items}}
         \label{fig:five over 6}
     \end{subfigure}
     \hfill
     \begin{subfigure}[b]{0.14\linewidth}
         \centering
         \includegraphics[width=\textwidth]{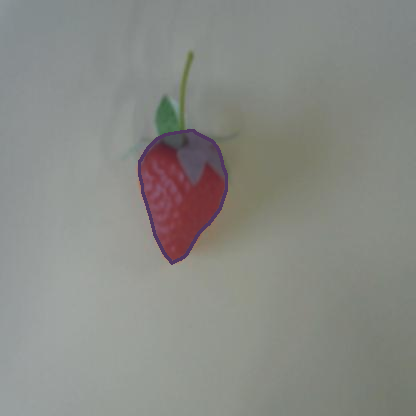}
         \caption{\textit{Strawberry}}
         \label{fig:five over 7}
     \end{subfigure}
     \hfill
     \begin{subfigure}[b]{0.14\linewidth}
         \centering
         \includegraphics[width=\textwidth]{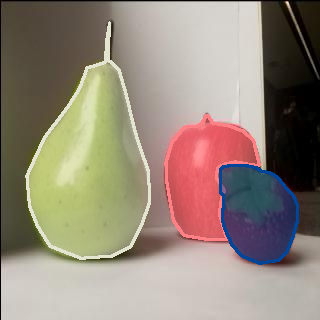}
         \caption{\textit{Fruits}}
         \label{fig:five over 8}
     \end{subfigure}
     \hfill
     \begin{subfigure}[b]{0.14\linewidth}
         \centering
         \includegraphics[width=\textwidth]{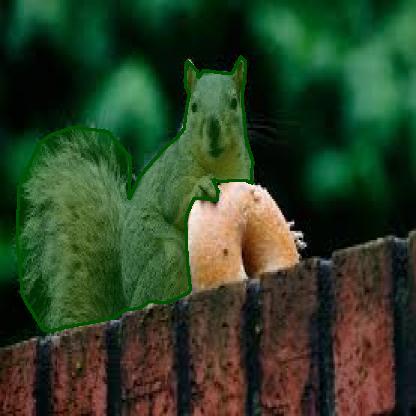}
         \caption{\textit{Butterfly-Squirrel}}
         \label{fig:five over 9}
     \end{subfigure}
     \hfill
     \begin{subfigure}[b]{0.14\linewidth}
         \centering
         \includegraphics[width=\textwidth]{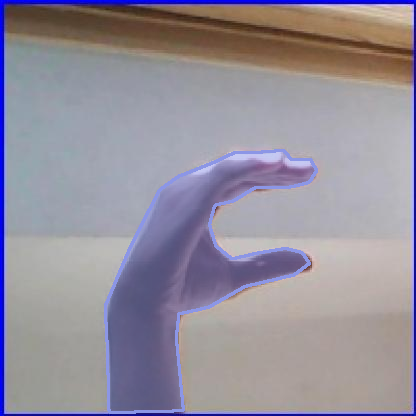}
         \caption{\textit{Hand}}
         \label{fig:five over 10}
     \end{subfigure}       
     \vspace{10pt}
     
    \centering
     \begin{subfigure}[b]{0.14\linewidth}
         \centering
         \includegraphics[width=\textwidth]{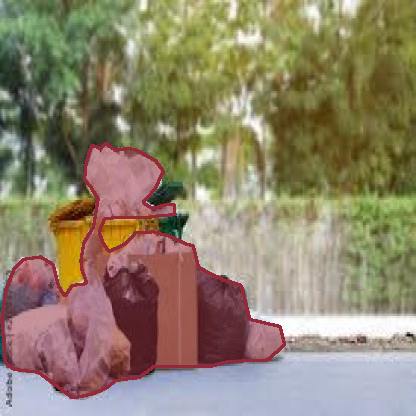}
         \caption{\textit{Garbage}}
         \label{fig:five over 11}
     \end{subfigure}
     \hfill
     \begin{subfigure}[b]{0.14\linewidth}
         \centering
         \includegraphics[width=\textwidth]{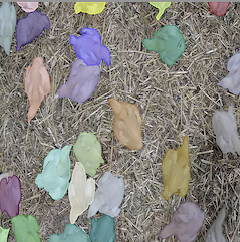}
         \caption{\textit{Chicken}}
         \label{fig:five over 12}
     \end{subfigure}
     \hfill
     \begin{subfigure}[b]{0.14\linewidth}
         \centering
          \includegraphics[width=\textwidth, height=\textwidth]{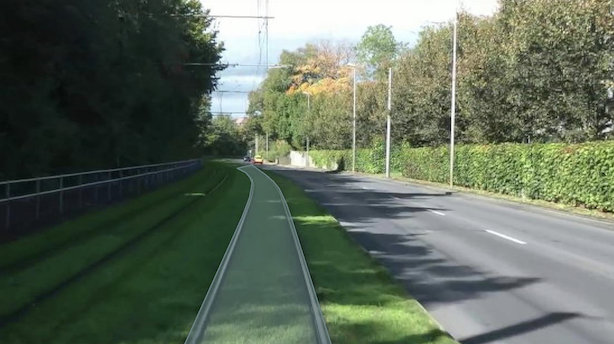}
         \caption{\textit{Rail}}
         \label{fig:five over 13}
     \end{subfigure}
     \hfill
     \begin{subfigure}[b]{0.14\linewidth}
         \centering
         \includegraphics[width=\textwidth]{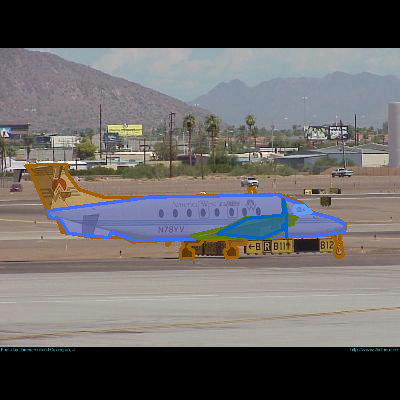}
         \caption{\textit{Airplane-parts}}
         \label{fig:five over 14}
     \end{subfigure}     
     \hfill
     \begin{subfigure}[b]{0.14\linewidth}
         \centering
         \includegraphics[width=\textwidth]{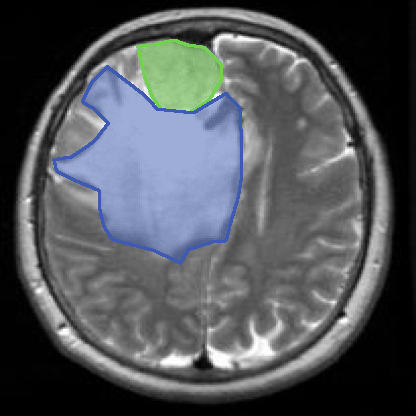}
         \caption{\textit{Brain-tumor}}
         \label{fig:five over 15}
     \end{subfigure}     
     \vspace{10pt}
     
    \centering
     \begin{subfigure}[b]{0.14\linewidth}
         \centering
         \includegraphics[width=\textwidth]{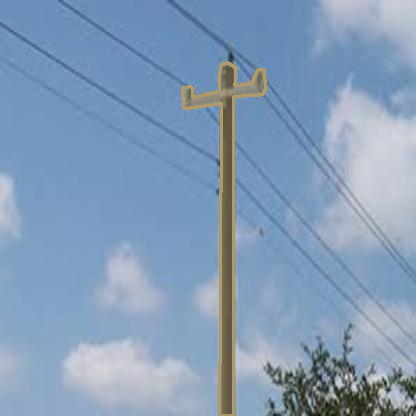}
         \caption{\textit{Poles}}
         \label{fig:five over 16}
     \end{subfigure}
     \hfill
     \begin{subfigure}[b]{0.14\linewidth}
         \centering
         \includegraphics[width=\textwidth]{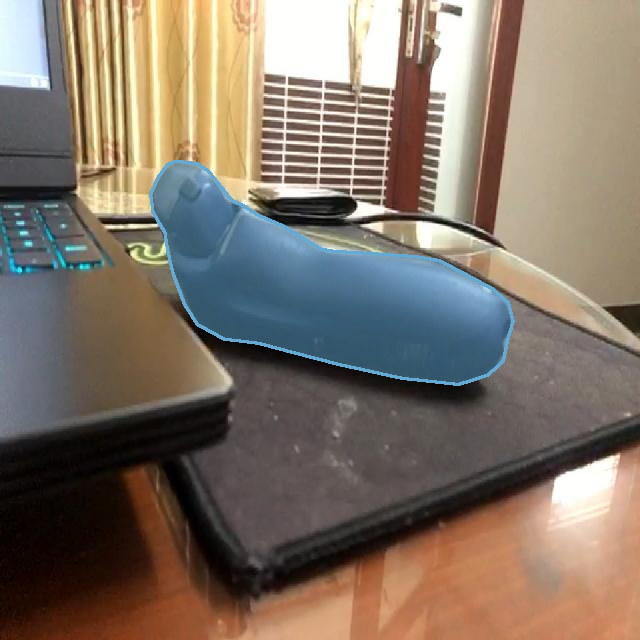}
         \caption{\textit{Electric-shaver}}
         \label{fig:five over 17}
     \end{subfigure}
     \hfill
     \begin{subfigure}[b]{0.14\linewidth}
         \centering
         \includegraphics[width=\textwidth, height=\textwidth]{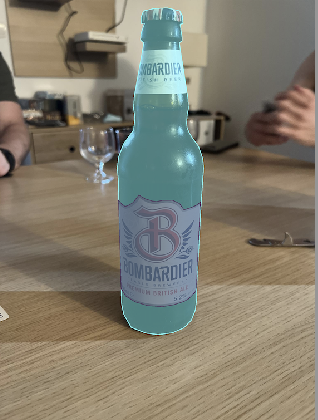}
         \caption{\textit{Bottles}}
         \label{fig:five over 18}
     \end{subfigure}
     \hfill
     \begin{subfigure}[b]{0.14\linewidth}
         \centering
         \includegraphics[width=\textwidth, height=\textwidth]{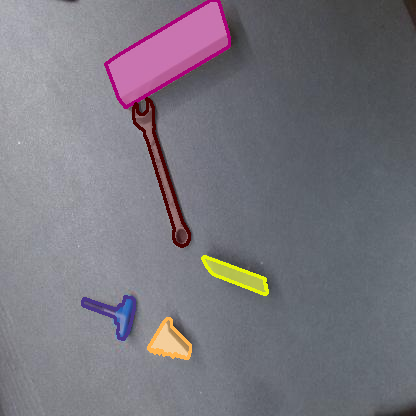}
         \caption{\textit{Toolkits}}
         \label{fig:five over 19}
     \end{subfigure}     
     \hfill
     \begin{subfigure}[b]{0.14\linewidth}
         \centering
         \includegraphics[width=\textwidth, height=\textwidth]{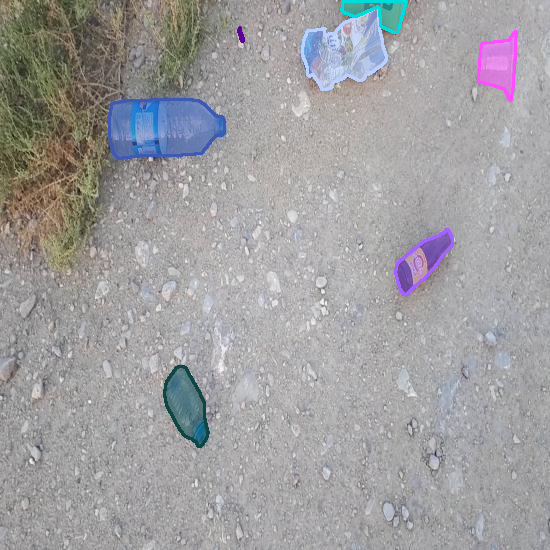}
         \caption{\textit{Trash}}
         \label{fig:five over 20}
     \end{subfigure}          
     \vspace{10pt}
     
    \centering
     \begin{subfigure}[b]{0.14\linewidth}
         \centering
         \includegraphics[width=\textwidth]{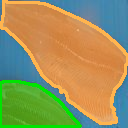}
         \caption{\textit{Salmon}}
         \label{fig:five over 21}
     \end{subfigure}
     \hfill
     \begin{subfigure}[b]{0.14\linewidth}
         \centering
         \includegraphics[width=\textwidth]{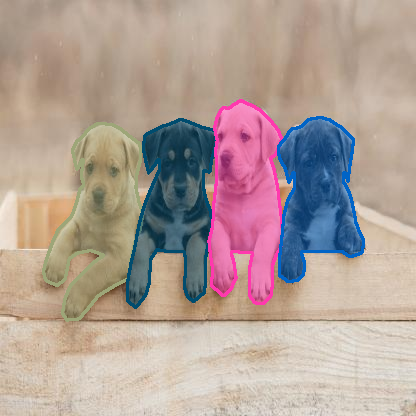}
         \caption{\textit{Puppies}}
         \label{fig:five over 22}
     \end{subfigure}
     \hfill
     \begin{subfigure}[b]{0.14\linewidth}
         \centering
         \includegraphics[width=\textwidth]{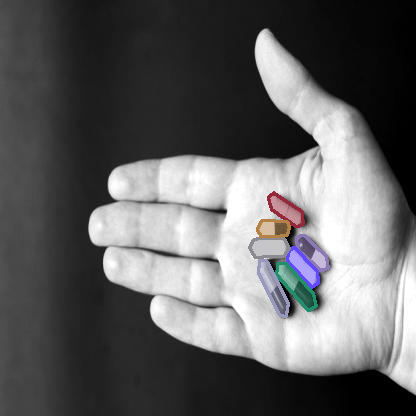}
         \caption{\textit{Tablets}}
         \label{fig:five over 23}
     \end{subfigure}
     \hfill
     \begin{subfigure}[b]{0.14\linewidth}
         \centering
         \includegraphics[width=\textwidth, height=\textwidth]{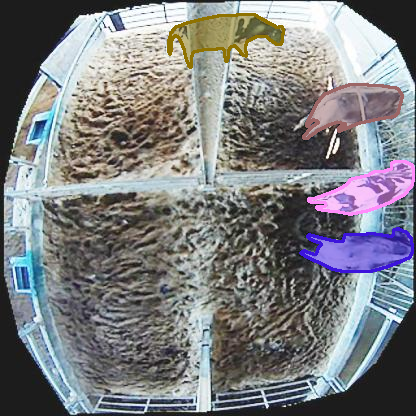}
         \caption{\textit{Cows}}
         \label{fig:five over 24}
     \end{subfigure}
     \hfill
     \begin{subfigure}[b]{0.14\linewidth}
         \centering
         \includegraphics[width=\textwidth, height=\textwidth]{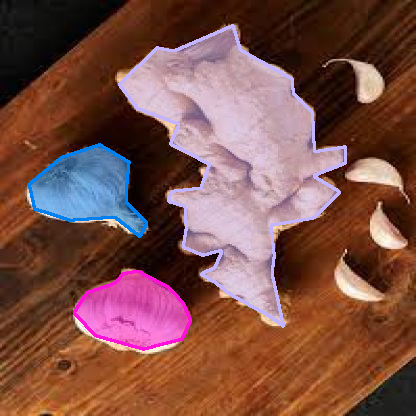}
         \caption{\textit{Ginger-Garlic}}
         \label{fig:five over 25}
     \end{subfigure}    
     \vspace{10pt}
     \caption{Examplar images and annotations in \textit{SegInW} benchmark. The benchmark covers a diversity of visual domains and concepts in the daily life.}
     \label{fig:seginw_exemplar}
\end{figure*}

\begin{figure*}[t]
    \centering
    \vspace{-12pt}
    \includegraphics[width=0.95\linewidth]{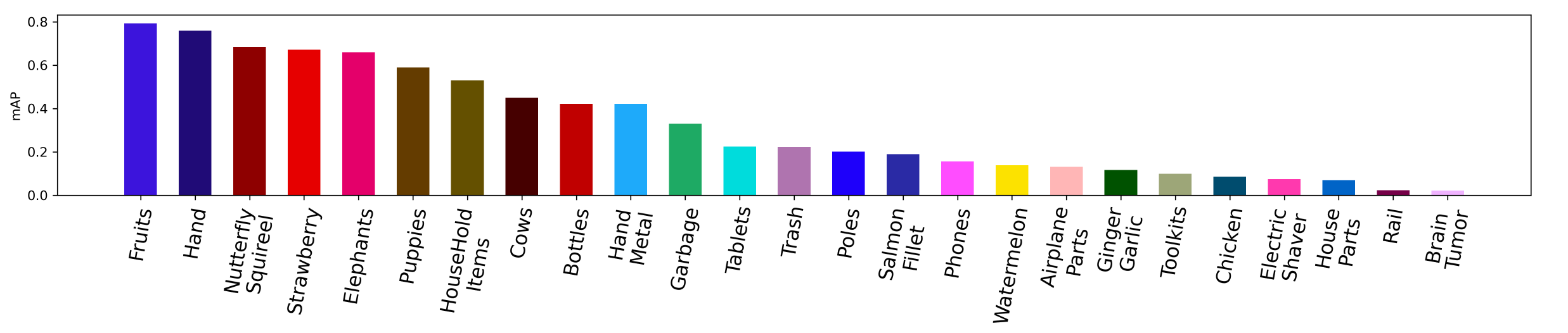}
    \vspace{0pt}
    \caption{Zero-shot segmentation performance on \textit{SeginW} with X-Decoder-L model. We report the mAP in descending order.}
    \label{fig:barchat}    
\end{figure*}

\begin{figure*}[t]
    \centering
     \begin{subfigure}[b]{0.24\linewidth}
         \centering
         \includegraphics[width=\textwidth]{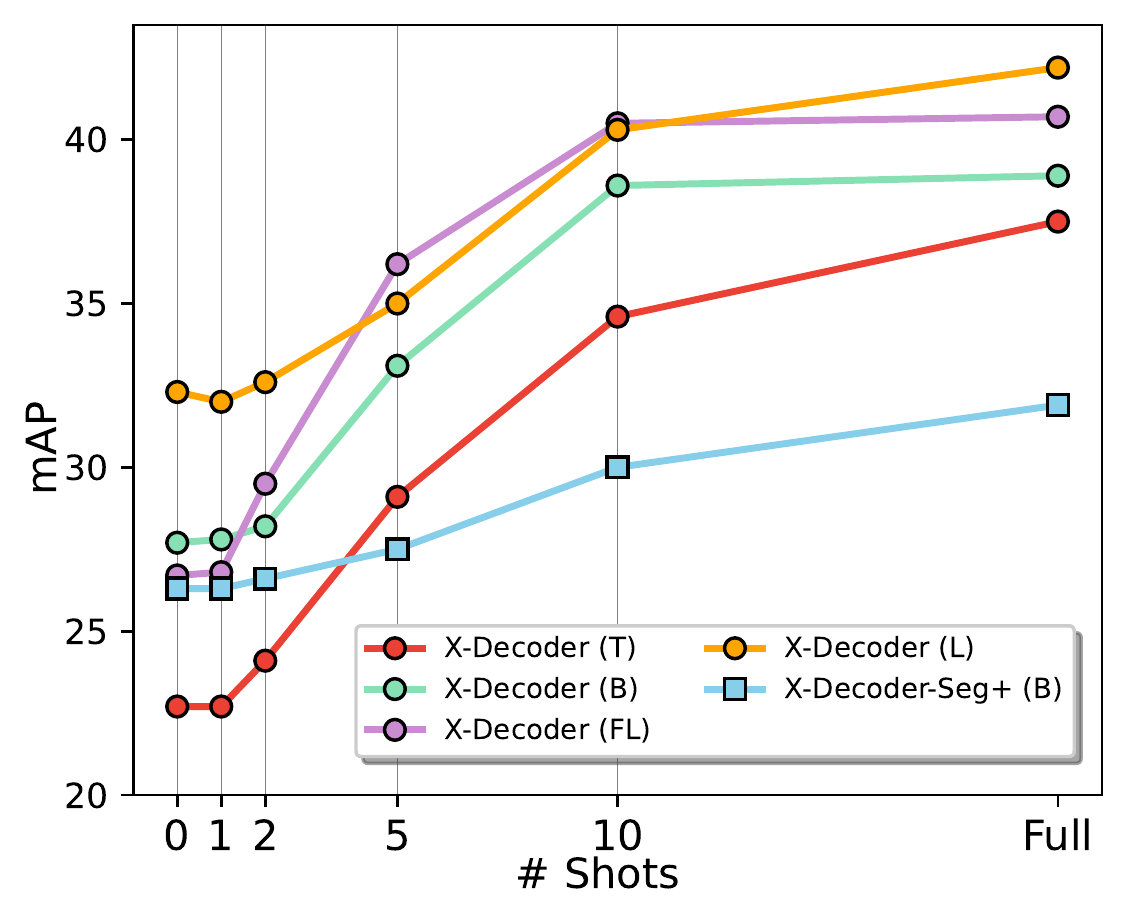}
         \caption{\textit{Tune Linear (0.26M)}}
         \label{fig:seginw_chart1}
     \end{subfigure}
     \hfill
     \begin{subfigure}[b]{0.24\linewidth}
         \centering
         \includegraphics[width=\textwidth]{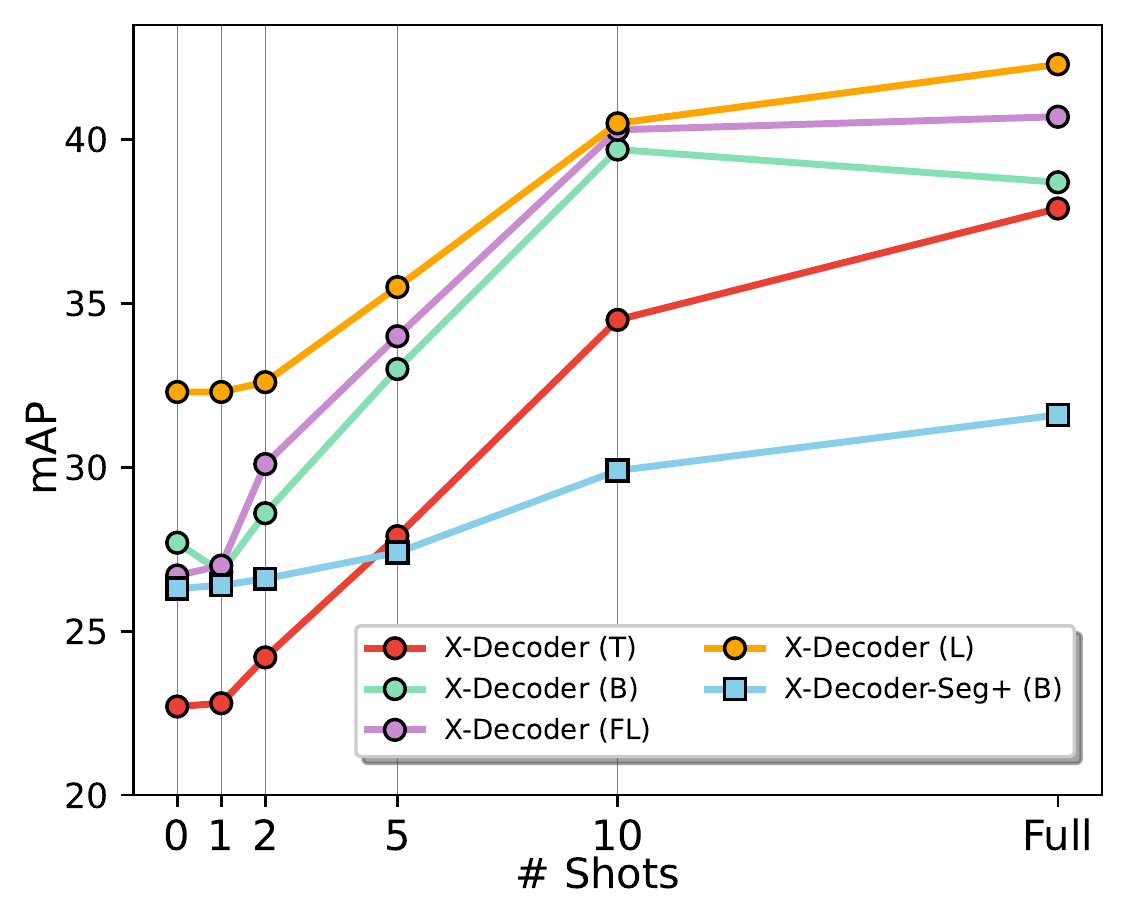}
         \caption{\textit{Tune Prompt (1.15M)}}
         \label{fig:seginw_chart2}
     \end{subfigure}
     \hfill
     \begin{subfigure}[b]{0.24\linewidth}
         \centering
         \includegraphics[width=\textwidth]{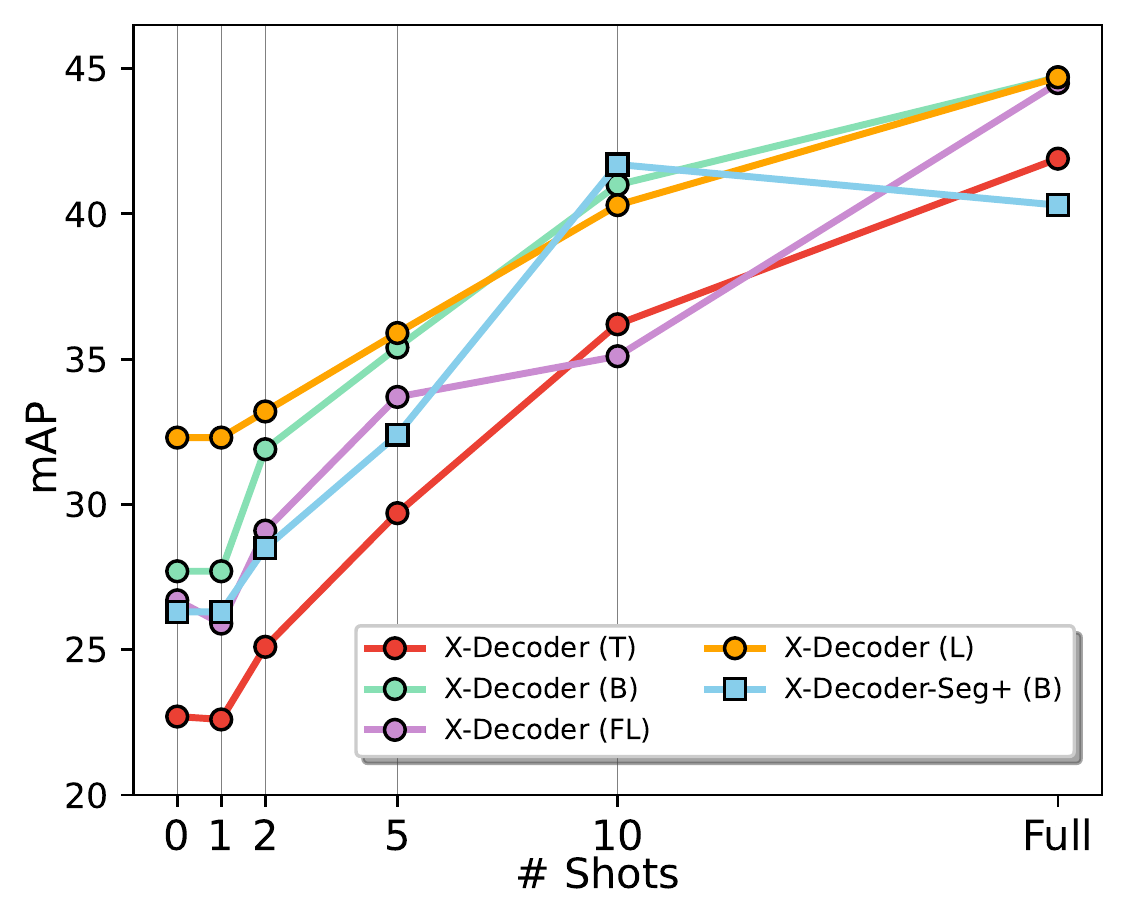}
         \caption{\textit{Tune Decoder (39.3M)}}
         \label{fig:seginw_chart3}
     \end{subfigure}
     \hfill
     \begin{subfigure}[b]{0.24\linewidth}
         \centering
         \includegraphics[width=\textwidth]{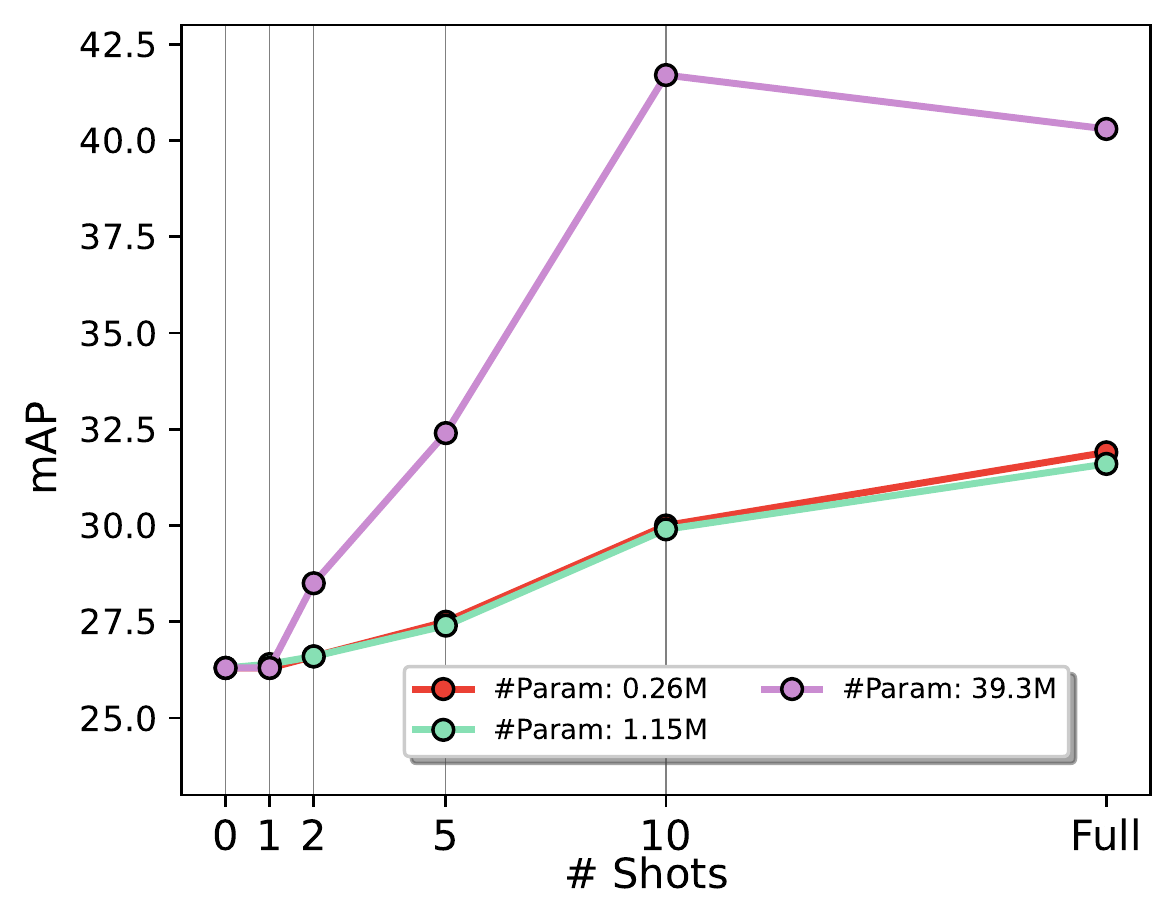}
         \caption{\textit{X-Decoder-Seg+ (B)}}
         \label{fig:seginw_chart4}
     \end{subfigure}
     \vspace{15pt}

    \centering
     \begin{subfigure}[b]{0.24\linewidth}
         \centering
         \includegraphics[width=\textwidth]{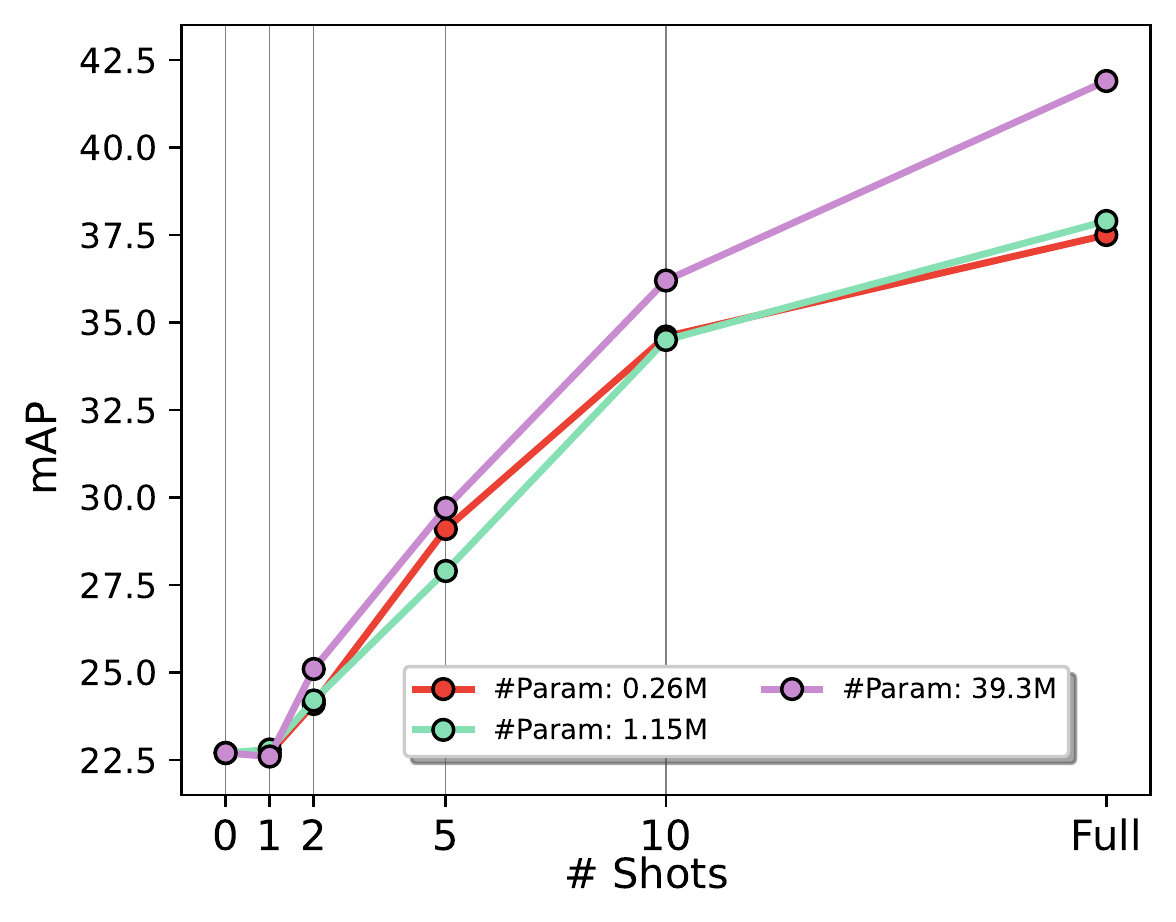}
         \caption{\textit{X-Decoder (T)}}
         \label{fig:seginw_chart5}
     \end{subfigure}
     \hfill
     \begin{subfigure}[b]{0.24\linewidth}
         \centering
         \includegraphics[width=\textwidth]{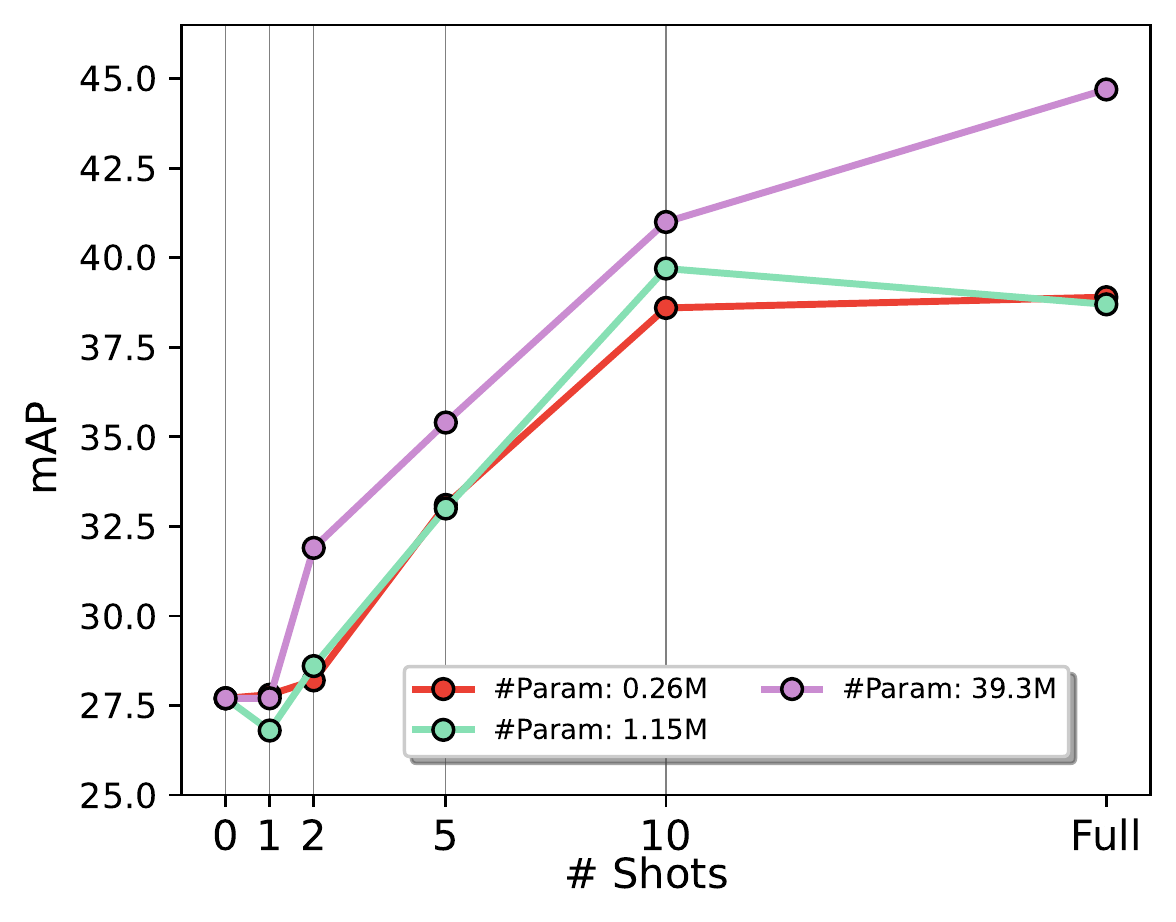}
         \caption{\textit{X-Decoder (B)}}
         \label{fig:seginw_chart6}
     \end{subfigure}
     \hfill
     \begin{subfigure}[b]{0.24\linewidth}
         \centering
         \includegraphics[width=\textwidth]{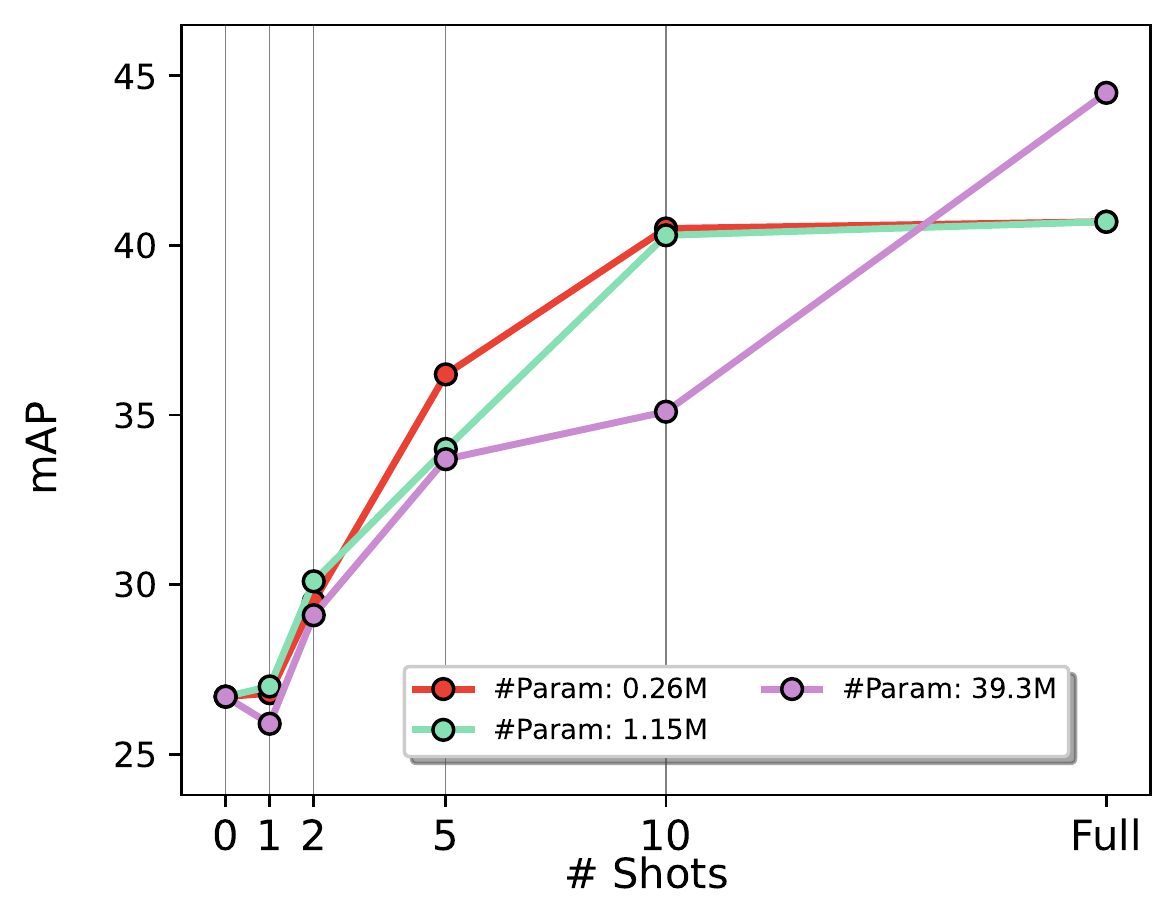}
         \caption{\textit{X-Decoder (L-IN21K)}}
         \label{fig:seginw_chart7}
     \end{subfigure}
     \hfill
     \begin{subfigure}[b]{0.24\linewidth}
         \centering
         \includegraphics[width=\textwidth]{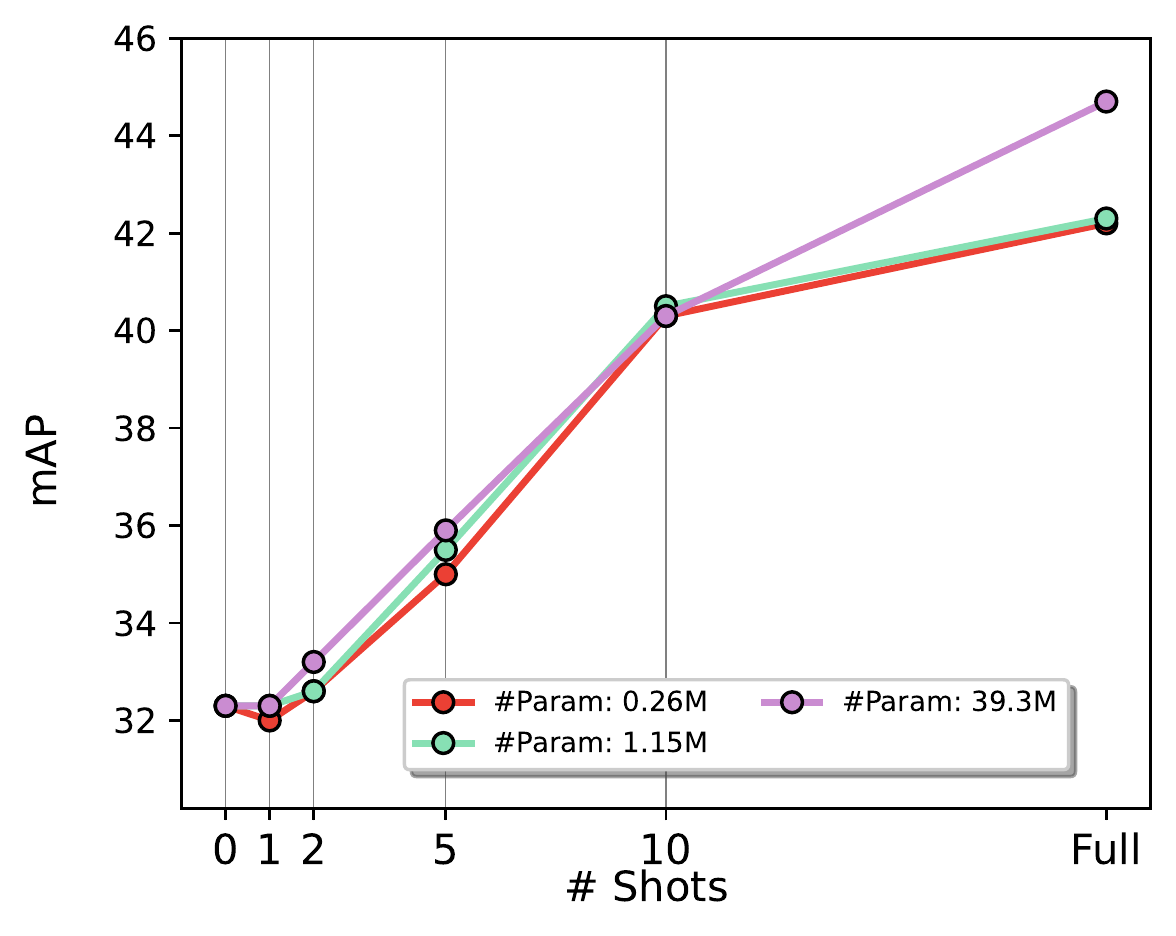}
         \caption{\textit{X-Decoder (L)}}
         \label{fig:seginw_chart8}
     \end{subfigure}
     \vspace{15pt}
     \caption{(a-c) Line chart of tuning shot and mAP with different tuning strategies on each backbone architecture. (d-h) Line chart of tuning shot and mAP with different backbone architecture on different number of tuning parameters (strategies). }
     \label{fig:seginw_linechart}
\end{figure*}

\section{Segmentation In the Wild Benchmark}
As shown in the main submission, our X-Decoder exhibits a strong generalization ability to segment images in ten settings of seven datasets from different domains, without any dataset-specific finetuning. Inspired by the object detection in the wild setting proposed in GLIP~\cite{li2022grounded}, we resort to more domain-specific datasets on the web to further examine the generality of our model. Specifically, we download 55 instance segmentation datasets from Roboflow~\footnote{\url{https://roboflow.com/}}. Afterward, we clean the datasets by excluding those containing visually undetectable categories (e.g. Different species of plant) or categories labeled with other languages. In the end, we compile 25 datasets that are suitable for evaluation into \textit{segmentation in the wild} (\textit{SegInW}) benchmark and report instance segmentation mAP. The dataset meta information is listed in Tab.~\ref{tab:seginw_meta}, and examplar images are shown in Fig.~\ref{fig:seginw_exemplar}.

On the \textit{SegInW} benchmark, we evaluate zero-shot, few-shot, and fine-tuned segmentation for five models (X-Decoder-Seg$^+$ as baselines, and \ourmodel{} with different visual backbone) on three different tuning scales. In Fig.~\ref{fig:barchat}, we report the zero-shot instance segmentation performance on 25 datasets separately in a descending order. Accordingly, \ourmodel{} shows reasonably good generalization ability to a wide range of visual and concept domains.  Specifically, it achieves higher mAP on common objects like fruits and animals but lower ones on fine-grained datasets like toolkits and rare concepts like rail and brain tumor. In Fig.~\ref{fig:seginw_linechart}, we further show the line chars for few-shot learning and fully-finetuning, and observe that:

\noindent
\textbf{\ourmodel{} has privilege on small-scale tuning.} As shown in Fig.~\ref{fig:seginw_linechart}~(a-b), comparing with X-Decoder-Seg$^+$ that only extract noun phrase to increase vocabulary size, \ourmodel{} performs much better with few-shot/finetune setting. Although \ourmodel{} (B) and X-Decoder-Seg$^+$ (B) have similar zero-shot performance, the gap increases with the number of images tuned. However, as the number of parameters tuned increased by a large margin Fig.~\ref{fig:seginw_linechart} (c), the performance gap between \ourmodel{} and X-Decoder-Seg$^+$ is shrunk to a small margin.

\noindent
\textbf{Zero-Shot gap could be bridged by tuning.}
\ourmodel{} (L) and \ourmodel{} (L-IN21K) are initialized with different pre-trained image backbones. Specifically, \ourmodel{} (L) is initialized by Florence~\cite{yuan2021florence} pre-trained Davit-d5, whereas \ourmodel{} (L-IN21K) is initialized with FocalNet-L pretrained on ImageNet-21k~\cite{deng2009imagenet}. As shown in Fig.~\ref{fig:seginw_linechart}~(a-c), although the gap between \ourmodel{}-L and \ourmodel{}-L-IN21K on the zero-shot setting is relatively large. However, the gap on 5/10/full finetuned settings is much smaller and even cross in some settings. 

\noindent
\textbf{Tuning class embedding is enough for few-shot settings.} As shown in Fig.~\ref{fig:seginw_linechart} (e-h), on the smaller scale backbone including (T/B), although tuning the full decoder has a better result, the gap is not obvious on 0-10 shots. And on larger scale models including L/L-IN21K, tuning with class embedding has similar/better results on 0-10 shots.

We show more detailed results in Table~\ref{tab:seginw_linear},  Table~\ref{tab:seginw_prompt} and Table~\ref{tab:seginw_finetune}. Similar to Table~\ref{tab:efficient-finetuning}, we report the number of parameters tuned in each setting. 

\section{Extra Visualization}

In this part, we demonstrate the generalization ability to video datasets and flexibility to support task compositions for \ourmodel{} with more qualitative visualizations. 

\subsection{Zero-Shot Generic Video Segmentation}
\begin{figure*}[h]
    \centering
    \includegraphics[width=.95\linewidth]{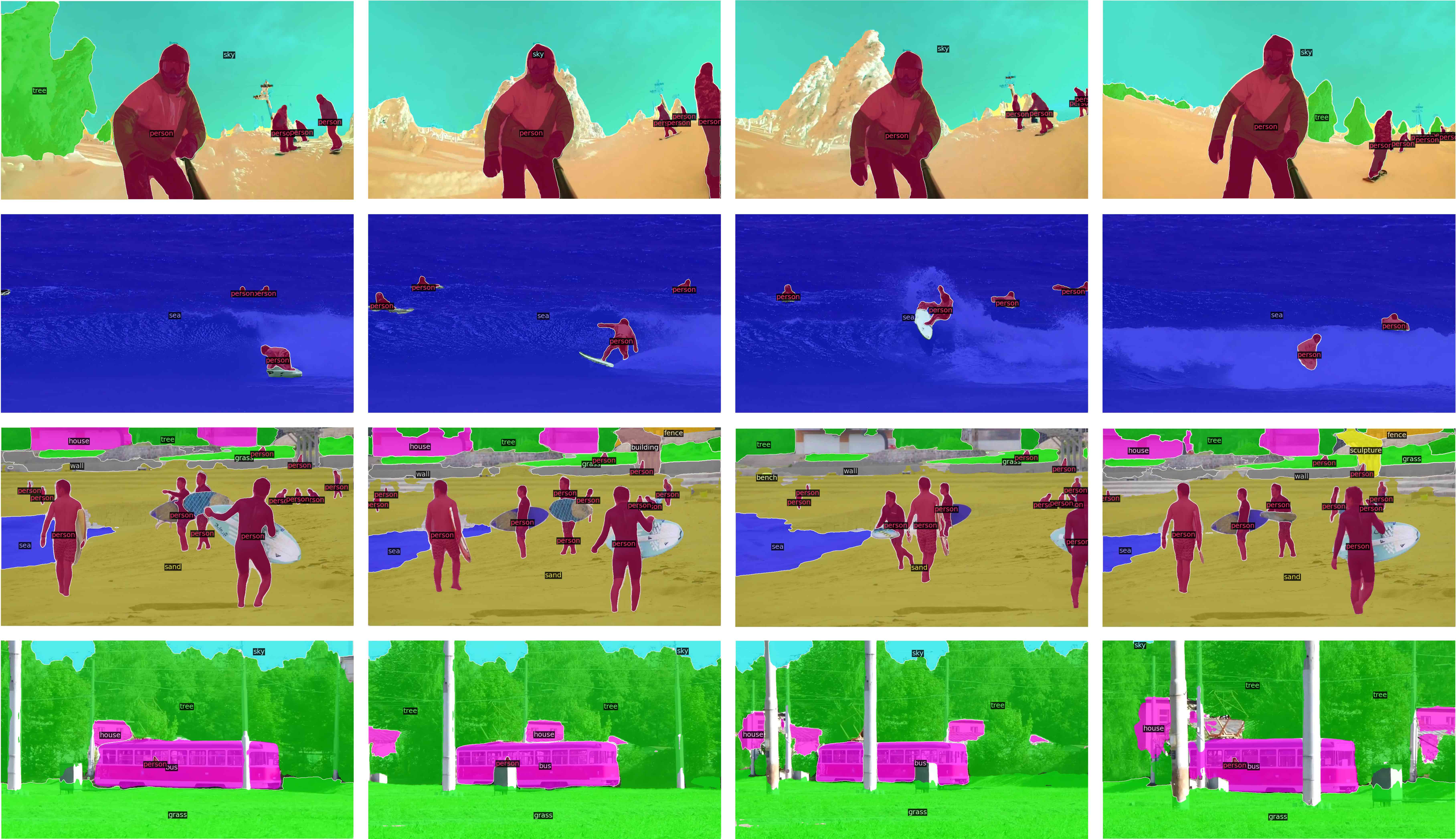}
    \vspace{2pt}
    \caption{Zero-Shot Video Generic Segmentation. (Source: YoutubeVOS videos)}
    \label{fig:generic_seg}
\end{figure*}

Open-vocabulary generic segmentation is one of the main advantages of \ourmodel{}. We also apply generic segmentation in a zero-shot manner to the YoutubeVOS~\cite{xu2018youtube} dataset. As shown in Fig.~\ref{fig:generic_seg}, our model can be well generalized to video zero-shot generic segmentation and make predictions that are consistent across frames. As a result, our model can be used in video segmentation directly or a good initialization for further finetuning.

\subsection{Zero-Shot Referring Video Segmentation}
\begin{figure*}[h]
    \centering
    \includegraphics[width=.95\linewidth]{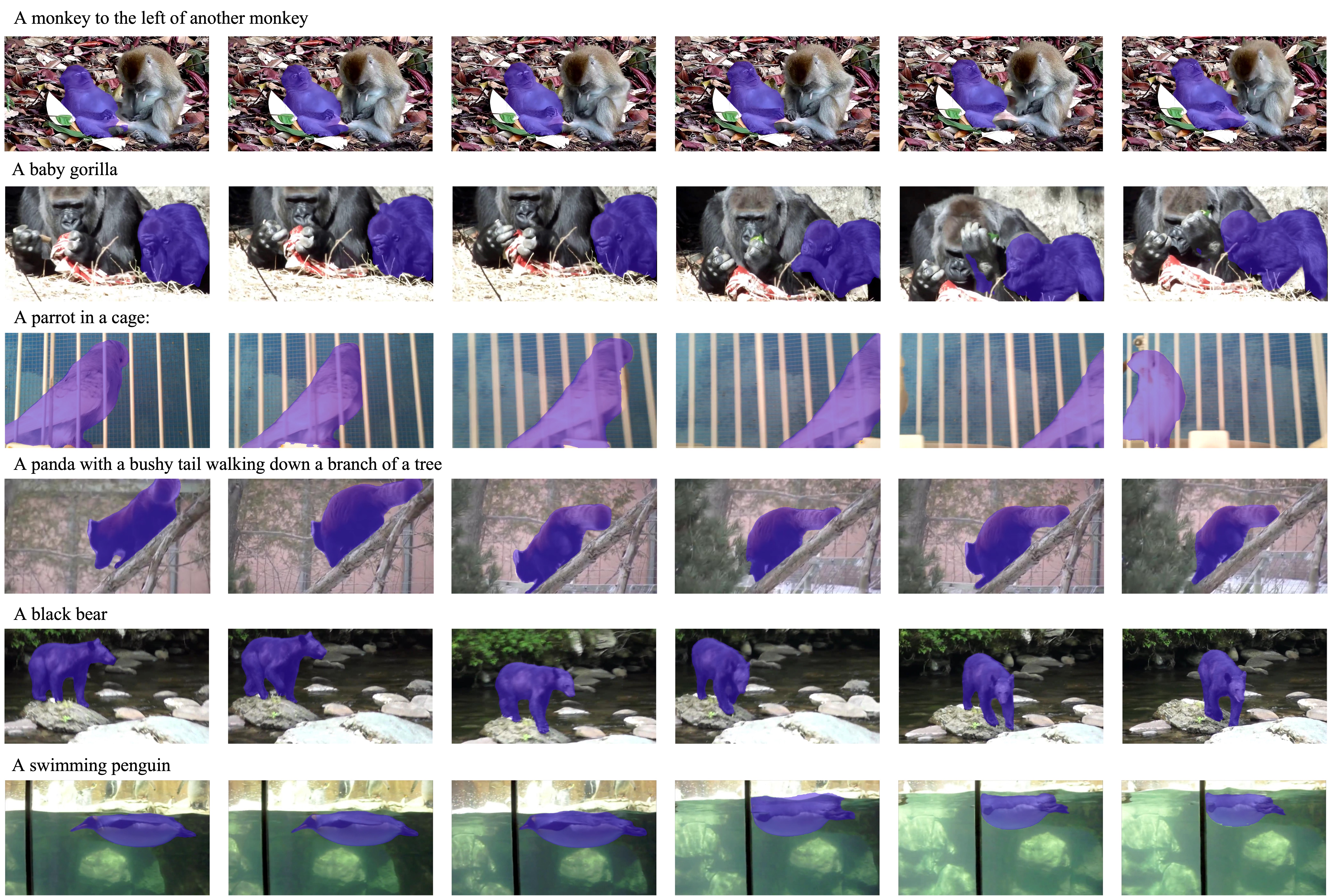}
    \caption{Zero-Shot Referring Video Segmentation. (Source: YoutubeVOS videos)}
    \label{fig:refer_video_seg}
\end{figure*}

Besides the generic segmentaton on video frames, our \ourmodel{} can be easily adapted to referring video segmentation as well without any architectural change or finetuning. In Fig.~\ref{fig:refer_video_seg}, we visualize some examples of referring video segmentation on the YoutubeVOS~\cite{xu2018youtube} dataset in a zero-shot manner. We can see that our model can generate rather accurate outputs given various referring phrases. Notably, in addition to the strong segmentation performance for given concepts, the model can also correctly distinguish the spatial locations (\textit{e.g.}, left \textit{v.s.} right in the first row), and object attributes (\textit{e.g.}, a baby gorilla instead of an adult gorilla in the second row) in these unseen videos.

\subsection{Zero-Shot Image Captioning}
\begin{figure*}[h]
    \centering
    \includegraphics[width=.95\linewidth]{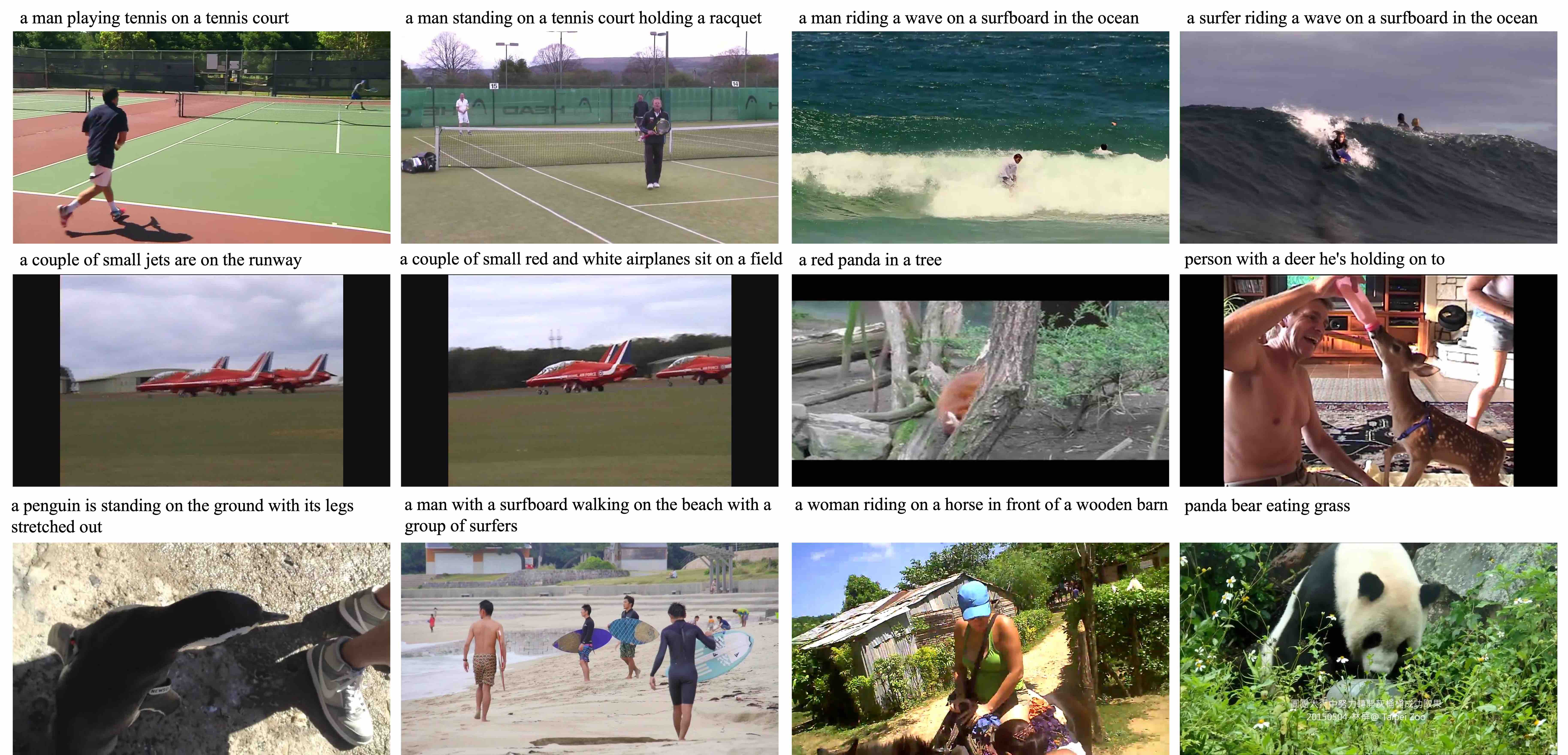}
    \caption{Zero-Shot Image Captioning. (Source: YoutubeVOS videos)}
    \label{fig:image_captioning}
\end{figure*}

To test the generalization ability of \ourmodel{}, we also ask the model generate image captions on the YoutubeVOS~\cite{xu2018youtube} dataset, which is in a different domain from the image data. As we can see from the examples in Fig.~\ref{fig:image_captioning}, the model can correctly predict the object, activity, and environment in an image. Interestingly, the captions for the first 6 images sampled from 3 different videos show that our approach can correctly differentiate the movements from similar scenarios (\textit{e.g.}, a man playing vs. a man standing in the first two samples.).

\subsection{Zero-Shot Referring Captioning}
\begin{figure*}[h]
    \centering
    \includegraphics[width=.95\linewidth]{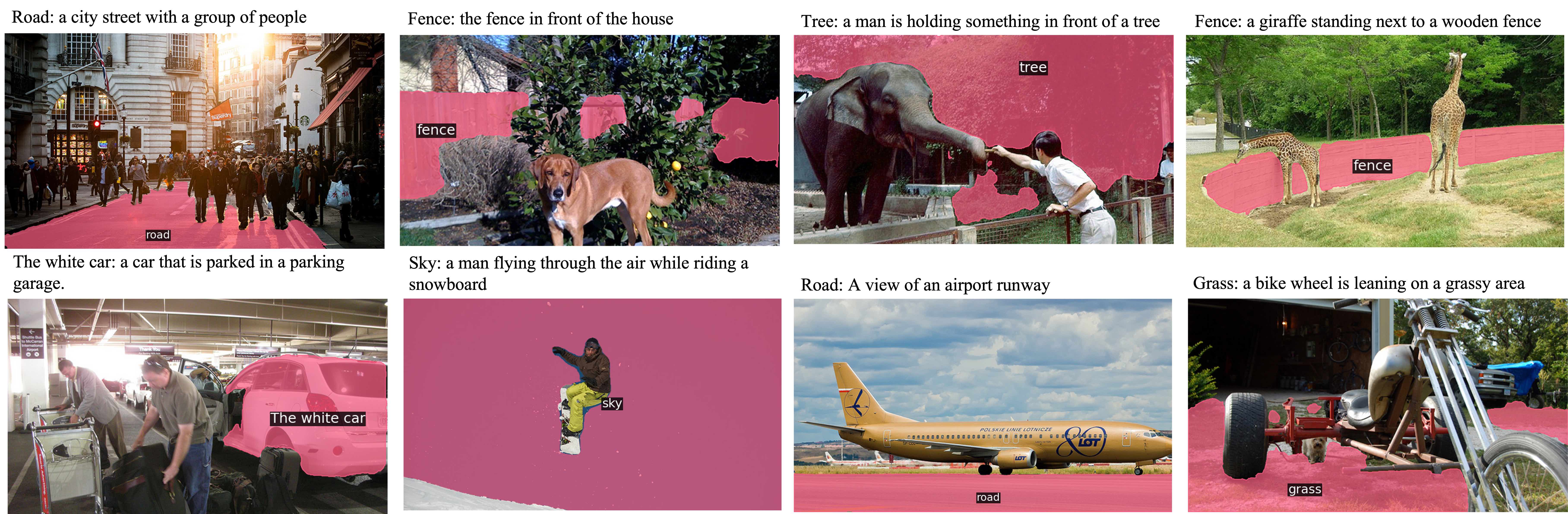}
    \caption{Referring Captioning. (Source: COCO 2017 val images)}
    \label{fig:ref_captioning}
\end{figure*}

In compensating for the visualization of the main paper, we add more referring captioning samples in Fig~\ref{fig:ref_captioning}. The phrase before ``:" is the referring phrase, and the sentence after ``:" is the generated caption. The grounding mask of the referring phrase is highlighted in pink. Clearly, our model can simultaneously segments the referred region and generates a region-specific caption. Complementary to regular image captioning systems, such a novel functionality provides a way of interpreting images in a more fine-grained manner. Note that our \ourmodel{} was never trained to generate such regional captions.

\subsection{Zero-Shot Referring Image Editing}

\begin{figure*}[h]
    \centering
    \includegraphics[width=.95\linewidth]{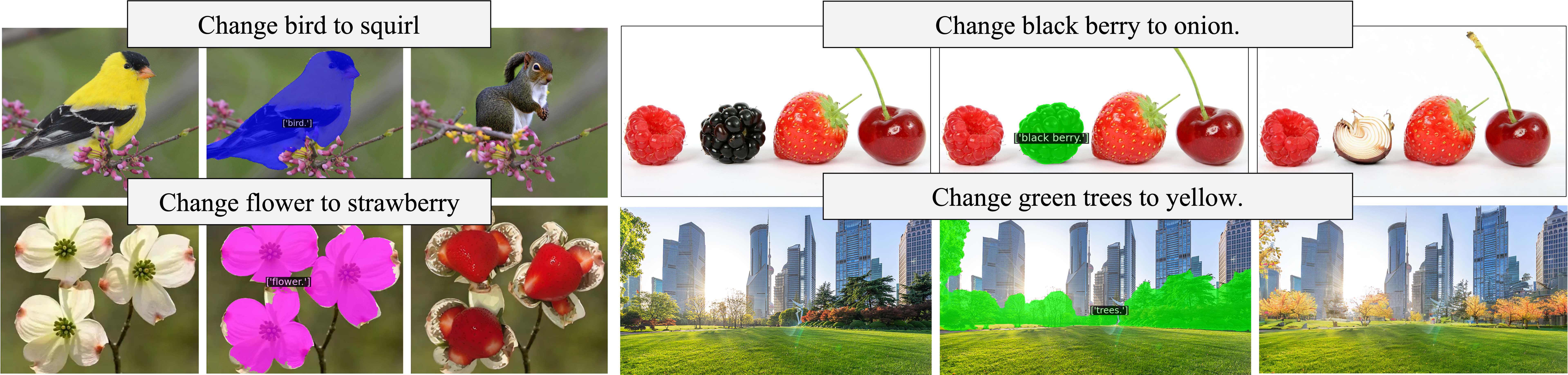}
    \caption{Referring Image Inpainting. (Source: web images)}
    \label{fig:ref_image_inpainting}
\end{figure*}

Finally, given the high-quality referring segmentation results with \ourmodel{}, we can effortlessly combine it with off-the-shelf Stable-Diffusion image inpainting model~\cite{rombach2022high} and perform zero-shot referring image editing. As shown in Fig.~\ref{fig:ref_image_inpainting}, the model first performs referring segmentation, then the original image and the segmentation mask are fed into the inpainting model to generate the inpainted image. For example, given ``change bird to squirrel'', it first extracts the bird segment (blue region) from the input image and then replace the segmented region with a generated squirrel. Likewise in other samples, we can see all the generated images look natural and follow the inpainting instructions very well. These impressive plug-and-play results imply a great potential of combining our \ourmodel{} and advanced generative AI models for fine-grained precise image editing.

\section{Discussions}

\textbf{Future Directions}. The extensive quantitative and qualitative results have demonstrated the strong performance and generalization ability of our \ourmodel{} for a variety of vision and vision-language tasks at different granularities. Upon the current \ourmodel{} design, we see two directions worth future explorations: (1) \textit{Pretrain the whole model in one stage effectively and efficiently}. Currently, the model still requires a separate pretraining for the image and text encoders. However, since our model supports large-scale image-text contrastive learning thanks to the decoupled design, we can easily unify the CLIP-style pretraining with the decoder pretraining in an end-to-end manner. (2) \textit{Unify all level of supervisions}. Due to high annotation costs, the pixel-level segmentation annotations by nature are much less than the region-level box and image-level annotations. It is worth building a more unified learning paradigm to jointly learn from pixel-level, region-level and image-level supervision to attain a more powerful unified model.

\textbf{Social Impact}. This work is mainly focused on the design of a generalized decoder for various vision and vision-language tasks. We have used a pretrained image and text encoder and further pretrained the models on a combination of various datasets and tasks. Since the models are trained on large-scale webly-crawled image-text pairs, the negative impact might arise due to the potential offensive or biased content in the data. To mitigate this issue, we need to have a careful sanity check on the training data and model predictions before deploying it in practical scenarios.

\begin{table*}
\footnotesize \setlength{\tabcolsep}{1.0pt}
\centering
\resizebox{1.0\linewidth}{!}{
\begin{tabular}{lcc|c|ccccccccccccccccccccccccc} 
\hline
Model       & Shot  & \#Param   & Avg  & \begin{tabular}[c]{@{}c@{}}Airplane-\\Parts\end{tabular} & Bottles                                               & \begin{tabular}[c]{@{}c@{}}Brain-\\Tumor\end{tabular} & Chicken                                                & Cows                                                  & \begin{tabular}[c]{@{}c@{}}Electric-\\Shaver\end{tabular} & Elephants                                             & Fruits                                                & Garbage                                                & \begin{tabular}[c]{@{}c@{}}Ginger-\\Garlic\end{tabular} & Hand                                                   & \begin{tabular}[c]{@{}c@{}}Hand-\\Metal\end{tabular}   & \begin{tabular}[c]{@{}c@{}}House-\\Parts\end{tabular} & \begin{tabular}[c]{@{}c@{}}HH.-\\Items\end{tabular} & \begin{tabular}[c]{@{}c@{}}Nutterfly-\\Squireel\end{tabular} & Phones                                                & Poles                                                 & Puppies                                               & Rail                                                  & \begin{tabular}[c]{@{}c@{}}Salmon-\\Fillet\end{tabular} & Strawberry                                             & Tablets                                                & Toolkits                                              & Trash                                                  & Watermelon                                             \\ 
\hline
X-Decoder (T)      & 0 & 0.0M & 22.7 & 10.5 & 19.0 & 1.1 & 12.0 & 12.0 & 1.2  & 65.6 & 66.5 & 28.7 & 7.9  & 0.6  & 22.4 & 5.5 & 50.6 & 62.1 & 29.9 & 3.6  & 48.9 & 0.7 & 15.0 & 41.6 & 15.2 & 9.5  & 19.3 & 16.2  \\
X-Decoder (T)      & 1   & 39.3M & 22.7 & 10.5{\tiny $\pm$0.0} & 19.0{\tiny $\pm$0.0} & 1.1{\tiny $\pm$0.0} & 12.0{\tiny $\pm$0.0}  & 12.0{\tiny $\pm$0.0} & 0.8{\tiny $\pm$0.7}   & 65.6{\tiny $\pm$0.0} & 66.5{\tiny $\pm$0.0} & 28.7{\tiny $\pm$0.0}  & 7.9{\tiny $\pm$5.8}   & 0.6{\tiny $\pm$0.0}   & 22.4{\tiny $\pm$0.0}  & 5.5{\tiny $\pm$0.0} & 50.6{\tiny $\pm$4.6} & 62.1{\tiny $\pm$0.0} & 29.9{\tiny $\pm$2.3} & 3.6{\tiny $\pm$0.0}  & 48.9{\tiny $\pm$4.6} & 0.7{\tiny $\pm$0.0}  & 15.0{\tiny $\pm$1.1} & 41.6{\tiny $\pm$0.0}  & 15.2{\tiny $\pm$0.0}  & 9.5{\tiny $\pm$0.0}  & 19.3{\tiny $\pm$0.0}  & 16.2{\tiny $\pm$0.0}  \\
X-Decoder (T)      & 3   & 39.3M & 24.1 & 10.5{\tiny $\pm$0.0} & 19.0{\tiny $\pm$0.0} & 1.1{\tiny $\pm$0.0} & 27.2{\tiny $\pm$26.3} & 13.4{\tiny $\pm$0.6} & 1.2{\tiny $\pm$0.0}   & 65.6{\tiny $\pm$0.0} & 66.9{\tiny $\pm$0.6} & 28.7{\tiny $\pm$0.0}  & 9.9{\tiny $\pm$2.5}   & 0.6{\tiny $\pm$0.0}   & 21.2{\tiny $\pm$2.2}  & 6.1{\tiny $\pm$0.1} & 50.6{\tiny $\pm$4.6} & 62.1{\tiny $\pm$0.0} & 36.6{\tiny $\pm$7.5} & 7.6{\tiny $\pm$8.0}  & 49.8{\tiny $\pm$2.5} & 0.7{\tiny $\pm$0.0}  & 15.0{\tiny $\pm$1.1} & 41.0{\tiny $\pm$1.0}  & 14.3{\tiny $\pm$1.5}  & 11.5{\tiny $\pm$0.6} & 19.8{\tiny $\pm$0.7}  & 20.3{\tiny $\pm$5.7}  \\
X-Decoder (T)      & 5   & 39.3M & 29.1 & 10.0{\tiny $\pm$0.9} & 30.2{\tiny $\pm$1.2} & 6.3{\tiny $\pm$3.9} & 51.8{\tiny $\pm$6.6}  & 18.2{\tiny $\pm$2.7} & 2.3{\tiny $\pm$1.5}   & 64.9{\tiny $\pm$0.8} & 67.2{\tiny $\pm$2.1} & 33.2{\tiny $\pm$1.2}  & 16.9{\tiny $\pm$3.3}  & 30.3{\tiny $\pm$22.8} & 23.8{\tiny $\pm$6.4}  & 7.1{\tiny $\pm$1.2} & 50.6{\tiny $\pm$0.1} & 66.4{\tiny $\pm$1.1} & 46.4{\tiny $\pm$5.9} & 14.0{\tiny $\pm$4.8} & 49.0{\tiny $\pm$0.4} & 1.5{\tiny $\pm$0.9}  & 15.7{\tiny $\pm$4.0} & 42.0{\tiny $\pm$0.7}  & 16.1{\tiny $\pm$3.2}  & 12.7{\tiny $\pm$1.6} & 21.1{\tiny $\pm$1.0}  & 27.9{\tiny $\pm$6.7}  \\
X-Decoder (T)      & 10  & 39.3M & 34.6 & 10.0{\tiny $\pm$2.0} & 34.1{\tiny $\pm$6.3} & 9.7{\tiny $\pm$2.3} & 59.6{\tiny $\pm$2.4}  & 19.8{\tiny $\pm$5.8} & 17.4{\tiny $\pm$23.3} & 64.8{\tiny $\pm$2.1} & 68.2{\tiny $\pm$7.1} & 38.1{\tiny $\pm$1.9}  & 17.6{\tiny $\pm$5.9}  & 81.6{\tiny $\pm$4.0}  & 45.6{\tiny $\pm$7.2}  & 8.4{\tiny $\pm$0.9} & 51.0{\tiny $\pm$0.4} & 63.9{\tiny $\pm$3.1} & 41.3{\tiny $\pm$8.4} & 3.6{\tiny $\pm$0.0}  & 48.4{\tiny $\pm$2.4} & 1.7{\tiny $\pm$1.5}  & 29.7{\tiny $\pm$7.0} & 44.4{\tiny $\pm$3.0}  & 24.1{\tiny $\pm$5.7}  & 14.1{\tiny $\pm$2.1} & 21.6{\tiny $\pm$1.1}  & 44.5{\tiny $\pm$2.4}  \\
X-Decoder (T)      & 3$\times$All & 39.3M & 37.5 & 10.5{\tiny $\pm$0.5} & 35.1{\tiny $\pm$4.4} & 6.6{\tiny $\pm$1.2} & 14.5{\tiny $\pm$4.3}  & 30.9{\tiny $\pm$0.9} & 7.1{\tiny $\pm$0.5}   & 68.3{\tiny $\pm$0.4} & 70.3{\tiny $\pm$6.9} & 36.9{\tiny $\pm$0.8}  & 9.9{\tiny $\pm$3.5}   & 37.7{\tiny $\pm$20.1} & 46.0{\tiny $\pm$3.7}  & 8.5{\tiny $\pm$0.0} & 50.6{\tiny $\pm$0.0} & 79.8{\tiny $\pm$0.3} & 48.7{\tiny $\pm$0.8} & 13.1{\tiny $\pm$9.4} & 50.6{\tiny $\pm$0.2} & 59.6{\tiny $\pm$0.9} & 54.2{\tiny $\pm$0.5} & 83.4{\tiny $\pm$10.5} & 27.4{\tiny $\pm$0.8}  & 13.2{\tiny $\pm$1.0} & 27.6{\tiny $\pm$0.9}  & 44.5{\tiny $\pm$0.9}  \\ 
\hline
X-Decoder-Seg$^+$ (B) & 0 & 0.0M & 26.3 & 13.2 & 17.2 & 0.8 & 33.0 & 28.6 & 4.9  & 67.9 & 71.1 & 28.8 & 5.2  & 0.0  & 0.8  & 6.8 & 50.6 & 53.2 & 18.8 & 17.9 & 68.2 & 0.7 & 21.1 & 86.3 & 5.8  & 11.5 & 12.1 & 31.7  \\
X-Decoder-Seg$^+$  (B) & 1   & 39.3M & 26.3 & 13.2{\tiny $\pm$0.0} & 17.2{\tiny $\pm$0.0} & 0.8{\tiny $\pm$0.0} & 33.0{\tiny $\pm$0.0}  & 28.6{\tiny $\pm$2.3} & 4.9{\tiny $\pm$0.0}   & 68.0{\tiny $\pm$0.2} & 71.1{\tiny $\pm$0.0} & 28.8{\tiny $\pm$0.0}  & 5.2{\tiny $\pm$0.0}   & 0.0{\tiny $\pm$0.0}   & 0.8{\tiny $\pm$0.0}   & 6.8{\tiny $\pm$0.0} & 50.7{\tiny $\pm$0.0} & 53.2{\tiny $\pm$0.0} & 18.8{\tiny $\pm$0.0} & 17.9{\tiny $\pm$0.0} & 68.2{\tiny $\pm$0.0} & 0.7{\tiny $\pm$0.0}  & 21.0{\tiny $\pm$0.0} & 86.5{\tiny $\pm$0.3}  & 5.8{\tiny $\pm$5.8}   & 11.5{\tiny $\pm$0.0} & 12.1{\tiny $\pm$1.1}  & 31.7{\tiny $\pm$2.3}  \\
X-Decoder-Seg$^+$  (B) & 3   & 39.3M & 26.6 & 13.2{\tiny $\pm$0.0} & 17.2{\tiny $\pm$0.2} & 0.9{\tiny $\pm$0.1} & 33.0{\tiny $\pm$0.0}  & 29.6{\tiny $\pm$0.6} & 4.4{\tiny $\pm$0.8}   & 67.3{\tiny $\pm$0.5} & 74.4{\tiny $\pm$4.9} & 28.6{\tiny $\pm$0.3}  & 6.0{\tiny $\pm$0.8}   & 0.0{\tiny $\pm$0.0}   & 1.1{\tiny $\pm$0.2}   & 7.0{\tiny $\pm$0.2} & 50.7{\tiny $\pm$0.0} & 53.1{\tiny $\pm$0.1} & 23.3{\tiny $\pm$3.3} & 17.9{\tiny $\pm$0.0} & 67.1{\tiny $\pm$1.9} & 0.7{\tiny $\pm$0.0}  & 21.4{\tiny $\pm$0.4} & 86.5{\tiny $\pm$0.7}  & 5.8{\tiny $\pm$0.0}   & 10.3{\tiny $\pm$1.0} & 12.2{\tiny $\pm$0.0}  & 32.9{\tiny $\pm$1.0}  \\
X-Decoder-Seg$^+$  (B) & 5   & 39.3M & 27.5 & 13.3{\tiny $\pm$0.1} & 18.6{\tiny $\pm$1.3} & 1.4{\tiny $\pm$0.1} & 34.3{\tiny $\pm$3.5}  & 30.6{\tiny $\pm$1.1} & 5.2{\tiny $\pm$0.7}   & 67.6{\tiny $\pm$2.2} & 79.5{\tiny $\pm$1.0} & 28.8{\tiny $\pm$0.1}  & 4.3{\tiny $\pm$0.8}   & 0.0{\tiny $\pm$0.0}   & 1.4{\tiny $\pm$0.3}   & 7.5{\tiny $\pm$0.1} & 50.7{\tiny $\pm$0.0} & 54.9{\tiny $\pm$0.9} & 26.9{\tiny $\pm$4.1} & 18.8{\tiny $\pm$1.1} & 66.9{\tiny $\pm$3.0} & 0.8{\tiny $\pm$0.1}  & 21.6{\tiny $\pm$0.9} & 85.5{\tiny $\pm$2.1}  & 7.4{\tiny $\pm$0.1}   & 13.0{\tiny $\pm$2.7} & 12.4{\tiny $\pm$0.0}  & 35.7{\tiny $\pm$1.8}  \\
X-Decoder-Seg$^+$  (B) & 10  & 39.3M & 30.0 & 13.5{\tiny $\pm$0.3} & 18.0{\tiny $\pm$0.2} & 2.1{\tiny $\pm$0.8} & 49.7{\tiny $\pm$1.9}  & 35.9{\tiny $\pm$1.8} & 8.0{\tiny $\pm$3.1}   & 68.2{\tiny $\pm$2.8} & 79.9{\tiny $\pm$4.0} & 28.6{\tiny $\pm$0.3}  & 7.3{\tiny $\pm$2.8}   & 0.0{\tiny $\pm$0.0}   & 6.0{\tiny $\pm$1.6}   & 8.1{\tiny $\pm$0.3} & 50.5{\tiny $\pm$0.0} & 55.1{\tiny $\pm$2.4} & 33.6{\tiny $\pm$1.7} & 17.9{\tiny $\pm$0.0} & 69.4{\tiny $\pm$3.9} & 1.6{\tiny $\pm$0.3}  & 24.8{\tiny $\pm$2.5} & 87.9{\tiny $\pm$0.6}  & 9.0{\tiny $\pm$1.3}   & 14.3{\tiny $\pm$0.6} & 13.0{\tiny $\pm$0.4}  & 45.7{\tiny $\pm$2.8}  \\
X-Decoder-Seg$^+$  (B) & 3$\times$All  & 39.3M  & 31.9 & 13.5{\tiny $\pm$0.2} & 19.8{\tiny $\pm$0.4} & 2.2{\tiny $\pm$0.1} & 31.7{\tiny $\pm$2.3}  & 37.7{\tiny $\pm$0.3} & 4.1{\tiny $\pm$0.3}   & 73.9{\tiny $\pm$0.2} & 78.8{\tiny $\pm$1.9} & 29.0{\tiny $\pm$0.0}  & 5.3{\tiny $\pm$0.2}   & 0.0{\tiny $\pm$0.0}   & 3.3{\tiny $\pm$0.6}   & 8.6{\tiny $\pm$0.1} & 50.7{\tiny $\pm$0.0} & 63.9{\tiny $\pm$0.0} & 24.3{\tiny $\pm$1.1} & 20.1{\tiny $\pm$0.0} & 65.4{\tiny $\pm$0.0} & 46.4{\tiny $\pm$1.5} & 54.0{\tiny $\pm$1.6} & 89.6{\tiny $\pm$0.0}  & 8.4{\tiny $\pm$0.8}   & 12.7{\tiny $\pm$0.9} & 13.7{\tiny $\pm$0.1}  & 39.1{\tiny $\pm$1.0}  \\ 
\hline
X-Decoder (B)     & 0 & 0.0M & 27.7 & 13.0 & 45.9 & 0.3 & 13.6 & 36.8 & 4.2  & 68.0 & 76.7 & 30.2 & 19.4 & 20.6 & 18.5 & 6.7 & 51.7 & 53.1 & 8.9  & 5.6  & 55.4 & 0.8 & 18.2 & 81.6 & 8.0  & 13.9 & 27.3 & 13.0  \\
X-Decoder (B)     & 1   & 39.3M & 27.8 & 13.0{\tiny $\pm$0.0} & 45.9{\tiny $\pm$0.0} & 0.3{\tiny $\pm$0.0} & 13.6{\tiny $\pm$1.1}  & 36.8{\tiny $\pm$0.0} & 4.2{\tiny $\pm$0.0}   & 68.0{\tiny $\pm$0.0} & 76.7{\tiny $\pm$0.0} & 30.2{\tiny $\pm$0.0}  & 19.4{\tiny $\pm$0.0}  & 20.6{\tiny $\pm$0.0}  & 18.5{\tiny $\pm$0.0}  & 6.7{\tiny $\pm$5.8} & 51.7{\tiny $\pm$0.0} & 53.1{\tiny $\pm$4.6} & 10.2{\tiny $\pm$2.1} & 5.6{\tiny $\pm$0.0}  & 55.4{\tiny $\pm$0.0} & 0.8{\tiny $\pm$0.0}  & 18.2{\tiny $\pm$0.0} & 81.6{\tiny $\pm$0.0}  & 8.0{\tiny $\pm$0.0}   & 13.9{\tiny $\pm$0.0} & 27.3{\tiny $\pm$0.0}  & 13.0{\tiny $\pm$0.0}  \\
X-Decoder (B)     & 3   & 39.3M & 28.2 & 12.6{\tiny $\pm$0.3} & 41.6{\tiny $\pm$5.2} & 0.3{\tiny $\pm$0.0} & 13.6{\tiny $\pm$1.1}  & 37.2{\tiny $\pm$0.6} & 4.2{\tiny $\pm$0.0}   & 68.0{\tiny $\pm$0.1} & 77.4{\tiny $\pm$1.1} & 30.4{\tiny $\pm$0.2}  & 19.8{\tiny $\pm$0.6}  & 26.7{\tiny $\pm$9.6}  & 22.2{\tiny $\pm$6.4}  & 7.2{\tiny $\pm$0.2} & 51.9{\tiny $\pm$0.3} & 53.1{\tiny $\pm$4.6} & 12.2{\tiny $\pm$2.8} & 5.5{\tiny $\pm$0.4}  & 55.4{\tiny $\pm$0.0} & 0.3{\tiny $\pm$0.5}  & 19.4{\tiny $\pm$3.4} & 82.9{\tiny $\pm$1.5}  & 8.7{\tiny $\pm$0.7}   & 10.9{\tiny $\pm$3.1} & 28.0{\tiny $\pm$0.6}  & 15.3{\tiny $\pm$2.0}  \\
X-Decoder (B)     & 5   & 39.3M & 33.1 & 12.6{\tiny $\pm$0.4} & 41.2{\tiny $\pm$0.9} & 2.2{\tiny $\pm$2.0} & 27.5{\tiny $\pm$7.5}  & 37.9{\tiny $\pm$0.3} & 9.8{\tiny $\pm$4.6}   & 67.6{\tiny $\pm$1.2} & 78.3{\tiny $\pm$1.7} & 30.4{\tiny $\pm$0.3}  & 33.9{\tiny $\pm$5.8}  & 75.2{\tiny $\pm$14.1} & 35.5{\tiny $\pm$16.5} & 8.5{\tiny $\pm$0.4} & 53.0{\tiny $\pm$1.8} & 61.9{\tiny $\pm$7.8} & 12.3{\tiny $\pm$2.7} & 4.5{\tiny $\pm$0.5}  & 54.7{\tiny $\pm$0.8} & 0.9{\tiny $\pm$0.1}  & 20.7{\tiny $\pm$0.4} & 86.7{\tiny $\pm$0.8}  & 8.8{\tiny $\pm$2.5}   & 14.9{\tiny $\pm$6.8} & 28.9{\tiny $\pm$1.3}  & 19.4{\tiny $\pm$6.8}  \\
X-Decoder (B)     & 10  & 39.3M & 38.6 & 10.9{\tiny $\pm$1.0} & 38.3{\tiny $\pm$3.5} & 1.8{\tiny $\pm$0.9} & 46.3{\tiny $\pm$4.2}  & 38.6{\tiny $\pm$1.8} & 24.5{\tiny $\pm$10.3} & 67.0{\tiny $\pm$4.4} & 78.4{\tiny $\pm$3.3} & 30.8{\tiny $\pm$0.4}  & 37.9{\tiny $\pm$2.1}  & 89.1{\tiny $\pm$2.9}  & 63.5{\tiny $\pm$1.0}  & 8.7{\tiny $\pm$0.5} & 58.4{\tiny $\pm$2.9} & 72.8{\tiny $\pm$3.0} & 19.5{\tiny $\pm$7.4} & 5.6{\tiny $\pm$0.0}  & 57.1{\tiny $\pm$1.1} & 2.3{\tiny $\pm$0.7}  & 24.6{\tiny $\pm$2.4} & 87.8{\tiny $\pm$2.8}  & 13.4{\tiny $\pm$5.0}  & 15.0{\tiny $\pm$3.3} & 32.6{\tiny $\pm$0.4}  & 39.9{\tiny $\pm$3.1}  \\
X-Decoder (B)     & 3$\times$All & 39.3M & 38.9 & 12.5{\tiny $\pm$0.0} & 42.7{\tiny $\pm$0.2} & 1.0{\tiny $\pm$0.0} & 14.6{\tiny $\pm$1.4}  & 36.8{\tiny $\pm$0.3} & 17.4{\tiny $\pm$3.1}  & 71.7{\tiny $\pm$0.2} & 79.7{\tiny $\pm$0.6} & 31.5{\tiny $\pm$0.1}  & 29.1{\tiny $\pm$1.6}  & 53.6{\tiny $\pm$0.5}  & 65.8{\tiny $\pm$0.5}  & 9.2{\tiny $\pm$0.0} & 54.0{\tiny $\pm$0.9} & 82.3{\tiny $\pm$0.2} & 17.1{\tiny $\pm$3.2} & 5.7{\tiny $\pm$1.2}  & 55.4{\tiny $\pm$0.4} & 48.9{\tiny $\pm$2.7} & 48.3{\tiny $\pm$1.4} & 90.3{\tiny $\pm$0.0}  & 18.8{\tiny $\pm$2.5}  & 15.0{\tiny $\pm$0.0} & 36.3{\tiny $\pm$0.5}  & 33.0{\tiny $\pm$2.1}  \\ 
\hline
X-Decoder (L-IN21K)      & 0 & 0.0M & 26.7 & 12.3 & 43.2 & 0.5 & 3.5  & 12.3 & 18.8 & 63.9 & 79.1 & 24.3 & 15.6 & 0.0  & 20.3 & 4.9 & 50.5 & 58.8 & 43.4 & 13.4 & 57.3 & 1.3 & 12.3 & 74.4 & 6.9  & 14.6 & 20.1 & 13.5  \\
X-Decoder (L-IN21K)      & 1  & 39.3M & 26.8 & 12.3{\tiny $\pm$0.0} & 43.2{\tiny $\pm$4.6} & 0.5{\tiny $\pm$0.0} & 3.5{\tiny $\pm$0.0}   & 13.9{\tiny $\pm$2.8} & 18.8{\tiny $\pm$0.0}  & 63.9{\tiny $\pm$4.6} & 79.1{\tiny $\pm$0.0} & 25.1{\tiny $\pm$1.4}  & 15.6{\tiny $\pm$1.1}  & 0.0{\tiny $\pm$0.0}   & 21.4{\tiny $\pm$1.9}  & 4.9{\tiny $\pm$0.0} & 50.5{\tiny $\pm$0.0} & 58.8{\tiny $\pm$0.0} & 43.4{\tiny $\pm$0.0} & 14.0{\tiny $\pm$0.9} & 57.3{\tiny $\pm$0.0} & 1.3{\tiny $\pm$0.0}  & 12.3{\tiny $\pm$0.0} & 74.4{\tiny $\pm$0.0}  & 6.9{\tiny $\pm$0.0}   & 14.6{\tiny $\pm$0.0} & 20.1{\tiny $\pm$0.0}  & 13.5{\tiny $\pm$0.0}  \\
X-Decoder (L-IN21K)      & 3  & 39.3M & 29.5 & 14.0{\tiny $\pm$0.4} & 44.9{\tiny $\pm$3.0} & 0.9{\tiny $\pm$0.7} & 3.5{\tiny $\pm$1.3}   & 27.0{\tiny $\pm$1.0} & 21.6{\tiny $\pm$4.9}  & 63.7{\tiny $\pm$0.8} & 79.1{\tiny $\pm$0.0} & 24.3{\tiny $\pm$0.0}  & 13.8{\tiny $\pm$4.2}  & 29.6{\tiny $\pm$51.3} & 16.9{\tiny $\pm$5.8}  & 6.0{\tiny $\pm$0.4} & 50.6{\tiny $\pm$0.1} & 59.6{\tiny $\pm$1.3} & 45.2{\tiny $\pm$0.7} & 18.4{\tiny $\pm$1.7} & 57.7{\tiny $\pm$1.4} & 0.5{\tiny $\pm$0.9}  & 16.7{\tiny $\pm$7.7} & 83.1{\tiny $\pm$1.8}  & 5.4{\tiny $\pm$3.0}   & 15.3{\tiny $\pm$0.9} & 23.3{\tiny $\pm$3.1}  & 14.7{\tiny $\pm$2.0}  \\
X-Decoder (L-IN21K)      & 5  & 39.3M & 36.2 & 12.1{\tiny $\pm$1.6} & 50.4{\tiny $\pm$4.3} & 0.4{\tiny $\pm$0.0} & 31.7{\tiny $\pm$6.9}  & 32.7{\tiny $\pm$0.7} & 51.9{\tiny $\pm$18.6} & 64.2{\tiny $\pm$0.7} & 75.7{\tiny $\pm$3.9} & 27.8{\tiny $\pm$0.7}  & 22.4{\tiny $\pm$13.4} & 60.0{\tiny $\pm$33.8} & 23.9{\tiny $\pm$10.0} & 7.1{\tiny $\pm$0.1} & 51.4{\tiny $\pm$1.0} & 63.0{\tiny $\pm$3.6} & 42.7{\tiny $\pm$2.9} & 15.7{\tiny $\pm$5.6} & 59.7{\tiny $\pm$1.7} & 1.8{\tiny $\pm$0.1}  & 21.4{\tiny $\pm$9.0} & 83.7{\tiny $\pm$2.5}  & 16.3{\tiny $\pm$11.6} & 16.5{\tiny $\pm$0.4} & 34.0{\tiny $\pm$4.7}  & 37.1{\tiny $\pm$2.7}  \\
X-Decoder (L-IN21K)      & 10 & 39.3M & 40.5 & 11.8{\tiny $\pm$0.4} & 52.0{\tiny $\pm$2.3} & 0.6{\tiny $\pm$0.2} & 34.1{\tiny $\pm$6.1}  & 34.3{\tiny $\pm$1.1} & 48.7{\tiny $\pm$16.6} & 65.3{\tiny $\pm$1.7} & 80.0{\tiny $\pm$0.9} & 30.4{\tiny $\pm$11.3} & 28.0{\tiny $\pm$10.8} & 91.5{\tiny $\pm$2.8}  & 47.4{\tiny $\pm$29.2} & 7.0{\tiny $\pm$0.5} & 54.2{\tiny $\pm$5.1} & 73.0{\tiny $\pm$6.9} & 44.6{\tiny $\pm$1.8} & 13.4{\tiny $\pm$0.0} & 55.0{\tiny $\pm$5.2} & 4.6{\tiny $\pm$1.3}  & 24.4{\tiny $\pm$3.6} & 85.3{\tiny $\pm$1.1}  & 24.7{\tiny $\pm$18.6} & 20.2{\tiny $\pm$1.3} & 37.0{\tiny $\pm$1.5}  & 43.8{\tiny $\pm$3.1}  \\
X-Decoder (L-IN21K)      & 3$\times$All & 39.3M & 40.7 & 14.1{\tiny $\pm$1.0} & 53.2{\tiny $\pm$0.7} & 0.4{\tiny $\pm$0.1} & 4.5{\tiny $\pm$1.7}   & 36.6{\tiny $\pm$0.4} & 62.7{\tiny $\pm$4.2}  & 70.3{\tiny $\pm$0.1} & 80.0{\tiny $\pm$1.2} & 31.7{\tiny $\pm$0.7}  & 15.6{\tiny $\pm$6.0}  & 22.2{\tiny $\pm$0.5}  & 61.7{\tiny $\pm$0.8}  & 7.8{\tiny $\pm$0.2} & 51.8{\tiny $\pm$1.0} & 85.2{\tiny $\pm$0.0} & 44.1{\tiny $\pm$0.3} & 12.6{\tiny $\pm$5.9} & 60.7{\tiny $\pm$0.0} & 41.7{\tiny $\pm$1.0} & 48.3{\tiny $\pm$0.6} & 90.4{\tiny $\pm$0.0}  & 22.9{\tiny $\pm$1.7}  & 18.7{\tiny $\pm$1.2} & 42.3{\tiny $\pm$0.6}  & 36.8{\tiny $\pm$2.4}  \\ 
\hline
X-Decoder (L)     & 0 & 0.0M & 32.3 & 13.1 & 42.1 & 2.2 & 8.6  & 44.9 & 7.5  & 66.0 & 79.2 & 33.0 & 11.6 & 75.9 & 42.1 & 7.0 & 53.0 & 68.4 & 15.6 & 20.1 & 59.0 & 2.3 & 19.0 & 67.1 & 22.5 & 9.9  & 22.3 & 13.8  \\
X-Decoder (L)     & 1   & 39.3M & 32.0 & 13.1{\tiny $\pm$0.0} & 42.1{\tiny $\pm$0.0} & 2.2{\tiny $\pm$0.0} & 8.6{\tiny $\pm$0.0}   & 44.9{\tiny $\pm$0.0} & 7.5{\tiny $\pm$0.0}   & 66.0{\tiny $\pm$0.0} & 79.2{\tiny $\pm$0.0} & 33.0{\tiny $\pm$0.0}  & 11.6{\tiny $\pm$1.1}  & 75.9{\tiny $\pm$0.0}  & 42.1{\tiny $\pm$0.0}  & 7.0{\tiny $\pm$0.0} & 53.0{\tiny $\pm$0.0} & 68.4{\tiny $\pm$0.0} & 15.6{\tiny $\pm$1.1} & 20.1{\tiny $\pm$0.0} & 59.0{\tiny $\pm$0.0} & 2.3{\tiny $\pm$0.0}  & 19.0{\tiny $\pm$0.0} & 67.1{\tiny $\pm$0.0}  & 22.5{\tiny $\pm$0.0}  & 9.9{\tiny $\pm$0.0}  & 15.4{\tiny $\pm$13.4} & 13.8{\tiny $\pm$0.0}  \\
X-Decoder (L)     & 3   & 39.3M & 32.6 & 13.1{\tiny $\pm$0.0} & 42.1{\tiny $\pm$0.0} & 2.2{\tiny $\pm$0.0} & 12.3{\tiny $\pm$6.4}  & 45.1{\tiny $\pm$0.2} & 7.5{\tiny $\pm$0.0}   & 66.0{\tiny $\pm$0.0} & 78.6{\tiny $\pm$0.5} & 33.3{\tiny $\pm$0.5}  & 11.6{\tiny $\pm$1.1}  & 75.9{\tiny $\pm$0.0}  & 42.1{\tiny $\pm$0.0}  & 7.0{\tiny $\pm$0.0} & 53.0{\tiny $\pm$0.0} & 68.4{\tiny $\pm$0.0} & 17.0{\tiny $\pm$5.2} & 21.6{\tiny $\pm$1.7} & 59.0{\tiny $\pm$0.0} & 2.4{\tiny $\pm$0.1}  & 19.0{\tiny $\pm$0.0} & 67.1{\tiny $\pm$0.0}  & 23.3{\tiny $\pm$1.4}  & 9.5{\tiny $\pm$0.7}  & 22.3{\tiny $\pm$0.0}  & 13.8{\tiny $\pm$0.0}  \\
X-Decoder (L)     & 5   & 39.3M & 35.0 & 14.0{\tiny $\pm$0.3} & 45.3{\tiny $\pm$3.6} & 4.1{\tiny $\pm$0.4} & 24.9{\tiny $\pm$11.0} & 46.1{\tiny $\pm$0.1} & 11.2{\tiny $\pm$7.1}  & 65.8{\tiny $\pm$0.8} & 77.9{\tiny $\pm$1.1} & 33.6{\tiny $\pm$0.5}  & 13.2{\tiny $\pm$1.6}  & 85.1{\tiny $\pm$4.1}  & 43.5{\tiny $\pm$4.1}  & 7.4{\tiny $\pm$0.0} & 52.9{\tiny $\pm$0.2} & 69.2{\tiny $\pm$1.4} & 16.9{\tiny $\pm$8.2} & 21.6{\tiny $\pm$1.6} & 58.5{\tiny $\pm$3.1} & 2.6{\tiny $\pm$0.2}  & 18.4{\tiny $\pm$0.9} & 81.2{\tiny $\pm$3.9}  & 25.8{\tiny $\pm$4.8}  & 9.7{\tiny $\pm$0.3}  & 24.9{\tiny $\pm$1.8}  & 19.6{\tiny $\pm$2.6}  \\
X-Decoder (L)     & 10  & 39.3M & 40.3 & 13.3{\tiny $\pm$0.3} & 45.2{\tiny $\pm$3.5} & 3.2{\tiny $\pm$1.7} & 42.3{\tiny $\pm$3.9}  & 45.8{\tiny $\pm$0.1} & 29.3{\tiny $\pm$3.5}  & 68.3{\tiny $\pm$2.0} & 76.0{\tiny $\pm$3.1} & 37.9{\tiny $\pm$1.9}  & 24.4{\tiny $\pm$1.3}  & 93.7{\tiny $\pm$0.4}  & 57.5{\tiny $\pm$1.2}  & 7.9{\tiny $\pm$0.5} & 52.1{\tiny $\pm$0.3} & 78.8{\tiny $\pm$1.3} & 27.0{\tiny $\pm$1.5} & 20.1{\tiny $\pm$0.0} & 56.7{\tiny $\pm$4.9} & 3.3{\tiny $\pm$0.3}  & 17.5{\tiny $\pm$0.9} & 85.2{\tiny $\pm$1.4}  & 40.1{\tiny $\pm$7.0}  & 8.4{\tiny $\pm$0.6}  & 31.4{\tiny $\pm$0.7}  & 42.0{\tiny $\pm$8.7}  \\
X-Decoder (L)     & 3$\times$All & 39.3M & 42.2 & 13.9{\tiny $\pm$0.5} & 48.4{\tiny $\pm$0.2} & 7.9{\tiny $\pm$2.8} & 8.6{\tiny $\pm$0.0}   & 45.3{\tiny $\pm$0.2} & 20.5{\tiny $\pm$0.2}  & 72.4{\tiny $\pm$0.0} & 80.5{\tiny $\pm$1.0} & 36.7{\tiny $\pm$1.1}  & 14.8{\tiny $\pm$1.4}  & 86.7{\tiny $\pm$1.8}  & 63.8{\tiny $\pm$0.3}  & 7.5{\tiny $\pm$0.2} & 52.8{\tiny $\pm$0.8} & 83.3{\tiny $\pm$0.1} & 20.1{\tiny $\pm$1.2} & 18.1{\tiny $\pm$6.6} & 57.4{\tiny $\pm$3.0} & 45.1{\tiny $\pm$0.9} & 50.2{\tiny $\pm$1.1} & 92.0{\tiny $\pm$0.1}  & 40.4{\tiny $\pm$1.0}  & 10.4{\tiny $\pm$0.7} & 36.3{\tiny $\pm$0.6}  & 40.2{\tiny $\pm$0.7}  \\
\hline
\end{tabular}
}
\vspace{3pt}
\caption{SegInW results with tuning on class embedding for different image shots and backbone architectures. (39.3M parameters tuned in the setting.)}
\label{tab:seginw_linear}
\end{table*}

\begin{table*}
\footnotesize \setlength{\tabcolsep}{1.0pt}
\centering
\resizebox{1.0\linewidth}{!}{
\begin{tabular}{lcc|c|ccccccccccccccccccccccccc} 
\hline
Model       & Shot  & \#Param   & Avg  & \begin{tabular}[c]{@{}c@{}}Airplane-\\Parts\end{tabular} & Bottles                                               & \begin{tabular}[c]{@{}c@{}}Brain-\\Tumor\end{tabular} & Chicken                                                & Cows                                                  & \begin{tabular}[c]{@{}c@{}}Electric-\\Shaver\end{tabular} & Elephants                                             & Fruits                                                & Garbage                                                & \begin{tabular}[c]{@{}c@{}}Ginger-\\Garlic\end{tabular} & Hand                                                   & \begin{tabular}[c]{@{}c@{}}Hand-\\Metal\end{tabular}   & \begin{tabular}[c]{@{}c@{}}House-\\Parts\end{tabular} & \begin{tabular}[c]{@{}c@{}}HH.-\\Items\end{tabular} & \begin{tabular}[c]{@{}c@{}}Nutterfly-\\Squireel\end{tabular} & Phones                                                & Poles                                                 & Puppies                                               & Rail                                                  & \begin{tabular}[c]{@{}c@{}}Salmon-\\Fillet\end{tabular} & Strawberry                                             & Tablets                                                & Toolkits                                              & Trash                                                  & Watermelon                                             \\ 
\hline
X-Decoder (T)      & 0 & 0.0M & 22.7 & 10.5 & 19.0 & 1.1 & 12.0 & 12.0 & 1.2  & 65.6 & 66.5 & 28.7 & 7.9  & 0.6  & 22.4 & 5.5 & 50.6 & 62.1 & 29.9 & 3.6  & 48.9 & 0.7 & 15.0 & 41.6 & 15.2 & 9.5  & 19.3 & 16.2  \\
X-Decoder (T)      & 1    & 1.15M & 22.8 & 10.5{\tiny $\pm$0.0} & 19.0{\tiny $\pm$0.0} & 1.1{\tiny $\pm$0.0} & 12.0{\tiny $\pm$0.0}  & 12.0{\tiny $\pm$0.0} & 1.2{\tiny $\pm$0.0}   & 65.6{\tiny $\pm$0.0} & 66.5{\tiny $\pm$0.0} & 28.7{\tiny $\pm$0.0} & 7.9{\tiny $\pm$5.8}   & 0.6{\tiny $\pm$0.0}   & 22.4{\tiny $\pm$0.0}  & 5.5{\tiny $\pm$0.0} & 50.6{\tiny $\pm$4.6} & 62.1{\tiny $\pm$0.0}  & 33.9{\tiny $\pm$4.9}  & 3.6{\tiny $\pm$0.0}  & 48.9{\tiny $\pm$4.6} & 0.7{\tiny $\pm$0.0}  & 15.0{\tiny $\pm$1.1} & 41.6{\tiny $\pm$0.0} & 15.2{\tiny $\pm$0.0}  & 9.5{\tiny $\pm$0.0}  & 19.3{\tiny $\pm$0.0}  & 16.2{\tiny $\pm$0.0}   \\
X-Decoder (T)      & 3    & 1.15M & 24.2 & 10.1{\tiny $\pm$0.1} & 22.9{\tiny $\pm$6.7} & 1.6{\tiny $\pm$0.6} & 19.2{\tiny $\pm$12.4} & 13.5{\tiny $\pm$0.4} & 1.2{\tiny $\pm$0.0}   & 66.0{\tiny $\pm$0.4} & 66.5{\tiny $\pm$0.0} & 29.9{\tiny $\pm$2.0} & 7.9{\tiny $\pm$5.8}   & 0.6{\tiny $\pm$0.0}   & 22.4{\tiny $\pm$0.0}  & 5.8{\tiny $\pm$0.4} & 50.6{\tiny $\pm$4.6} & 62.1{\tiny $\pm$0.0}  & 38.1{\tiny $\pm$3.9}  & 7.2{\tiny $\pm$8.4}  & 48.9{\tiny $\pm$4.6} & 0.9{\tiny $\pm$0.3}  & 15.0{\tiny $\pm$1.1} & 42.4{\tiny $\pm$1.3} & 16.5{\tiny $\pm$2.4}  & 10.2{\tiny $\pm$0.6} & 19.9{\tiny $\pm$1.0}  & 24.1{\tiny $\pm$5.0}   \\
X-Decoder (T)      & 5    & 1.15M & 27.9 & 10.5{\tiny $\pm$0.3} & 30.1{\tiny $\pm$1.8} & 4.1{\tiny $\pm$3.3} & 36.6{\tiny $\pm$3.4}  & 15.6{\tiny $\pm$1.4} & 2.2{\tiny $\pm$0.7}   & 66.0{\tiny $\pm$0.4} & 69.7{\tiny $\pm$2.9} & 31.6{\tiny $\pm$1.2} & 7.2{\tiny $\pm$0.0}   & 22.3{\tiny $\pm$25.9} & 31.8{\tiny $\pm$5.9}  & 7.9{\tiny $\pm$0.5} & 50.6{\tiny $\pm$0.0} & 65.4{\tiny $\pm$1.4}  & 43.7{\tiny $\pm$5.4}  & 17.9{\tiny $\pm$1.9} & 51.6{\tiny $\pm$1.0} & 1.0{\tiny $\pm$0.3}  & 15.7{\tiny $\pm$1.7} & 39.4{\tiny $\pm$4.7} & 17.8{\tiny $\pm$4.2}  & 11.8{\tiny $\pm$1.3} & 21.4{\tiny $\pm$0.2}  & 24.1{\tiny $\pm$7.8}   \\
X-Decoder (T)      & 10   & 1.15M & 34.5 & 10.1{\tiny $\pm$0.0} & 38.4{\tiny $\pm$2.6} & 4.7{\tiny $\pm$1.1} & 56.3{\tiny $\pm$7.4}  & 20.7{\tiny $\pm$5.6} & 2.6{\tiny $\pm$1.5}   & 66.8{\tiny $\pm$2.7} & 64.0{\tiny $\pm$7.7} & 36.1{\tiny $\pm$2.1} & 21.8{\tiny $\pm$5.5}  & 84.9{\tiny $\pm$2.6}  & 41.1{\tiny $\pm$14.3} & 8.3{\tiny $\pm$0.6} & 53.1{\tiny $\pm$3.2} & 68.0{\tiny $\pm$1.6}  & 45.4{\tiny $\pm$5.6}  & 3.6{\tiny $\pm$0.0}  & 49.7{\tiny $\pm$0.8} & 6.7{\tiny $\pm$2.4}  & 24.3{\tiny $\pm$5.3} & 43.0{\tiny $\pm$1.5} & 32.1{\tiny $\pm$5.8}  & 14.6{\tiny $\pm$3.1} & 22.6{\tiny $\pm$1.7}  & 41.8{\tiny $\pm$12.1}  \\
X-Decoder (T)      & 3$\times$All & 1.15M & 37.9 & 10.7{\tiny $\pm$0.1} & 37.1{\tiny $\pm$3.6} & 7.7{\tiny $\pm$2.9} & 13.1{\tiny $\pm$2.0}  & 30.5{\tiny $\pm$0.8} & 7.3{\tiny $\pm$3.5}   & 68.1{\tiny $\pm$0.5} & 73.7{\tiny $\pm$1.2} & 38.0{\tiny $\pm$0.8} & 10.0{\tiny $\pm$2.0}  & 38.1{\tiny $\pm$20.3} & 41.5{\tiny $\pm$2.9}  & 8.6{\tiny $\pm$0.1} & 50.6{\tiny $\pm$0.0} & 80.0{\tiny $\pm$0.2}  & 45.4{\tiny $\pm$2.1}  & 17.9{\tiny $\pm$1.9} & 50.7{\tiny $\pm$0.9} & 59.6{\tiny $\pm$2.0} & 54.6{\tiny $\pm$1.0} & 89.7{\tiny $\pm$0.1} & 28.9{\tiny $\pm$2.1}  & 12.9{\tiny $\pm$0.2} & 27.6{\tiny $\pm$0.4}  & 43.8{\tiny $\pm$0.5}   \\
\hline
X-Decoder-Seg$^+$ (B) & 0 & 0.0M & 26.3 & 13.2 & 17.2 & 0.8 & 33.0 & 28.6 & 4.9  & 67.9 & 71.1 & 28.8 & 5.2  & 0.0  & 0.8  & 6.8 & 50.6 & 53.2 & 18.8 & 17.9 & 68.2 & 0.7 & 21.1 & 86.3 & 5.8  & 11.5 & 12.1 & 31.7  \\
X-Decoder-Seg$^+$  (B) & 1    & 1.15M & 26.4 & 13.2{\tiny $\pm$0.0} & 17.2{\tiny $\pm$0.0} & 0.8{\tiny $\pm$0.0} & 33.0{\tiny $\pm$0.0}  & 28.6{\tiny $\pm$2.3} & 4.9{\tiny $\pm$0.0}   & 67.9{\tiny $\pm$0.0} & 74.1{\tiny $\pm$5.1} & 28.8{\tiny $\pm$0.0} & 5.2{\tiny $\pm$0.0}   & 0.0{\tiny $\pm$0.0}   & 0.8{\tiny $\pm$0.0}   & 6.8{\tiny $\pm$0.0} & 50.6{\tiny $\pm$0.0} & 53.2{\tiny $\pm$0.0}  & 18.8{\tiny $\pm$0.0}  & 17.9{\tiny $\pm$0.0} & 68.2{\tiny $\pm$0.0} & 0.7{\tiny $\pm$0.0}  & 21.1{\tiny $\pm$0.0} & 86.3{\tiny $\pm$0.0} & 5.8{\tiny $\pm$5.8}   & 11.5{\tiny $\pm$0.0} & 12.1{\tiny $\pm$1.1}  & 31.7{\tiny $\pm$2.3}   \\
X-Decoder-Seg$^+$  (B) & 3    & 1.15M & 26.6 & 13.3{\tiny $\pm$0.1} & 17.2{\tiny $\pm$0.0} & 1.1{\tiny $\pm$0.3} & 33.0{\tiny $\pm$0.0}  & 28.9{\tiny $\pm$0.5} & 4.9{\tiny $\pm$0.0}   & 67.2{\tiny $\pm$1.3} & 74.4{\tiny $\pm$4.9} & 28.7{\tiny $\pm$0.0} & 5.2{\tiny $\pm$0.1}   & 0.0{\tiny $\pm$0.0}   & 1.0{\tiny $\pm$0.0}   & 7.0{\tiny $\pm$0.2} & 50.6{\tiny $\pm$0.0} & 53.3{\tiny $\pm$0.0}  & 22.1{\tiny $\pm$3.5}  & 18.3{\tiny $\pm$0.3} & 66.1{\tiny $\pm$1.8} & 0.7{\tiny $\pm$0.0}  & 21.4{\tiny $\pm$0.8} & 86.6{\tiny $\pm$0.0} & 6.2{\tiny $\pm$0.7}   & 11.1{\tiny $\pm$0.5} & 12.2{\tiny $\pm$0.2}  & 33.5{\tiny $\pm$0.7}   \\
X-Decoder-Seg$^+$  (B) & 5    & 1.15M & 27.4 & 13.5{\tiny $\pm$0.1} & 17.7{\tiny $\pm$1.0} & 1.6{\tiny $\pm$0.8} & 31.5{\tiny $\pm$6.9}  & 32.7{\tiny $\pm$3.0} & 4.6{\tiny $\pm$0.9}   & 68.9{\tiny $\pm$2.9} & 79.7{\tiny $\pm$1.5} & 28.7{\tiny $\pm$0.0} & 6.7{\tiny $\pm$2.8}   & 0.0{\tiny $\pm$0.0}   & 1.2{\tiny $\pm$0.1}   & 7.6{\tiny $\pm$0.3} & 50.6{\tiny $\pm$0.0} & 53.7{\tiny $\pm$0.6}  & 25.6{\tiny $\pm$2.6}  & 18.6{\tiny $\pm$1.3} & 66.1{\tiny $\pm$1.4} & 0.8{\tiny $\pm$0.1}  & 21.2{\tiny $\pm$0.3} & 87.2{\tiny $\pm$0.6} & 6.9{\tiny $\pm$0.9}   & 10.1{\tiny $\pm$0.8} & 12.6{\tiny $\pm$0.2}  & 35.5{\tiny $\pm$3.2}   \\
X-Decoder-Seg$^+$  (B) & 10   & 1.15M & 29.9 & 13.5{\tiny $\pm$0.5} & 19.4{\tiny $\pm$0.9} & 2.5{\tiny $\pm$1.1} & 45.1{\tiny $\pm$10.9} & 37.5{\tiny $\pm$2.8} & 7.2{\tiny $\pm$3.9}   & 69.4{\tiny $\pm$2.9} & 81.4{\tiny $\pm$0.4} & 28.8{\tiny $\pm$0.3} & 8.2{\tiny $\pm$2.3}   & 0.0{\tiny $\pm$0.0}   & 2.1{\tiny $\pm$0.3}   & 8.2{\tiny $\pm$0.1} & 51.9{\tiny $\pm$1.8} & 54.6{\tiny $\pm$1.7}  & 32.2{\tiny $\pm$7.7}  & 17.9{\tiny $\pm$0.0} & 71.7{\tiny $\pm$1.5} & 1.7{\tiny $\pm$0.1}  & 24.5{\tiny $\pm$0.8} & 87.5{\tiny $\pm$0.2} & 10.7{\tiny $\pm$1.3}  & 14.9{\tiny $\pm$1.5} & 12.5{\tiny $\pm$0.1}  & 42.0{\tiny $\pm$3.8}   \\
X-Decoder-Seg$^+$  (B) & 3$\times$All & 1.15M & 31.6 & 13.5{\tiny $\pm$0.1} & 19.0{\tiny $\pm$0.5} & 1.7{\tiny $\pm$0.0} & 33.0{\tiny $\pm$0.0}  & 37.7{\tiny $\pm$0.3} & 3.8{\tiny $\pm$0.1}   & 73.7{\tiny $\pm$0.2} & 75.0{\tiny $\pm$3.8} & 29.0{\tiny $\pm$0.1} & 5.0{\tiny $\pm$0.4}   & 0.0{\tiny $\pm$0.0}   & 3.6{\tiny $\pm$0.5}   & 8.6{\tiny $\pm$0.2} & 50.7{\tiny $\pm$0.0} & 63.7{\tiny $\pm$0.9}  & 24.7{\tiny $\pm$1.2}  & 19.4{\tiny $\pm$1.2} & 65.3{\tiny $\pm$0.1} & 45.7{\tiny $\pm$1.0} & 53.0{\tiny $\pm$1.0} & 89.6{\tiny $\pm$0.0} & 8.4{\tiny $\pm$0.7}   & 13.5{\tiny $\pm$1.3} & 13.4{\tiny $\pm$0.1}  & 38.2{\tiny $\pm$0.5}   \\ 
\hline
X-Decoder (B)     & 0 & 0.0M & 27.7 & 13.0 & 45.9 & 0.3 & 13.6 & 36.8 & 4.2  & 68.0 & 76.7 & 30.2 & 19.4 & 20.6 & 18.5 & 6.7 & 51.7 & 53.1 & 8.9  & 5.6  & 55.4 & 0.8 & 18.2 & 81.6 & 8.0  & 13.9 & 27.3 & 13.0  \\
X-Decoder (B)     & 1    & 1.15M & 26.8 & 13.0{\tiny $\pm$0.0} & 45.9{\tiny $\pm$0.0} & 0.3{\tiny $\pm$0.0} & 9.0{\tiny $\pm$7.8}   & 36.8{\tiny $\pm$0.0} & 4.1{\tiny $\pm$0.2}   & 68.0{\tiny $\pm$0.0} & 76.7{\tiny $\pm$0.0} & 30.2{\tiny $\pm$0.0} & 19.4{\tiny $\pm$0.0}  & 20.6{\tiny $\pm$0.0}  & 18.5{\tiny $\pm$0.0}  & 6.7{\tiny $\pm$5.8} & 51.7{\tiny $\pm$0.0} & 35.4{\tiny $\pm$30.6} & 8.9{\tiny $\pm$0.0}   & 4.9{\tiny $\pm$0.5}  & 55.4{\tiny $\pm$0.0} & 0.8{\tiny $\pm$0.0}  & 18.2{\tiny $\pm$0.0} & 81.6{\tiny $\pm$0.0} & 8.0{\tiny $\pm$0.0}   & 13.9{\tiny $\pm$0.0} & 27.3{\tiny $\pm$0.0}  & 13.0{\tiny $\pm$0.0}   \\
X-Decoder (B)     & 3    & 1.15M & 28.6 & 12.5{\tiny $\pm$0.1} & 44.3{\tiny $\pm$2.4} & 0.2{\tiny $\pm$0.0} & 21.4{\tiny $\pm$7.2}  & 37.7{\tiny $\pm$0.8} & 5.1{\tiny $\pm$1.5}   & 67.9{\tiny $\pm$0.1} & 76.6{\tiny $\pm$0.2} & 30.3{\tiny $\pm$0.0} & 24.9{\tiny $\pm$4.9}  & 23.4{\tiny $\pm$2.4}  & 18.5{\tiny $\pm$0.0}  & 7.1{\tiny $\pm$0.6} & 51.7{\tiny $\pm$0.0} & 53.1{\tiny $\pm$4.6}  & 10.7{\tiny $\pm$5.4}  & 6.7{\tiny $\pm$0.4}  & 55.1{\tiny $\pm$0.3} & 0.8{\tiny $\pm$0.0}  & 18.6{\tiny $\pm$2.7} & 84.2{\tiny $\pm$2.3} & 8.1{\tiny $\pm$1.5}   & 12.6{\tiny $\pm$2.3} & 27.3{\tiny $\pm$0.0}  & 14.0{\tiny $\pm$1.8}   \\
X-Decoder (B)     & 5    & 1.15M & 33.0 & 12.4{\tiny $\pm$0.9} & 43.7{\tiny $\pm$0.9} & 0.4{\tiny $\pm$0.3} & 40.0{\tiny $\pm$4.1}  & 38.1{\tiny $\pm$0.7} & 9.0{\tiny $\pm$3.7}   & 67.4{\tiny $\pm$0.4} & 79.9{\tiny $\pm$1.2} & 30.9{\tiny $\pm$0.8} & 34.8{\tiny $\pm$4.8}  & 34.4{\tiny $\pm$10.0} & 51.1{\tiny $\pm$13.7} & 8.4{\tiny $\pm$0.1} & 54.2{\tiny $\pm$2.1} & 65.6{\tiny $\pm$2.2}  & 15.1{\tiny $\pm$5.3}  & 4.5{\tiny $\pm$0.4}  & 55.4{\tiny $\pm$0.0} & 1.1{\tiny $\pm$0.2}  & 24.4{\tiny $\pm$5.6} & 87.0{\tiny $\pm$2.3} & 8.5{\tiny $\pm$0.1}   & 12.5{\tiny $\pm$3.5} & 28.5{\tiny $\pm$0.4}  & 16.6{\tiny $\pm$2.1}   \\
X-Decoder (B)     & 10   & 1.15M & 39.7 & 10.3{\tiny $\pm$1.8} & 36.8{\tiny $\pm$4.1} & 1.4{\tiny $\pm$0.8} & 49.3{\tiny $\pm$0.9}  & 38.8{\tiny $\pm$1.3} & 28.4{\tiny $\pm$2.2}  & 69.2{\tiny $\pm$1.6} & 76.3{\tiny $\pm$0.4} & 31.4{\tiny $\pm$0.5} & 37.2{\tiny $\pm$1.1}  & 92.2{\tiny $\pm$0.2}  & 60.1{\tiny $\pm$6.0}  & 8.8{\tiny $\pm$0.7} & 55.8{\tiny $\pm$3.8} & 74.2{\tiny $\pm$1.8}  & 28.7{\tiny $\pm$11.4} & 5.6{\tiny $\pm$0.0}  & 57.0{\tiny $\pm$2.4} & 3.2{\tiny $\pm$0.4}  & 30.2{\tiny $\pm$2.9} & 88.0{\tiny $\pm$1.7} & 19.8{\tiny $\pm$2.3}  & 17.6{\tiny $\pm$6.0} & 30.9{\tiny $\pm$1.4}  & 40.4{\tiny $\pm$2.5}   \\
X-Decoder (B)     & 3$\times$All & 1.15M & 38.7 & 12.6{\tiny $\pm$0.2} & 41.7{\tiny $\pm$1.7} & 1.1{\tiny $\pm$0.1} & 13.6{\tiny $\pm$1.1}  & 36.5{\tiny $\pm$0.1} & 23.1{\tiny $\pm$7.9}  & 71.5{\tiny $\pm$0.3} & 77.3{\tiny $\pm$0.3} & 31.5{\tiny $\pm$0.2} & 32.4{\tiny $\pm$6.1}  & 47.3{\tiny $\pm$14.1} & 65.6{\tiny $\pm$0.5}  & 9.2{\tiny $\pm$0.0} & 53.8{\tiny $\pm$0.3} & 82.2{\tiny $\pm$0.1}  & 17.1{\tiny $\pm$1.2}  & 5.5{\tiny $\pm$1.4}  & 55.8{\tiny $\pm$0.7} & 48.4{\tiny $\pm$3.8} & 47.9{\tiny $\pm$1.2} & 90.0{\tiny $\pm$0.3} & 18.3{\tiny $\pm$0.3}  & 12.8{\tiny $\pm$0.7} & 36.9{\tiny $\pm$1.0}  & 33.2{\tiny $\pm$0.9}   \\ 
\hline
X-Decoder (L-IN21K)      & 0 & 0.0M & 26.7 & 12.3 & 43.2 & 0.5 & 3.5  & 12.3 & 18.8 & 63.9 & 79.1 & 24.3 & 15.6 & 0.0  & 20.3 & 4.9 & 50.5 & 58.8 & 43.4 & 13.4 & 57.3 & 1.3 & 12.3 & 74.4 & 6.9  & 14.6 & 20.1 & 13.5  \\
X-Decoder (L-IN21K)      & 1    & 1.15M & 27.0 & 12.3{\tiny $\pm$0.0} & 43.2{\tiny $\pm$4.6} & 0.5{\tiny $\pm$0.0} & 3.5{\tiny $\pm$0.0}   & 20.1{\tiny $\pm$6.8} & 18.8{\tiny $\pm$0.0}  & 63.9{\tiny $\pm$4.6} & 78.6{\tiny $\pm$0.4} & 24.3{\tiny $\pm$0.0} & 15.6{\tiny $\pm$1.1}  & 0.0{\tiny $\pm$0.0}   & 20.3{\tiny $\pm$0.0}  & 4.9{\tiny $\pm$0.0} & 50.5{\tiny $\pm$0.0} & 58.8{\tiny $\pm$0.0}  & 43.4{\tiny $\pm$0.0}  & 13.6{\tiny $\pm$0.3} & 57.3{\tiny $\pm$0.0} & 1.3{\tiny $\pm$0.0}  & 12.3{\tiny $\pm$0.0} & 76.4{\tiny $\pm$3.4} & 7.1{\tiny $\pm$0.2}   & 14.6{\tiny $\pm$0.0} & 20.1{\tiny $\pm$0.0}  & 13.5{\tiny $\pm$0.0}   \\
X-Decoder (L-IN21K)      & 3    & 1.15M & 30.1 & 12.7{\tiny $\pm$0.8} & 45.8{\tiny $\pm$4.5} & 0.6{\tiny $\pm$0.1} & 15.4{\tiny $\pm$9.0}  & 28.6{\tiny $\pm$1.9} & 38.0{\tiny $\pm$18.2} & 64.1{\tiny $\pm$0.2} & 78.2{\tiny $\pm$1.5} & 24.3{\tiny $\pm$0.0} & 15.6{\tiny $\pm$1.1}  & 26.3{\tiny $\pm$23.1} & 20.9{\tiny $\pm$3.3}  & 5.6{\tiny $\pm$0.4} & 50.5{\tiny $\pm$0.0} & 60.8{\tiny $\pm$3.5}  & 42.2{\tiny $\pm$2.3}  & 12.2{\tiny $\pm$3.1} & 57.3{\tiny $\pm$0.0} & 1.5{\tiny $\pm$0.3}  & 17.8{\tiny $\pm$9.5} & 77.8{\tiny $\pm$5.7} & 7.0{\tiny $\pm$4.4}   & 15.2{\tiny $\pm$1.0} & 13.4{\tiny $\pm$11.6} & 19.3{\tiny $\pm$9.9}   \\
X-Decoder (L-IN21K)      & 5    & 1.15M & 34.0 & 13.8{\tiny $\pm$1.4} & 48.2{\tiny $\pm$3.1} & 0.7{\tiny $\pm$0.3} & 26.3{\tiny $\pm$6.2}  & 28.7{\tiny $\pm$3.9} & 33.3{\tiny $\pm$25.4} & 62.8{\tiny $\pm$0.2} & 79.0{\tiny $\pm$1.9} & 29.3{\tiny $\pm$2.0} & 16.4{\tiny $\pm$10.1} & 60.1{\tiny $\pm$52.0} & 34.5{\tiny $\pm$22.5} & 7.2{\tiny $\pm$0.4} & 52.2{\tiny $\pm$2.5} & 63.5{\tiny $\pm$1.7}  & 37.0{\tiny $\pm$10.8} & 7.6{\tiny $\pm$6.2}  & 59.8{\tiny $\pm$0.8} & 2.3{\tiny $\pm$0.2}  & 21.8{\tiny $\pm$3.5} & 85.4{\tiny $\pm$1.1} & 5.7{\tiny $\pm$2.3}   & 16.0{\tiny $\pm$2.1} & 27.7{\tiny $\pm$10.8} & 29.8{\tiny $\pm$8.7}   \\
X-Decoder (L-IN21K)      & 10   & 1.15M & 40.3 & 12.5{\tiny $\pm$1.4} & 46.5{\tiny $\pm$7.0} & 0.8{\tiny $\pm$0.5} & 36.2{\tiny $\pm$3.6}  & 31.9{\tiny $\pm$3.9} & 44.7{\tiny $\pm$29.6} & 63.1{\tiny $\pm$1.6} & 79.8{\tiny $\pm$1.0} & 33.2{\tiny $\pm$5.5} & 34.3{\tiny $\pm$3.6}  & 88.9{\tiny $\pm$3.1}  & 62.2{\tiny $\pm$0.8}  & 7.2{\tiny $\pm$1.4} & 53.2{\tiny $\pm$2.2} & 76.0{\tiny $\pm$2.1}  & 27.2{\tiny $\pm$21.0} & 13.4{\tiny $\pm$0.0} & 58.8{\tiny $\pm$3.7} & 2.4{\tiny $\pm$0.1}  & 23.3{\tiny $\pm$4.1} & 88.0{\tiny $\pm$1.4} & 25.9{\tiny $\pm$12.8} & 17.6{\tiny $\pm$2.7} & 34.3{\tiny $\pm$6.9}  & 43.9{\tiny $\pm$1.3}   \\
X-Decoder (L-IN21K)      & 3$\times$All & 1.15M & 40.7 & 14.5{\tiny $\pm$0.5} & 53.3{\tiny $\pm$0.4} & 0.5{\tiny $\pm$0.1} & 4.1{\tiny $\pm$1.0}   & 36.8{\tiny $\pm$0.0} & 64.3{\tiny $\pm$0.2}  & 70.7{\tiny $\pm$0.4} & 80.7{\tiny $\pm$1.1} & 32.1{\tiny $\pm$0.1} & 13.2{\tiny $\pm$5.6}  & 20.4{\tiny $\pm$1.9}  & 61.5{\tiny $\pm$0.7}  & 7.9{\tiny $\pm$0.1} & 51.6{\tiny $\pm$0.7} & 84.8{\tiny $\pm$0.0}  & 43.2{\tiny $\pm$1.1}  & 13.5{\tiny $\pm$6.0} & 60.5{\tiny $\pm$0.3} & 42.9{\tiny $\pm$2.7} & 48.8{\tiny $\pm$0.9} & 90.4{\tiny $\pm$0.3} & 24.5{\tiny $\pm$3.6}  & 19.2{\tiny $\pm$0.8} & 41.6{\tiny $\pm$0.3}  & 35.7{\tiny $\pm$1.2}   \\ 
\hline
X-Decoder (L)     & 0 & 0.0M & 32.3 & 13.1 & 42.1 & 2.2 & 8.6  & 44.9 & 7.5  & 66.0 & 79.2 & 33.0 & 11.6 & 75.9 & 42.1 & 7.0 & 53.0 & 68.4 & 15.6 & 20.1 & 59.0 & 2.3 & 19.0 & 67.1 & 22.5 & 9.9  & 22.3 & 13.8  \\
X-Decoder (L)     & 1    & 1.15M & 32.3 & 13.1{\tiny $\pm$0.0} & 42.1{\tiny $\pm$0.0} & 2.2{\tiny $\pm$0.0} & 8.6{\tiny $\pm$0.0}   & 44.9{\tiny $\pm$0.0} & 7.5{\tiny $\pm$0.0}   & 66.0{\tiny $\pm$0.0} & 79.2{\tiny $\pm$0.0} & 33.0{\tiny $\pm$0.0} & 11.6{\tiny $\pm$1.1}  & 75.9{\tiny $\pm$0.0}  & 42.1{\tiny $\pm$0.0}  & 7.0{\tiny $\pm$0.0} & 53.0{\tiny $\pm$0.0} & 68.4{\tiny $\pm$0.0}  & 15.6{\tiny $\pm$1.1}  & 20.1{\tiny $\pm$0.0} & 59.0{\tiny $\pm$0.0} & 2.3{\tiny $\pm$0.0}  & 19.0{\tiny $\pm$0.0} & 67.1{\tiny $\pm$0.0} & 22.5{\tiny $\pm$0.0}  & 9.9{\tiny $\pm$0.0}  & 22.3{\tiny $\pm$0.0}  & 13.8{\tiny $\pm$0.0}   \\
X-Decoder (L)     & 3    & 1.15M & 32.6 & 13.3{\tiny $\pm$0.3} & 41.4{\tiny $\pm$1.2} & 2.2{\tiny $\pm$0.0} & 8.6{\tiny $\pm$0.0}   & 45.5{\tiny $\pm$0.4} & 7.5{\tiny $\pm$0.0}   & 66.4{\tiny $\pm$0.6} & 79.2{\tiny $\pm$0.0} & 33.0{\tiny $\pm$0.0} & 11.6{\tiny $\pm$1.1}  & 75.9{\tiny $\pm$0.0}  & 42.1{\tiny $\pm$0.0}  & 7.1{\tiny $\pm$0.2} & 53.0{\tiny $\pm$0.0} & 68.4{\tiny $\pm$0.0}  & 18.0{\tiny $\pm$3.3}  & 20.7{\tiny $\pm$0.4} & 59.0{\tiny $\pm$0.0} & 2.3{\tiny $\pm$0.0}  & 19.0{\tiny $\pm$0.0} & 67.1{\tiny $\pm$0.0} & 22.0{\tiny $\pm$0.8}  & 9.9{\tiny $\pm$0.0}  & 22.9{\tiny $\pm$0.9}  & 17.1{\tiny $\pm$5.6}   \\
X-Decoder (L)     & 5    & 1.15M & 35.5 & 13.7{\tiny $\pm$0.5} & 46.9{\tiny $\pm$4.1} & 4.0{\tiny $\pm$1.6} & 33.2{\tiny $\pm$1.0}  & 45.7{\tiny $\pm$0.6} & 12.1{\tiny $\pm$4.8}  & 65.9{\tiny $\pm$1.6} & 77.6{\tiny $\pm$0.7} & 32.9{\tiny $\pm$0.5} & 20.3{\tiny $\pm$10.3} & 75.9{\tiny $\pm$0.0}  & 41.4{\tiny $\pm$1.2}  & 6.9{\tiny $\pm$0.5} & 53.0{\tiny $\pm$0.2} & 70.3{\tiny $\pm$2.0}  & 20.2{\tiny $\pm$1.5}  & 22.4{\tiny $\pm$1.9} & 59.1{\tiny $\pm$0.9} & 3.1{\tiny $\pm$0.4}  & 16.8{\tiny $\pm$2.1} & 80.1{\tiny $\pm$2.3} & 27.7{\tiny $\pm$4.3}  & 9.4{\tiny $\pm$0.8}  & 23.9{\tiny $\pm$0.4}  & 23.6{\tiny $\pm$8.2}   \\
X-Decoder (L)     & 10   & 1.15M & 40.5 & 13.6{\tiny $\pm$0.1} & 45.4{\tiny $\pm$1.8} & 4.5{\tiny $\pm$2.4} & 44.9{\tiny $\pm$0.4}  & 46.0{\tiny $\pm$1.1} & 35.2{\tiny $\pm$10.8} & 66.7{\tiny $\pm$3.8} & 78.6{\tiny $\pm$1.7} & 39.2{\tiny $\pm$2.6} & 20.1{\tiny $\pm$5.2}  & 94.3{\tiny $\pm$0.3}  & 59.9{\tiny $\pm$3.4}  & 7.2{\tiny $\pm$0.2} & 52.0{\tiny $\pm$0.7} & 77.8{\tiny $\pm$1.3}  & 24.5{\tiny $\pm$2.8}  & 20.1{\tiny $\pm$0.0} & 56.7{\tiny $\pm$1.3} & 3.2{\tiny $\pm$0.6}  & 19.7{\tiny $\pm$2.5} & 86.9{\tiny $\pm$0.4} & 38.8{\tiny $\pm$4.5}  & 8.8{\tiny $\pm$3.2}  & 30.7{\tiny $\pm$1.2}  & 36.5{\tiny $\pm$4.5}   \\
X-Decoder (L)     & 3$\times$All & 1.15M & 42.3 & 13.4{\tiny $\pm$0.0} & 48.8{\tiny $\pm$0.7} & 6.3{\tiny $\pm$1.2} & 8.6{\tiny $\pm$0.0}   & 45.1{\tiny $\pm$0.0} & 20.5{\tiny $\pm$1.0}  & 72.1{\tiny $\pm$0.1} & 79.3{\tiny $\pm$0.9} & 36.9{\tiny $\pm$0.9} & 12.8{\tiny $\pm$1.1}  & 88.5{\tiny $\pm$2.5}  & 63.1{\tiny $\pm$1.9}  & 7.6{\tiny $\pm$0.0} & 52.8{\tiny $\pm$0.8} & 83.6{\tiny $\pm$0.2}  & 22.1{\tiny $\pm$1.1}  & 21.7{\tiny $\pm$1.9} & 59.2{\tiny $\pm$0.8} & 43.7{\tiny $\pm$2.7} & 50.0{\tiny $\pm$1.0} & 91.7{\tiny $\pm$0.0} & 40.9{\tiny $\pm$1.4}  & 9.8{\tiny $\pm$0.3}  & 36.6{\tiny $\pm$0.3}  & 40.7{\tiny $\pm$1.7}   \\
\hline
\end{tabular}
}
\vspace{3pt}
\caption{SegInW results with tuning on class \& mask embeddings and latent queries for different image shots and backbone architectures. (1.15M parameters tuned in the setting.)}
\label{tab:seginw_prompt}
\end{table*}

\begin{table*}
\footnotesize \setlength{\tabcolsep}{1.0pt}
\centering
\resizebox{1.0\linewidth}{!}{
\begin{tabular}{lcc|c|ccccccccccccccccccccccccc} 
\hline
Model       & Shot     & \#Param & Avg  & \begin{tabular}[c]{@{}c@{}}Airplane-\\Parts\end{tabular} & Bottles                                                & \begin{tabular}[c]{@{}c@{}}Brain-\\Tumor\end{tabular} & Chicken                                                & Cows                                                   & \begin{tabular}[c]{@{}c@{}}Electric-\\Shaver\end{tabular} & Elephants                                             & Fruits                                                 & Garbage                                                & \begin{tabular}[c]{@{}c@{}}Ginger-\\Garlic\end{tabular} & Hand                                                   & \begin{tabular}[c]{@{}c@{}}Hand-\\Metal\end{tabular}   & \begin{tabular}[c]{@{}c@{}}House-\\Parts\end{tabular} & \begin{tabular}[c]{@{}c@{}}HH.-\\Items\end{tabular} & \begin{tabular}[c]{@{}c@{}}Nutterfly-\\Squireel\end{tabular} & Phones                                                 & Poles                                                  & Puppies                                                & Rail                                                   & \begin{tabular}[c]{@{}c@{}}Salmon-\\Fillet\end{tabular} & Strawberry                                             & Tablets                                                & Toolkits                                              & Trash                                                 & Watermelon                                              \\ 
\hline
X-Decoder (T)      & 0 & 0.0M & 22.7 & 10.5 & 19.0 & 1.1 & 12.0 & 12.0 & 1.2  & 65.6 & 66.5 & 28.7 & 7.9  & 0.6  & 22.4 & 5.5 & 50.6 & 62.1 & 29.9 & 3.6  & 48.9 & 0.7 & 15.0 & 41.6 & 15.2 & 9.5  & 19.3 & 16.2  \\
X-Decoder (T)      & 1    & 0.26M & 22.6 & 10.5{\tiny $\pm$0.0}    & 19.0{\tiny $\pm$0.0}  & 1.1{\tiny $\pm$0.0}  & 12.0{\tiny $\pm$0.0}  & 12.0{\tiny $\pm$0.0}  & 1.2{\tiny $\pm$0.0}      & 65.6{\tiny $\pm$0.0} & 66.5{\tiny $\pm$0.0}  & 28.7{\tiny $\pm$0.0}  & 7.9{\tiny $\pm$5.8}    & 0.6{\tiny $\pm$0.0}   & 22.4{\tiny $\pm$0.0}  & 5.5{\tiny $\pm$0.0}  & 50.6{\tiny $\pm$4.6}     & 62.1{\tiny $\pm$0.0}        & 28.5{\tiny $\pm$2.1}  & 3.6{\tiny $\pm$0.0}   & 48.9{\tiny $\pm$4.6}  & 0.7{\tiny $\pm$0.0}   & 15.0{\tiny $\pm$1.1}   & 41.6{\tiny $\pm$0.0}  & 15.2{\tiny $\pm$0.0}  & 9.5{\tiny $\pm$0.0}  & 19.3{\tiny $\pm$0.0} & 16.2{\tiny $\pm$0.0}   \\
X-Decoder (T)      & 3    & 0.26M & 25.1 & 10.4{\tiny $\pm$0.1}    & 20.2{\tiny $\pm$2.1}  & 3.4{\tiny $\pm$4.0}  & 42.7{\tiny $\pm$17.8} & 14.4{\tiny $\pm$2.3}  & 1.2{\tiny $\pm$0.0}      & 66.1{\tiny $\pm$0.8} & 65.5{\tiny $\pm$3.1}  & 28.8{\tiny $\pm$0.2}  & 11.6{\tiny $\pm$6.3}   & 0.6{\tiny $\pm$0.0}   & 22.4{\tiny $\pm$0.0}  & 7.4{\tiny $\pm$1.1}  & 50.6{\tiny $\pm$0.0}     & 62.1{\tiny $\pm$0.0}        & 33.4{\tiny $\pm$14.3} & 9.6{\tiny $\pm$6.6}   & 49.3{\tiny $\pm$0.6}  & 0.7{\tiny $\pm$0.0}   & 15.0{\tiny $\pm$1.1}   & 41.6{\tiny $\pm$0.0}  & 15.3{\tiny $\pm$3.2}  & 11.8{\tiny $\pm$2.5} & 20.3{\tiny $\pm$1.7} & 20.5{\tiny $\pm$3.7}   \\
X-Decoder (T)      & 5    & 0.26M & 29.7 & 10.5{\tiny $\pm$0.7}    & 33.7{\tiny $\pm$3.6}  & 7.8{\tiny $\pm$2.6}  & 33.0{\tiny $\pm$16.3} & 15.2{\tiny $\pm$2.6}  & 14.2{\tiny $\pm$12.7}    & 65.5{\tiny $\pm$1.7} & 65.5{\tiny $\pm$8.8}  & 34.7{\tiny $\pm$3.1}  & 16.7{\tiny $\pm$1.6}   & 51.0{\tiny $\pm$18.1} & 30.3{\tiny $\pm$4.6}  & 7.9{\tiny $\pm$0.6}  & 51.1{\tiny $\pm$0.8}     & 63.6{\tiny $\pm$4.1}        & 46.0{\tiny $\pm$5.6}  & 13.8{\tiny $\pm$11.0} & 49.5{\tiny $\pm$0.8}  & 0.5{\tiny $\pm$0.4}   & 18.4{\tiny $\pm$8.5}   & 34.6{\tiny $\pm$2.8}  & 19.7{\tiny $\pm$3.7}  & 13.2{\tiny $\pm$1.7} & 18.9{\tiny $\pm$1.4} & 26.8{\tiny $\pm$6.4}   \\
X-Decoder (T)      & 10   & 0.26M & 36.2 & 10.9{\tiny $\pm$1.6}    & 35.2{\tiny $\pm$5.5}  & 6.2{\tiny $\pm$2.6}  & 61.8{\tiny $\pm$3.1}  & 19.8{\tiny $\pm$5.7}  & 46.2{\tiny $\pm$6.8}     & 66.0{\tiny $\pm$2.7} & 63.2{\tiny $\pm$10.3} & 34.9{\tiny $\pm$3.8}  & 19.5{\tiny $\pm$13.6}  & 92.0{\tiny $\pm$1.0}  & 46.1{\tiny $\pm$15.0} & 10.6{\tiny $\pm$1.2} & 56.3{\tiny $\pm$3.3}     & 67.9{\tiny $\pm$2.1}        & 33.3{\tiny $\pm$12.5} & 3.6{\tiny $\pm$0.0}   & 45.8{\tiny $\pm$3.5}  & 6.3{\tiny $\pm$0.6}   & 22.8{\tiny $\pm$8.7}   & 52.4{\tiny $\pm$21.2} & 25.7{\tiny $\pm$10.4} & 16.4{\tiny $\pm$6.3} & 21.2{\tiny $\pm$0.9} & 39.9{\tiny $\pm$2.1}   \\
X-Decoder (T)      & 3$\times$All & 0.26M & 41.9 & 10.7{\tiny $\pm$0.4}    & 42.8{\tiny $\pm$2.2}  & 8.7{\tiny $\pm$0.7}  & 13.6{\tiny $\pm$2.8}  & 30.5{\tiny $\pm$3.5}  & 27.5{\tiny $\pm$11.0}    & 69.0{\tiny $\pm$0.8} & 70.8{\tiny $\pm$2.8}  & 38.5{\tiny $\pm$0.0}  & 10.3{\tiny $\pm$1.6}   & 74.0{\tiny $\pm$1.2}  & 61.0{\tiny $\pm$2.0}  & 13.9{\tiny $\pm$0.3} & 50.7{\tiny $\pm$0.0}     & 81.3{\tiny $\pm$0.8}        & 44.2{\tiny $\pm$3.0}  & 20.1{\tiny $\pm$0.0}  & 50.3{\tiny $\pm$0.6}  & 62.6{\tiny $\pm$2.6}  & 55.0{\tiny $\pm$0.6}   & 90.8{\tiny $\pm$0.1}  & 28.0{\tiny $\pm$1.9}  & 18.4{\tiny $\pm$2.7} & 27.3{\tiny $\pm$0.9} & 45.3{\tiny $\pm$1.7}   \\ 
\hline
X-Decoder-Seg$^+$ (B) & 0 & 0.0M & 26.3 & 13.2 & 17.2 & 0.8 & 33.0 & 28.6 & 4.9  & 67.9 & 71.1 & 28.8 & 5.2  & 0.0  & 0.8  & 6.8 & 50.6 & 53.2 & 18.8 & 17.9 & 68.2 & 0.7 & 21.1 & 86.3 & 5.8  & 11.5 & 12.1 & 31.7  \\
X-Decoder-Seg$^+$ (B) & 1    & 0.26M & 26.3 & 13.2{\tiny $\pm$0.0}    & 17.2{\tiny $\pm$0.0}  & 0.8{\tiny $\pm$0.0}  & 33.0{\tiny $\pm$0.0}  & 28.6{\tiny $\pm$0.0}  & 4.9{\tiny $\pm$0.0}      & 67.9{\tiny $\pm$0.0} & 71.1{\tiny $\pm$0.0}  & 28.8{\tiny $\pm$0.0}  & 5.2{\tiny $\pm$0.0}    & 0.0{\tiny $\pm$0.0}   & 1.5{\tiny $\pm$1.2}   & 6.8{\tiny $\pm$5.8}  & 50.6{\tiny $\pm$0.0}     & 53.2{\tiny $\pm$0.0}        & 18.8{\tiny $\pm$0.0}  & 17.9{\tiny $\pm$0.0}  & 67.3{\tiny $\pm$1.6}  & 0.7{\tiny $\pm$0.0}   & 21.1{\tiny $\pm$0.0}   & 86.3{\tiny $\pm$0.0}  & 5.8{\tiny $\pm$5.8}   & 11.5{\tiny $\pm$0.0} & 12.1{\tiny $\pm$1.1} & 31.8{\tiny $\pm$0.0}   \\
X-Decoder-Seg$^+$ (B) & 3    & 0.26M & 28.5 & 13.2{\tiny $\pm$0.3}    & 17.2{\tiny $\pm$0.0}  & 1.4{\tiny $\pm$0.5}  & 33.0{\tiny $\pm$0.0}  & 31.2{\tiny $\pm$3.4}  & 9.5{\tiny $\pm$6.7}      & 67.9{\tiny $\pm$0.0} & 74.1{\tiny $\pm$5.2}  & 28.8{\tiny $\pm$0.0}  & 7.0{\tiny $\pm$3.1}    & 1.3{\tiny $\pm$0.7}   & 12.4{\tiny $\pm$16.4} & 7.3{\tiny $\pm$0.3}  & 52.6{\tiny $\pm$3.4}     & 55.8{\tiny $\pm$2.5}        & 31.3{\tiny $\pm$7.3}  & 17.6{\tiny $\pm$2.5}  & 68.2{\tiny $\pm$0.0}  & 0.7{\tiny $\pm$0.0}   & 21.9{\tiny $\pm$3.6}   & 86.9{\tiny $\pm$0.9}  & 9.7{\tiny $\pm$0.7}   & 11.9{\tiny $\pm$0.6} & 12.0{\tiny $\pm$0.7} & 39.0{\tiny $\pm$7.2}   \\
X-Decoder-Seg$^+$ (B) & 5    & 0.26M & 32.4 & 13.7{\tiny $\pm$0.0}    & 20.7{\tiny $\pm$0.9}  & 2.3{\tiny $\pm$0.5}  & 27.1{\tiny $\pm$14.4} & 30.2{\tiny $\pm$0.6}  & 41.7{\tiny $\pm$27.2}    & 67.3{\tiny $\pm$1.7} & 77.5{\tiny $\pm$0.2}  & 28.0{\tiny $\pm$1.0}  & 17.8{\tiny $\pm$7.7}   & 3.2{\tiny $\pm$1.8}   & 28.3{\tiny $\pm$11.3} & 7.2{\tiny $\pm$0.4}  & 50.8{\tiny $\pm$0.1}     & 58.5{\tiny $\pm$4.3}        & 39.6{\tiny $\pm$12.5} & 20.4{\tiny $\pm$0.3}  & 69.6{\tiny $\pm$2.3}  & 1.0{\tiny $\pm$0.4}   & 31.4{\tiny $\pm$5.7}   & 86.6{\tiny $\pm$1.5}  & 10.9{\tiny $\pm$3.5}  & 14.1{\tiny $\pm$1.3} & 13.0{\tiny $\pm$1.0} & 48.3{\tiny $\pm$5.4}   \\
X-Decoder-Seg$^+$ (B) & 10   & 0.26M & 41.7 & 14.2{\tiny $\pm$1.7}    & 24.9{\tiny $\pm$4.3}  & 5.5{\tiny $\pm$0.1}  & 66.5{\tiny $\pm$2.7}  & 36.5{\tiny $\pm$0.4}  & 68.0{\tiny $\pm$5.8}     & 69.3{\tiny $\pm$2.6} & 69.8{\tiny $\pm$9.4}  & 28.5{\tiny $\pm$0.5}  & 24.9{\tiny $\pm$6.0}   & 94.2{\tiny $\pm$0.6}  & 46.4{\tiny $\pm$7.1}  & 9.5{\tiny $\pm$0.0}  & 50.5{\tiny $\pm$0.1}     & 70.3{\tiny $\pm$3.8}        & 52.4{\tiny $\pm$2.2}  & 17.9{\tiny $\pm$0.0}  & 59.5{\tiny $\pm$5.2}  & 16.0{\tiny $\pm$12.1} & 37.6{\tiny $\pm$6.1}   & 87.6{\tiny $\pm$0.5}  & 13.4{\tiny $\pm$0.3}  & 13.1{\tiny $\pm$3.3} & 11.8{\tiny $\pm$0.7} & 51.8{\tiny $\pm$3.2}   \\
X-Decoder-Seg$^+$ (B) & 3$\times$All & 0.26M & 40.3 & 13.5{\tiny $\pm$0.2}    & 21.8{\tiny $\pm$1.2}  & 3.3{\tiny $\pm$0.5}  & 33.0{\tiny $\pm$0.0}  & 40.9{\tiny $\pm$0.9}  & 68.1{\tiny $\pm$6.8}     & 73.2{\tiny $\pm$0.0} & 73.0{\tiny $\pm$5.4}  & 30.3{\tiny $\pm$0.6}  & 11.6{\tiny $\pm$3.2}   & 26.1{\tiny $\pm$14.4} & 49.2{\tiny $\pm$2.1}  & 11.4{\tiny $\pm$0.0} & 50.8{\tiny $\pm$0.1}     & 81.6{\tiny $\pm$0.8}        & 35.6{\tiny $\pm$5.0}  & 20.7{\tiny $\pm$0.9}  & 64.7{\tiny $\pm$1.2}  & 59.2{\tiny $\pm$0.7}  & 51.7{\tiny $\pm$0.8}   & 90.5{\tiny $\pm$0.8}  & 12.8{\tiny $\pm$1.2}  & 15.5{\tiny $\pm$0.3} & 17.5{\tiny $\pm$0.5} & 50.6{\tiny $\pm$0.4}   \\ 
\hline
X-Decoder (B)     & 0 & 0.0M & 27.7 & 13.0 & 45.9 & 0.3 & 13.6 & 36.8 & 4.2  & 68.0 & 76.7 & 30.2 & 19.4 & 20.6 & 18.5 & 6.7 & 51.7 & 53.1 & 8.9  & 5.6  & 55.4 & 0.8 & 18.2 & 81.6 & 8.0  & 13.9 & 27.3 & 13.0  \\
X-Decoder (B)     & 1    & 0.26M & 27.7 & 13.0{\tiny $\pm$0.0}    & 45.9{\tiny $\pm$0.0}  & 0.3{\tiny $\pm$0.0}  & 13.6{\tiny $\pm$1.1}  & 36.8{\tiny $\pm$0.0}  & 4.2{\tiny $\pm$0.0}      & 68.0{\tiny $\pm$0.0} & 76.7{\tiny $\pm$0.0}  & 30.2{\tiny $\pm$0.0}  & 19.4{\tiny $\pm$0.0}   & 20.6{\tiny $\pm$0.0}  & 18.5{\tiny $\pm$0.0}  & 6.7{\tiny $\pm$5.8}  & 51.7{\tiny $\pm$0.0}     & 53.0{\tiny $\pm$4.6}        & 9.9{\tiny $\pm$1.6}   & 5.6{\tiny $\pm$0.0}   & 55.4{\tiny $\pm$0.0}  & 0.8{\tiny $\pm$0.0}   & 18.2{\tiny $\pm$0.0}   & 81.6{\tiny $\pm$0.0}  & 8.0{\tiny $\pm$0.0}   & 13.9{\tiny $\pm$0.0} & 27.3{\tiny $\pm$0.0} & 13.0{\tiny $\pm$0.0}   \\
X-Decoder (B)     & 3    & 0.26M & 31.9 & 13.1{\tiny $\pm$0.4}    & 47.4{\tiny $\pm$3.9}  & 0.3{\tiny $\pm$0.1}  & 18.9{\tiny $\pm$9.2}  & 38.1{\tiny $\pm$2.2}  & 9.4{\tiny $\pm$5.4}      & 69.6{\tiny $\pm$1.4} & 76.8{\tiny $\pm$0.0}  & 30.9{\tiny $\pm$1.1}  & 19.4{\tiny $\pm$0.0}   & 84.9{\tiny $\pm$1.7}  & 32.2{\tiny $\pm$23.7} & 8.2{\tiny $\pm$1.3}  & 52.8{\tiny $\pm$1.9}     & 53.0{\tiny $\pm$4.6}        & 10.9{\tiny $\pm$2.0}  & 5.3{\tiny $\pm$0.7}   & 55.4{\tiny $\pm$0.0}  & 0.8{\tiny $\pm$0.0}   & 18.5{\tiny $\pm$1.1}   & 81.6{\tiny $\pm$0.0}  & 8.5{\tiny $\pm$1.3}   & 15.5{\tiny $\pm$2.6} & 27.4{\tiny $\pm$0.2} & 17.5{\tiny $\pm$1.2}   \\
X-Decoder (B)     & 5    & 0.26M & 35.4 & 12.6{\tiny $\pm$0.5}    & 48.7{\tiny $\pm$1.8}  & 1.0{\tiny $\pm$0.8}  & 19.0{\tiny $\pm$16.0} & 37.1{\tiny $\pm$1.2}  & 21.3{\tiny $\pm$21.3}    & 67.9{\tiny $\pm$1.8} & 79.4{\tiny $\pm$3.6}  & 32.2{\tiny $\pm$1.7}  & 32.2{\tiny $\pm$3.0}   & 82.8{\tiny $\pm$7.8}  & 63.4{\tiny $\pm$5.0}  & 8.6{\tiny $\pm$0.2}  & 53.7{\tiny $\pm$2.7}     & 65.6{\tiny $\pm$10.9}       & 17.8{\tiny $\pm$6.9}  & 5.0{\tiny $\pm$1.4}   & 54.2{\tiny $\pm$2.1}  & 1.2{\tiny $\pm$0.4}   & 21.0{\tiny $\pm$1.3}   & 81.6{\tiny $\pm$10.0} & 8.6{\tiny $\pm$3.3}   & 18.2{\tiny $\pm$6.6} & 30.5{\tiny $\pm$0.6} & 19.2{\tiny $\pm$3.2}   \\
X-Decoder (B)     & 10   & 0.26M & 41.0 & 13.7{\tiny $\pm$0.8}    & 42.4{\tiny $\pm$2.8}  & 4.0{\tiny $\pm$2.2}  & 50.8{\tiny $\pm$3.8}  & 40.3{\tiny $\pm$0.8}  & 70.9{\tiny $\pm$6.9}     & 68.8{\tiny $\pm$2.9} & 78.1{\tiny $\pm$0.6}  & 30.8{\tiny $\pm$6.4}  & 40.4{\tiny $\pm$9.1}   & 76.6{\tiny $\pm$23.9} & 63.0{\tiny $\pm$3.6}  & 10.7{\tiny $\pm$0.6} & 60.6{\tiny $\pm$0.6}     & 69.8{\tiny $\pm$1.7}        & 21.0{\tiny $\pm$20.7} & 5.6{\tiny $\pm$0.0}   & 55.4{\tiny $\pm$1.4}  & 4.4{\tiny $\pm$2.1}   & 27.8{\tiny $\pm$5.4}   & 88.3{\tiny $\pm$2.1}  & 22.4{\tiny $\pm$8.9}  & 14.2{\tiny $\pm$2.7} & 31.1{\tiny $\pm$2.3} & 31.4{\tiny $\pm$9.1}   \\
X-Decoder (B)     & 3$\times$All & 0.26M & 44.7 & 13.0{\tiny $\pm$0.0}    & 43.8{\tiny $\pm$2.8}  & 3.3{\tiny $\pm$0.0}  & 15.4{\tiny $\pm$3.1}  & 36.5{\tiny $\pm$1.2}  & 69.3{\tiny $\pm$9.3}     & 72.2{\tiny $\pm$0.5} & 79.6{\tiny $\pm$1.1}  & 34.0{\tiny $\pm$0.7}  & 38.9{\tiny $\pm$1.0}   & 89.4{\tiny $\pm$3.3}  & 74.8{\tiny $\pm$0.9}  & 14.1{\tiny $\pm$0.2} & 57.9{\tiny $\pm$1.9}     & 84.0{\tiny $\pm$0.4}        & 17.8{\tiny $\pm$3.1}  & 5.1{\tiny $\pm$0.4}   & 55.9{\tiny $\pm$0.3}  & 57.5{\tiny $\pm$1.8}  & 48.4{\tiny $\pm$0.4}   & 90.0{\tiny $\pm$0.1}  & 18.4{\tiny $\pm$0.2}  & 21.0{\tiny $\pm$4.0} & 38.3{\tiny $\pm$0.5} & 37.0{\tiny $\pm$0.5}   \\ 
\hline
X-Decoder (L-IN21K)      & 0 & 0.0M & 26.7 & 12.3 & 43.2 & 0.5 & 3.5  & 12.3 & 18.8 & 63.9 & 79.1 & 24.3 & 15.6 & 0.0  & 20.3 & 4.9 & 50.5 & 58.8 & 43.4 & 13.4 & 57.3 & 1.3 & 12.3 & 74.4 & 6.9  & 14.6 & 20.1 & 13.5  \\
X-Decoder (L-IN21K)      & 1    & 0.26M & 25.9 & 12.3{\tiny $\pm$0.0}    & 28.8{\tiny $\pm$24.9} & 0.5{\tiny $\pm$0.0}  & 3.5{\tiny $\pm$0.0}   & 12.3{\tiny $\pm$1.1}  & 18.8{\tiny $\pm$0.0}     & 63.9{\tiny $\pm$4.6} & 79.1{\tiny $\pm$0.0}  & 24.3{\tiny $\pm$0.0}  & 15.6{\tiny $\pm$1.1}   & 0.0{\tiny $\pm$0.0}   & 20.3{\tiny $\pm$0.0}  & 4.9{\tiny $\pm$0.0}  & 50.5{\tiny $\pm$0.0}     & 58.8{\tiny $\pm$0.0}        & 43.4{\tiny $\pm$0.0}  & 11.2{\tiny $\pm$3.8}  & 57.3{\tiny $\pm$0.0}  & 1.3{\tiny $\pm$0.0}   & 12.3{\tiny $\pm$0.0}   & 74.4{\tiny $\pm$0.0}  & 5.4{\tiny $\pm$2.5}   & 14.6{\tiny $\pm$0.0} & 20.1{\tiny $\pm$0.0} & 13.5{\tiny $\pm$0.0}   \\
X-Decoder (L-IN21K)      & 3    & 0.26M & 29.1 & 12.6{\tiny $\pm$2.1}    & 44.7{\tiny $\pm$2.6}  & 1.1{\tiny $\pm$0.9}  & 8.0{\tiny $\pm$11.0}  & 15.7{\tiny $\pm$3.0}  & 32.7{\tiny $\pm$24.0}    & 63.9{\tiny $\pm$4.6} & 76.7{\tiny $\pm$4.4}  & 24.5{\tiny $\pm$0.4}  & 15.6{\tiny $\pm$1.1}   & 30.2{\tiny $\pm$52.4} & 16.9{\tiny $\pm$5.9}  & 6.8{\tiny $\pm$2.0}  & 51.0{\tiny $\pm$0.7}     & 61.1{\tiny $\pm$4.0}        & 43.0{\tiny $\pm$3.8}  & 14.6{\tiny $\pm$5.8}  & 57.5{\tiny $\pm$3.1}  & 1.4{\tiny $\pm$0.1}   & 12.3{\tiny $\pm$0.0}   & 74.0{\tiny $\pm$0.7}  & 4.4{\tiny $\pm$2.7}   & 14.8{\tiny $\pm$0.3} & 20.8{\tiny $\pm$1.2} & 21.0{\tiny $\pm$10.7}  \\
X-Decoder (L-IN21K)      & 5    & 0.26M & 33.7 & 13.9{\tiny $\pm$1.0}    & 46.4{\tiny $\pm$5.9}  & 2.1{\tiny $\pm$2.0}  & 9.5{\tiny $\pm$8.5}   & 31.4{\tiny $\pm$1.3}  & 52.6{\tiny $\pm$12.3}    & 64.1{\tiny $\pm$0.6} & 78.0{\tiny $\pm$6.2}  & 32.9{\tiny $\pm$1.2}  & 19.2{\tiny $\pm$8.2}   & 71.0{\tiny $\pm$24.3} & 26.3{\tiny $\pm$20.1} & 7.5{\tiny $\pm$0.8}  & 54.9{\tiny $\pm$4.7}     & 66.6{\tiny $\pm$1.7}        & 32.6{\tiny $\pm$9.3}  & 10.3{\tiny $\pm$9.2}  & 59.0{\tiny $\pm$1.6}  & 1.3{\tiny $\pm$0.6}   & 15.5{\tiny $\pm$8.3}   & 80.1{\tiny $\pm$5.5}  & 3.0{\tiny $\pm$3.6}   & 14.0{\tiny $\pm$5.0} & 26.6{\tiny $\pm$9.2} & 22.3{\tiny $\pm$8.5}   \\
X-Decoder (L-IN21K)      & 10   & 0.26M & 35.1 & 10.4{\tiny $\pm$1.5}    & 39.1{\tiny $\pm$7.2}  & 4.4{\tiny $\pm$1.1}  & 31.7{\tiny $\pm$8.5}  & 24.7{\tiny $\pm$11.7} & 55.8{\tiny $\pm$6.0}     & 61.4{\tiny $\pm$6.1} & 73.9{\tiny $\pm$7.1}  & 28.6{\tiny $\pm$1.8}  & 17.5{\tiny $\pm$10.2}  & 85.4{\tiny $\pm$7.0}  & 40.8{\tiny $\pm$28.3} & 6.4{\tiny $\pm$2.1}  & 58.4{\tiny $\pm$2.9}     & 54.2{\tiny $\pm$4.3}        & 32.2{\tiny $\pm$19.3} & 13.4{\tiny $\pm$0.0}  & 40.2{\tiny $\pm$13.5} & 2.2{\tiny $\pm$2.1}   & 20.8{\tiny $\pm$14.8}  & 81.0{\tiny $\pm$2.0}  & 17.9{\tiny $\pm$15.8} & 17.6{\tiny $\pm$3.1} & 26.4{\tiny $\pm$4.4} & 31.9{\tiny $\pm$9.4}   \\
X-Decoder (L-IN21K)      & 3$\times$All & 0.26M & 44.5 & 12.1{\tiny $\pm$0.8}    & 57.0{\tiny $\pm$0.4}  & 1.5{\tiny $\pm$0.5}  & 4.9{\tiny $\pm$2.3}   & 41.4{\tiny $\pm$1.0}  & 74.7{\tiny $\pm$2.9}     & 70.3{\tiny $\pm$0.4} & 79.1{\tiny $\pm$2.6}  & 36.6{\tiny $\pm$0.7}  & 23.6{\tiny $\pm$3.8}   & 54.6{\tiny $\pm$2.3}  & 70.0{\tiny $\pm$1.6}  & 12.7{\tiny $\pm$0.1} & 60.1{\tiny $\pm$0.0}     & 86.1{\tiny $\pm$0.3}        & 43.1{\tiny $\pm$2.1}  & 5.2{\tiny $\pm$2.2}   & 59.7{\tiny $\pm$0.9}  & 46.6{\tiny $\pm$0.8}  & 52.0{\tiny $\pm$0.4}   & 91.0{\tiny $\pm$0.0}  & 23.0{\tiny $\pm$5.5}  & 22.7{\tiny $\pm$1.7} & 43.8{\tiny $\pm$0.5} & 40.5{\tiny $\pm$1.3}   \\ 
\hline
X-Decoder (L)     & 0 & 0.0M & 32.3 & 13.1 & 42.1 & 2.2 & 8.6  & 44.9 & 7.5  & 66.0 & 79.2 & 33.0 & 11.6 & 75.9 & 42.1 & 7.0 & 53.0 & 68.4 & 15.6 & 20.1 & 59.0 & 2.3 & 19.0 & 67.1 & 22.5 & 9.9  & 22.3 & 13.8  \\
X-Decoder (L)     & 1    & 0.26M & 32.3 & 13.1{\tiny $\pm$0.0}    & 42.1{\tiny $\pm$0.0}  & 2.2{\tiny $\pm$0.0}  & 8.6{\tiny $\pm$0.0}   & 44.9{\tiny $\pm$0.0}  & 7.5{\tiny $\pm$0.0}      & 66.0{\tiny $\pm$0.0} & 79.2{\tiny $\pm$0.0}  & 33.0{\tiny $\pm$0.0}  & 11.6{\tiny $\pm$1.1}   & 75.9{\tiny $\pm$0.0}  & 42.1{\tiny $\pm$0.0}  & 7.0{\tiny $\pm$0.0}  & 53.0{\tiny $\pm$0.0}     & 68.4{\tiny $\pm$0.0}        & 15.6{\tiny $\pm$1.1}  & 20.1{\tiny $\pm$0.0}  & 59.0{\tiny $\pm$0.0}  & 2.3{\tiny $\pm$0.0}   & 19.0{\tiny $\pm$0.0}   & 67.1{\tiny $\pm$0.0}  & 22.5{\tiny $\pm$2.3}  & 9.9{\tiny $\pm$0.0}  & 22.3{\tiny $\pm$0.0} & 13.8{\tiny $\pm$0.0}   \\
X-Decoder (L)     & 3    & 0.26M & 33.2 & 12.9{\tiny $\pm$0.4}    & 45.9{\tiny $\pm$6.4}  & 1.8{\tiny $\pm$0.6}  & 8.6{\tiny $\pm$0.0}   & 44.9{\tiny $\pm$0.0}  & 7.5{\tiny $\pm$0.0}      & 66.0{\tiny $\pm$0.0} & 79.2{\tiny $\pm$0.0}  & 33.0{\tiny $\pm$0.0}  & 13.2{\tiny $\pm$2.7}   & 75.9{\tiny $\pm$0.0}  & 42.1{\tiny $\pm$0.0}  & 7.2{\tiny $\pm$0.3}  & 53.0{\tiny $\pm$0.0}     & 68.4{\tiny $\pm$0.0}        & 18.1{\tiny $\pm$0.9}  & 22.4{\tiny $\pm$1.9}  & 59.0{\tiny $\pm$0.0}  & 2.3{\tiny $\pm$0.0}   & 19.8{\tiny $\pm$1.3}   & 67.1{\tiny $\pm$0.0}  & 26.0{\tiny $\pm$6.0}  & 9.6{\tiny $\pm$0.4}  & 25.8{\tiny $\pm$6.0} & 18.3{\tiny $\pm$5.1}   \\
X-Decoder (L)     & 5    & 0.26M & 35.9 & 12.5{\tiny $\pm$0.5}    & 44.9{\tiny $\pm$2.2}  & 2.4{\tiny $\pm$2.5}  & 28.4{\tiny $\pm$6.8}  & 44.9{\tiny $\pm$1.0}  & 15.7{\tiny $\pm$1.6}     & 67.1{\tiny $\pm$2.0} & 77.1{\tiny $\pm$0.2}  & 36.3{\tiny $\pm$0.7}  & 9.8{\tiny $\pm$8.7}    & 93.1{\tiny $\pm$0.9}  & 45.6{\tiny $\pm$7.5}  & 7.6{\tiny $\pm$1.0}  & 53.0{\tiny $\pm$0.8}     & 71.3{\tiny $\pm$3.1}        & 19.4{\tiny $\pm$3.5}  & 22.5{\tiny $\pm$1.7}  & 55.8{\tiny $\pm$2.2}  & 2.4{\tiny $\pm$0.5}   & 12.0{\tiny $\pm$2.1}   & 78.4{\tiny $\pm$6.0}  & 30.1{\tiny $\pm$6.3}  & 10.0{\tiny $\pm$1.3} & 30.1{\tiny $\pm$1.7} & 25.9{\tiny $\pm$2.8}   \\
X-Decoder (L)     & 10   & 0.26M & 40.3 & 14.0{\tiny $\pm$1.5}    & 33.7{\tiny $\pm$14.6} & 4.4{\tiny $\pm$5.9}  & 41.2{\tiny $\pm$3.0}  & 44.6{\tiny $\pm$1.7}  & 73.0{\tiny $\pm$2.2}     & 68.8{\tiny $\pm$5.5} & 79.4{\tiny $\pm$1.8}  & 39.2{\tiny $\pm$3.4}  & 17.5{\tiny $\pm$4.7}   & 93.9{\tiny $\pm$0.6}  & 53.9{\tiny $\pm$3.1}  & 8.8{\tiny $\pm$3.0}  & 52.5{\tiny $\pm$2.5}     & 77.3{\tiny $\pm$0.9}        & 24.0{\tiny $\pm$7.0}  & 20.1{\tiny $\pm$0.0}  & 55.3{\tiny $\pm$0.7}  & 3.0{\tiny $\pm$1.6}   & 15.0{\tiny $\pm$6.6}   & 72.7{\tiny $\pm$16.1} & 39.1{\tiny $\pm$7.8}  & 9.0{\tiny $\pm$5.5}  & 32.1{\tiny $\pm$2.4} & 32.5{\tiny $\pm$9.4}   \\
X-Decoder (L)     & 3$\times$All & 0.26M & 44.7 & 13.6{\tiny $\pm$0.3}    & 49.3{\tiny $\pm$0.6}  & 4.2{\tiny $\pm$1.1}  & 8.8{\tiny $\pm$0.4}   & 44.6{\tiny $\pm$0.3}  & 51.1{\tiny $\pm$0.8}     & 72.8{\tiny $\pm$0.6} & 78.9{\tiny $\pm$1.2}  & 42.0{\tiny $\pm$0.0}  & 13.8{\tiny $\pm$3.1}   & 90.2{\tiny $\pm$0.6}  & 67.8{\tiny $\pm$0.3}  & 11.8{\tiny $\pm$0.1} & 52.7{\tiny $\pm$0.4}     & 84.3{\tiny $\pm$0.0}        & 18.9{\tiny $\pm$1.2}  & 21.6{\tiny $\pm$1.6}  & 54.3{\tiny $\pm$2.3}  & 56.8{\tiny $\pm$3.5}  & 50.4{\tiny $\pm$0.6}   & 90.7{\tiny $\pm$0.2}  & 42.1{\tiny $\pm$2.7}  & 11.1{\tiny $\pm$1.0} & 39.7{\tiny $\pm$1.2} & 44.3{\tiny $\pm$2.9}   \\
\hline
\end{tabular}
}
\vspace{3pt}
\caption{SegInW results with tuning on X-Decoder for different image shots and backbone architectures. (0.26M parameters tuned in the setting.)}
\label{tab:seginw_finetune}
\end{table*}

\clearpage


\end{document}